\documentclass[twoside,11pt]{article}

\usepackage{blindtext}

\usepackage[preprint]{jmlr2e}

\PassOptionsToPackage{breaklinks,colorlinks}{hyperref}
\usepackage[utf8]{inputenc}               
\usepackage[T1]{fontenc}                 
\usepackage[breaklinks,colorlinks]{hyperref} 
\usepackage{url}                    
\usepackage{booktabs}                  
\usepackage{amsfonts}                  
\usepackage{nicefrac}                  
\usepackage{microtype}                 
\usepackage{xcolor}     
\usepackage{graphicx}
\usepackage{tabularray}
\usepackage{float}
\usepackage{amsmath}
\usepackage{mathtools}
\usepackage{bm}
\usepackage{multirow}
\usepackage{listings}
\usepackage{pythonhighlight}

\newcommand{\dataset}[0]{ClimSim}
\newcommand{\datasetonline}[0]{ClimSim-Online}

\usepackage{lastpage}
\jmlrheading{25}{2024}{1-\pageref{LastPage}}{6/30}{x/xx}{xx-xxxx}{Yu and Hu et al.}

\ShortHeadings{\datasetonline{}: Dataset and Framework for Hybrid Climate Emulation}{Yu and Hu et al.}
\firstpageno{1}

\begin{document}

\title{\datasetonline{}: A Large Multi-scale Dataset\\and Framework for Hybrid ML-physics Climate Emulation}

% \author{\name Author One \email one@stat.washington.edu \\
%        \name Author Two \email two@cs.berkeley.edu \\
%        \name Author Three \email three@stat.washington.edu \\
%        \name Author Four \email four@cs.berkeley.edu \\
%        \AND

\author{Sungduk Yu$^{1, 2}$\thanks{Equal contribution}, Zeyuan Hu$^{3,4}$$^*$\thanks{Correspondence: zeyuan\_hu@fas.harvard.edu}, Akshay Subramaniam$^{3}$, Walter Hannah$^5$, Liran Peng$^1$, Jerry Lin$^1$, Mohamed Aziz Bhouri$^6$, \textbf{Ritwik Gupta$^7$, Björn Lütjens$^{8}$, Justus C. Will$^1$, Gunnar Behrens$^9$, Julius J. M. Busecke$^6$, Nora Loose$^{10}$, Charles I Stern$^6$} \textbf{Tom Beucler$^{11}$, Bryce Harrop$^{12}$, Helge Heuer$^9$, Benjamin R Hillman$^{13}$, Andrea Jenney$^{14}$, Nana Liu$^{15}$, Alistair White$^{16, 17}$,} \textbf{Tian Zheng$^6$, Zhiming Kuang$^{4}$, Fiaz Ahmed$^{18}$, Elizabeth Barnes$^{19}$, Noah D. Brenowitz$^{3}$, Christopher Bretherton$^{20}$,} \textbf{Veronika Eyring$^9$, Savannah Ferretti$^{1}$, Nicholas Lutsko$^{21}$, Pierre Gentine$^6$,} \textbf{Stephan Mandt$^1$, J. David Neelin$^{18}$, Rose Yu$^{21}$, Laure Zanna$^{22}$,} \textbf{Nathan Urban$^{23}$, Janni Yuval$^{24}$, Ryan Abernathey$^6$, Pierre Baldi$^1$, Wayne Chuang$^{6}$,} \textbf{Yu Huang$^6$, Fernando Iglesias-Suarez$^{25}$, Sanket Jantre$^{23}$, Po-Lun Ma$^{12}$, Sara Shamekh$^{22}$, Guang Zhang$^{21}$,} \textbf{Michael Pritchard$^{1, 3}$}
\\
\\
$^1$UCI, $^{2}$Intel Labs, $^{3}$NVIDIA, $^{4}$Harvard, $^5$LLNL, $^6$Columbia, $^7$UCB, $^8$MIT, $^9$DLR, $^{10}$Princeton, $^{11}$UNIL, $^{12}$PNNL, $^{13}$SNL, $^{14}$OSU, $^{15}$CIWRO/NOAA, $^{16}$PIK, $^{17}$TUM, $^{18}$UCLA, $^{19}$CSU, $^{20}$Allen AI, $^{21}$UCSD, $^{22}$NYU, $^{23}$BNL, $^{24}$Google Research, $^{25}$PRED}

\editor{}

\maketitle

\begin{abstract}%   <- trailing '%' for backward compatibility of .sty file
Modern climate projections lack adequate spatial and temporal resolution due to computational constraints, leading to inaccuracies in representing critical processes like thunderstorms that occur on the sub-resolution scale. Hybrid methods combining physics with machine learning (ML) offer faster, higher fidelity climate simulations by outsourcing compute-hungry, high-resolution simulations to ML emulators. However, these hybrid ML-physics simulations require domain-specific data and workflows that have been inaccessible to many ML experts. 

As an extension of the \dataset{} dataset \citep{yu2024climsim}, we present \datasetonline{}, which also includes an end-to-end workflow for developing hybrid ML-physics simulators. The \dataset{} dataset includes 5.7 billion pairs of multivariate input/output vectors, capturing the influence of high-resolution, high-fidelity physics on a host climate simulator's macro-scale state.

The dataset is global and spans ten years at a high sampling frequency. We provide a cross-platform, containerized pipeline to integrate ML models into operational climate simulators for hybrid testing. We also implement various ML baselines, alongside a hybrid baseline simulator, to highlight the ML challenges of building stable, skillful emulators. The data (\url{https://huggingface.co/datasets/LEAP/ClimSim_high-res}\footnote{Also in a low-resolution version (\url{https://huggingface.co/datasets/LEAP/ClimSim_low-res}) and an aquaplanet version (\url{https://huggingface.co/datasets/LEAP/ClimSim_low-res_aqua-planet}).}) and code (\url{https://leap-stc.github.io/ClimSim} and \url{https://github.com/leap-stc/climsim-online/}) are publicly released to support the development of hybrid ML-physics and high-fidelity climate simulations.

\end{abstract}


\section{Introduction}
\label{sec:introduction}
% ===============================
\subsection{Overview}
% ===============================

Projections from numerical Earth system model simulations are the primary tool informing climate change policy \citep{Tebaldi2021esd}. However, current climate simulators poorly represent clouds and extreme rainfall physics \citep{IPCC2021, Sherwood2020} despite stretching the limits of the world's most powerful supercomputers. This is because the required computational power to simulate Earth system complexity imposes significant restrictions on the simulations' spatial resolution \citep{Schneider2017,Gentine2021}. Physics occurring on scales smaller than the temporal and/or spatial resolutions of climate simulations are commonly represented using empirical or physically inspired mathematical representations called ``parameterizations''. Assumptions in these parameterizations often lead to errors that can grow into inaccuracies in projections of future climates. 

% SY: The paragraph was too long, so i broke it into 3 smaller chunks.
Machine learning (ML) is an attractive approach for learning the complex nonlinear sub-resolution physics---processes (and properties) occurring on scales smaller than typical climate model resolution---from short, higher fidelity simulations. The implementation of ML-physics parameterizations has the exciting possibility of resulting in hybrid climate simulations that are both cheaper and more accurate than current state-of-the-art model simulations \citep{Gentine2018, Eyring2021}. 

Traditional Earth system models have a typical smallest resolvable scale of 80--200 km in the horizontal direction \citep{eyring2016overview}, equivalent to the size of a typical U.S. county. In contrast, the community has achieved 1-10 km resolution, global storm-resolving simulators, though these models are still being tested for their use in long-term climate simulations and are very computationally demanding \citep[e.g.,][]{taylor2023simple,hohenegger2023icon, mooers2023comparing}. Accurately representing cloud formation requires a resolution of 100 m or finer, demanding six orders of magnitude increase in computational power. 

ML presents a conceivable solution to sidestep the limitations of classical computing \citep{Eyring2021,eyring24_hybridmodelling,eyring24natcc_AGCIMLClimate}. It enables hybrid-ML climate simulations that integrate traditional numerical methods---which solve the equations governing large-scale fluid motions of Earth's atmosphere---with ML-based parameterizations that emulate the macro-scale effects of small-scale physics. Instead of relying on heuristic assumptions about these small-scale processes, ML-based parameterizations learn directly from data generated by short-duration, high-resolution simulations \citep{Bretherton2022, Clark2022, Grundner2022Cloud, Sanford2023, Gentine2018, Rasp2018, Brenowitz2020, Han2020, ott2020, Mooers2021, Wang2022a, han2023ensemble, Iglesias2023, Yuval2020, Yuval2021, heuer2023interpretable, niu2024multi}. This task is a nonlinear regression problem: in the climate simulation, an ML-physics parameterization returns the large-scale outputs---changes in wind, moisture, or temperature---that occur due to unresolved small-scale (sub-resolution) physics, given large-scale resolved inputs (e.g., temperature, wind velocity; see Section \ref{sec:experiments}). 

While several proofs of concept have emerged in recent years, hybrid-ML climate simulations have yet to be stable and ready for operational use. Obtaining sufficient, complete training data is a major challenge impeding progress from the ML community. This data must contain all macro-scale variables that regulate the behavior of subgrid-scale physics and be compatible with hybrid ML-climate simulations. Addressing this using training data from uniformly high-resolution simulations has proven to be very expensive and requires coarse-graining the high-resolution data, potentially leading to issues when coupled with a host climate simulation \citep{ross2023benchmarking}.

A promising solution is utilizing multi-scale climate (see Section \ref{sec:concepts}) simulations to generate training data. Crucially, these provide a clean interface between the learned high-resolution physics and the host climate simulator's macro-scale dynamics \citep{Rasp2020b}. In theory, this makes hybrid simulations approachable and tractable. In practice, the full potential of multi-scale simulations remains largely untapped due to a scarcity of existing datasets, exacerbated by the combination of operational simulation code complexity and the need for domain expertise in choosing variables. To further complicate matters, the absence of a straightforward method for testing learned ML emulators in hybrid settings renders the problem even less approachable.

We introduce \datasetonline{}, the largest and most physically comprehensive dataset and end-to-end workflow for training and testing ML-based parameterizations of full subgrid-scale physics (atmospheric storms, clouds, turbulence, rainfall, and radiation) for use in hybrid-ML climate simulations. The \dataset{} dataset \citep{yu2024climsim} offers a comprehensive collection of inputs and outputs from multi-scale climate model simulations. As an extension to the \dataset{} dataset, \datasetonline{} also provides a containerized, end-to-end workflow for integrating ML models into host climate simulations, facilitating the evaluation of online performance (see Section \ref{sec:concepts} and \ref{sec:online}) in hybrid-ML simulations. This containerized workflow ensures reproducibility and ease of use, making it accessible to ML researchers without domain knowledge. \datasetonline{} was prepared by atmospheric scientists, climate model developers, and ML researchers to lower the entry barrier for ML experts on this important problem. Our benchmark dataset serves as a foundation for developing robust frameworks that learn subgrid-scale processes related to cloud physics. This framework enables online coupling within the host climate model, with the ultimate goal to help improve the performance and accuracy of climate models used for long-term projections.

% ===============================
\subsection{Concepts and Terminology from Earth Science}
% ===============================
\label{sec:concepts}

\textbf{Convection Parameterization:} In atmospheric science, ``convection'' refers to storm cloud and rain development, as well as the associated small-scale (100s m to $<$10km) turbulent air motions. Convective parameterizations represent the combined effects of these processes, such as the vertical transport of heat, moisture, and momentum within the atmosphere, and condensational heating and drying, on the temporal and spatial scale of the host climate model \citep{Emanuel1994, Randall2012, Siebesma2020}. Stochastic parameterizations represent sub-resolution (``sub-grid scale'' in the terminology of Earth science) effects as stochastic processes, dependent on grid-scale variable inputs \citep{Lin2000, Neelin2008} to capture variations arising from sub-grid scale dynamics.

\textbf{Multi-Scale Climate Simulations:} Multi-scale climate simulation is a technique that represents convection without a convective parameterization by deploying a smaller-scale, high-resolution cloud-resolving simulator nested within each host grid column of a climate simulator \citep{Grabowski1999, Benedict2009, Randall2013, Hannah2020, Norman2022}. The smaller-scale simulator explicitly resolves the detailed behavior of clouds and their turbulent motions at both a higher spatial and temporal resolution than the host climate model. This improves the accuracy of the host simulations but comes at a high computational cost \citep{Randall2003, Khairoutdinov2008}. The time-integrated and horizontally averaged influence of the resolved convection is fed upscale to the host climate model and is the target of hybrid ML-climate simulation approaches.

\textbf{Significance of Precipitation Processes for Climate Impacts:} In climate simulations, changes in precipitation with climate change are a particularly important issue. The frequency of extreme precipitation events increases with climate change \citep{Pall2007, Guerreiro2018, Neelin2022,seneviratne2021ipcc, mooers2022understanding}, with corresponding  societal impacts \citep{Davenport2021}. Current climate models agree on the direction of this change but exhibit a large spread in the quantitative rate of increase with climate change \citep{Pendergrass2014, Martinez-Villalobos2023,seneviratne2021ipcc}. 

\textbf{Offline Training vs. Online Evaluation:} In this manuscript, we define ``offline training'' as the traditional supervised learning task, where a regression or generative machine learning (ML) model is trained to map input features to target features. Offline metrics (Section \ref{subsection:offline_metrics}) assess how well an ML model performs this mapping for samples at each individual location and time step, using a fixed dataset. 

On the other hand, ``Online evaluation'' has domain-specific meaning in our context, and refers to evaluating the performance of the hybrid climate simulator in which many instances of its embedded high resolution physics solver are replaced with copies of a trained ML parameterization that is then allowed to feed back with resolved planetary scale climate dynamics. That is, the evaluation data changes dynamically. This involves assessing how accurately the hybrid simulation can replicate climate statistics, such as yearly-mean atmospheric states, compared to a pure physical climate simulation (see Section \ref{subsec:online_metric} for more online evaluation metrics).

% ======================================================
\section{Related Work}
\label{sec:relatedwork}
% ======================================================

Several benchmark datasets have been developed to facilitate AI tasks in weather and climate. For example, the European Center for Medium-Range Weather Forecasts (ECMWF) reanalysis v5 \citep[ERA5,][]{hersbach2020era5} is a comprehensive dataset of global weather from 1940 to present. WeatherBench 2 benchmark dataset provides data specifically designed for training and evaluating data-driven weather forecasting models, focusing on global, medium range (1-15days) prediction \cite{rasp2024weatherbench}. ClimateBench \citep{Watson-Parris2022} was designed for emulators that produce annual mean global predictions of temperature and precipitation given greenhouse gas concentrations and emissions. ClimateBench is limited to data from a single climate model. In contrast, ClimateSet \citep{kaltenborn2023climateset} expands ClimateBench by offering a large-scale dataset with inputs and outputs from 36 climate models. ClimART \citep{Cachay2021} was designed for the development of radiative energy transfer parameterization emulators for use in weather and climate modeling. Section 7 in the SI documented more climate or weather related benchmark datasets. 

However, \dataset{} is unique for its focus on learning ML parameterizations which can be used in hybrid climate simulations. Unlike other datasets, \dataset{} is designed to capture the nonlinear effects of clouds, rain, storms, and radiation at kilometer scales. It provides a end-to-end framework to emulate an embedded component—the cloud-resolving simulator—in multi-scale climate simulators and to evaluate the resulting hybrid climate simulations.

There have been several recent efforts to produce hybrid-ML models learning from multi-scale climate simulations, analogous to \dataset{} \citep{Gentine2018, Rasp2018, Han2020, ott2020, Mooers2021, Wang2022a, lin2023systematic, Iglesias2023, han2023ensemble}. Most of these focused on simple aquaplanets \citep{Gentine2018, Rasp2018, Han2020, ott2020, lin2023systematic, Iglesias2023}, while simulations that included real geography \citep{Mooers2021, Wang2022a, han2023ensemble, heuer2023interpretable} did not include enough variables for complete land-surface coupling, to our knowledge. Most examine simple multi-layer perceptrons, except for \cite{Han2020, Wang2022a, han2023ensemble}, who used a ResNet architecture, and \cite{behrens2024improving}, who used a variational encoder-decoder that accounts for stochasticity. Although hybrid testing in real-geography settings is error-prone, several studies \citep{Wang2022a,han2023ensemble,kochkov2023neural} have demonstrated some hybrid stability. Compressing input data to avoid causal confounders may improve online accuracy \citep{Iglesias2023,kuhbacher2024towards}, and methods have been developed to enforce physical constraints \citep{Beucler2021a, Reed2023}.

Compared to the training data used above, \dataset{}'s comprehensive variable coverage is unprecedented, including all variables needed to be coupled to a land system simulator and to enforce physical constraints. Its availability across coarse-resolution, high-resolution, aquaplanet and real-geography use cases is also new to the community. Successful ML innovations with \dataset{} can have a downstream impact since it is based on state-of-the-art multi-scale climate model simulations that are actively supported by a mission agency (U.S.\ Department of Energy). 

In non-multi-scale settings, an important body of related work \citep{Bretherton2022, Clark2022, Kwa2023, Sanford2023} has made exciting progress on using analogous hybrid ML approaches to reduce biases in uniform resolution climate simulations, including in an operational climate code with land coupling and online stability \citep{Yuval2020, Yuval2021} (see Supplementary Information; SI). Other related work includes full model emulation (FME) for short-term weather prediction~\citep{Pathak2022, Bonev2023, Lam2022} and for long-term climate simulation \citep{watt2023ace}. While \dataset{} is focused on hybrid-ML climate simulations and we do not demonstrate FME baselines, \dataset{} contains full atmospheric state variable samplings well suited for the FME task.

% ======================================================
\section{\dataset{} Dataset}
\label{sec:dataset}
% ======================================================

\begin{figure}
    \centering
    \includegraphics[width=\textwidth, trim={0 1.5cm 0 1.2cm},clip]{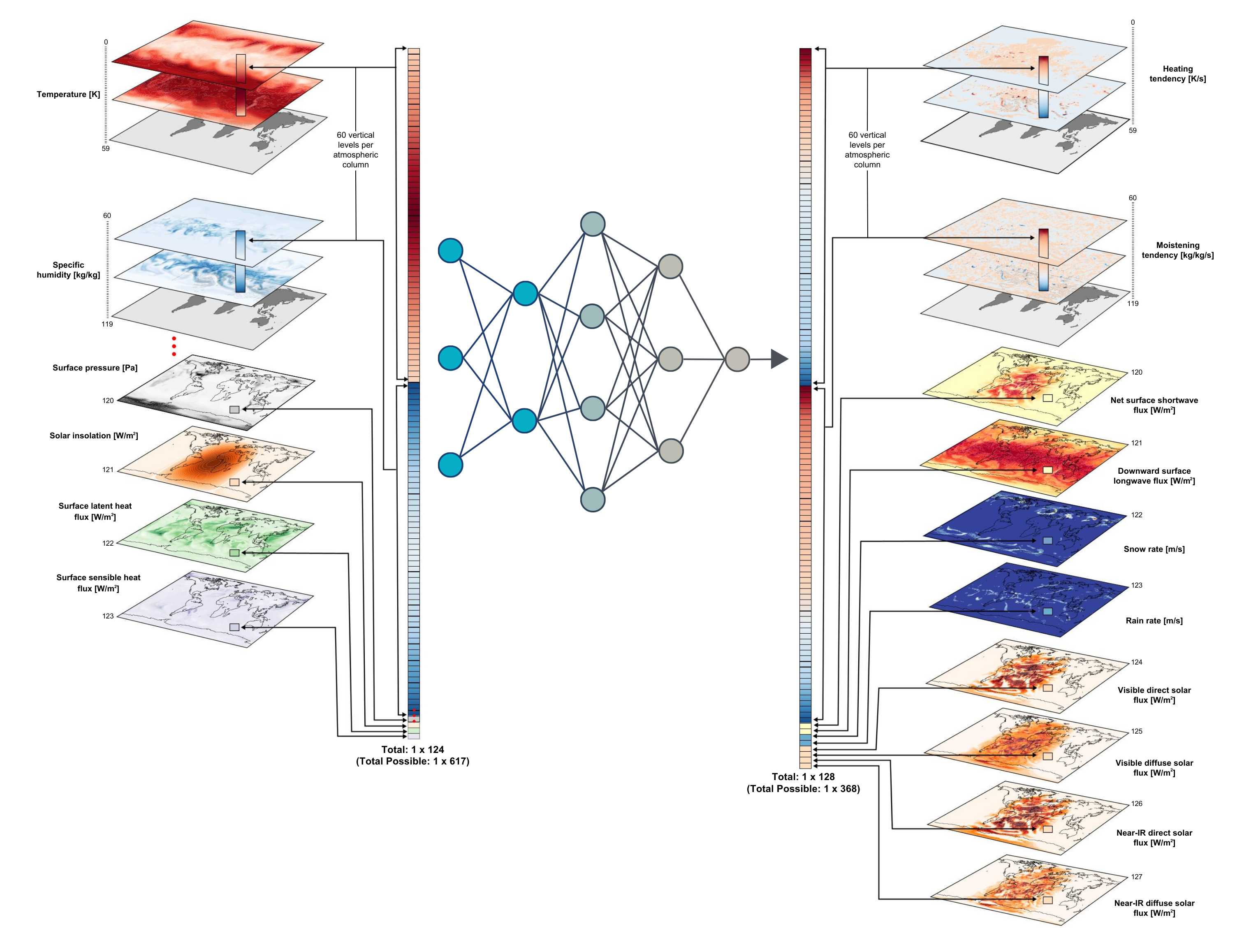}
    \setlength{\belowcaptionskip}{-0.5em}%
    \caption{The spatially-local version of \dataset{} that our baselines are scored on. A spatially-global version of the problem that expands to the full list of variables would be useful to try.}
    \label{fig:preprocessdiagram}
\end{figure}

\textbf{Dataset Overview:} \dataset{} is designed to facilitate the development of ML parameterizations for hybrid climate simulations. The dataset includes inputs $x\in\mathbb R^{d_i}$ (with $d_i$ = 124 for standard inputs and $d_i$ = 617 for expanded inputs) representing the local vertical structure of macro-scale state variables and boundary conditions, and targets $y\in\mathbb R^{d_o}$ (with $d_o$ = 128 for standard targets and $d_o$ = 368 for expanded targets) representing tendencies due to unresolved processes and surface fluxes for surface coupling. Generated using the E3SM-MMF multi-scale climate simulator over 10 simulated years, the dataset comprises 5.7 billion high-resolution samples (41.2TB) and 100 million low-resolution samples (744GB). Data is split into training/validation (first 8 years) and test (last 2 years) sets, ensuring no temporal overlap. Offline training involves learning an ML parameterization by mapping inputs to targets for each sample (an atmospheric column at a single timestep). \datasetonline{} provides an accessible end-to-end workflow to integrate a trained ML parameterization into a host climate simulator to perform hybrid simulations. Climate statistics in these hybrid simulations are evaluated against those from pure physical E3SM-MMF simulations.

\textbf{Experiment Outline:} \dataset{} presents a regression problem with mapping from a multivariate input vector, with inputs $x\in\mathbb R^{d_i}$ of size $d_i$ = 124 and targets $y\in\mathbb R^{d_o}$ of size $d_o$ = 128 (Figure \ref{fig:preprocessdiagram}). The input represents the local vertical structure (in horizontal location and time) of macro-scale state variables in a multi-scale climate simulator before any adjustments from sub-grid scale convection and radiation are made. The input also includes concatenated scalars containing boundary conditions of incoming radiation at the top of the atmospheric column and land surface model constraints at its base. The target vector contains the tendencies of the same state variables representing the impact of unresolved processes (e.g., redistribution of water, related phase changes, and radiative heating due to convection). The output vector represents the horizontally averaged change in atmospheric state after the computationally demanding subcycling of nested simulators. The ultimate goal is to outsource these physics to ML by mapping inputs to targets at comparable fidelity. The target vector includes scalar fields and fluxes from the bottom of the atmospheric column expected by the land surface model component that it must couple to; land-atmosphere coupling is crucial to predicting regional water cycle dynamics \citep{Fischer2007, Seneviratne2010}. Importantly, \dataset{} also includes the option for \textit{expanded inputs} $x\in\mathbb R^{d_i}$ of size $d_i$ = 617 and targets $y\in\mathbb R^{d_o}$ of size $d_o$ = 368, which we demonstrate in one of our experiments. 

\textbf{Dataset Collection:} We ran the E3SM-MMF multi-scale climate simulator \citep{Hannah2020, Norman2022, Hannah2022a, Hannah2021}, using multiple NVIDIA A100 GPUs for a total of $\sim$ 9,800 GPU-hours. We saved instantaneous values from every grid column of the atmospheric state before and after high-resolution calculations occurred, isolating state updates due to explicitly-resolved moist convection, boundary layer turbulence, and radiation; details of the E3SM-MMF climate simulator configuration can be found in SI. These data were saved at 20-minute intervals (i.e., the time step of the host climate model) for 10 simulated years (excluding one-month spinup), resulting in 5.7 billion samples for the high-resolution simulation that uses an unstructured ``cubed-sphere" horizontal grid with 21,600 grid columns spanning the globe. This grid yields an \textit{approximate} horizontal grid spacing of 1.5$^\circ$, but unlike a traditional climate model that maps points across the sphere using two dimensions aligned with cardinal north/south and east/west directions, unstructured grids use a single dimension to organize the horizontal location of points. The atmospheric columns at each location and time are treated as independent samples. Thus, the total number of samples can be understood by considering that atmospheric columns at each location and time are treated as independent samples, such that 5.7 billion $\approx$ 21,600 horizontal locations per time step $\times$ 72-time steps per simulated day $\times$ 3,650 simulated days). It is important to note that each sample retains a 1D structure corresponding to the vertical variation across 60 levels. 

We also ran two additional simulations with approximately ten times less horizontal resolution, with only 384 grid columns spanning the globe, resulting in 100 million samples for each simulation. These low-resolution options allow for fast prototyping of ML models due to smaller training data volumes and less geographic complexity. One low-resolution simulation uses an ``aquaplanet'' configuration, i.e., a lower boundary condition of specified sea surface temperature, invariant in the longitudinal dimension with no seasonal cycle. This is the simplest prototyping dataset, removing variance associated with continents and time-varying boundary conditions. The total data volume is 41.2TB for the high-resolution dataset and 744GB for each of the low-resolution datasets.

The input and output variables in this dataset were selected based on the design of E3SM-MMF. We have included all variables involved in the interface between the host climate simulator and the embedded cloud-resolving simulators. Specifically, the input variables are macro-state variables calculated by the host climate simulator and passed to the embedded cloud-resolving simulators. The output variables comprise the subgrid physics tendencies, which are simulated by the embedded simulators based on these macro-state inputs. This setup defines the pairing of input and output variables. 

\textbf{Locality vs. Nonlocality:} A spatially-global version of the problem could be of practical use for improving ML via helpful spatial context \citep{Wang2022b, Lütjens2022}. In this case, information from other grid columns across the globe is taken into account. Thus, the problem becomes a 2D $\rightarrow$ 2D regressions task with inputs $x\in\mathbb R^{d_i}$ of maximum size $d_i=617 \times 21,600$ (grid columns) and targets, $y\in\mathbb R^{d_o}$, of maximum size $d_o=368 \times 21,600$. Here, the second dimension represents the unstructured "cubed-sphere" computational mesh used by the climate model, which projects a cube onto the sphere, effectively avoiding the polar singularity problems associated with regular Cartesian grids \citep{ronchi1996cubed,Hannah2022a}. Further details about the climate simulator configuration, simulations, and data, including complete variable lists, can be found in SI.

\textbf{Dataset Interface:} Raw model outputs emerge from the climate simulator as standard NetCDF files, which can be easily parsed in any language. Each timestep yields files containing input and target vectors separately, resulting in a total of 525,600 files for each of the three datasets. To prevent redundancy, variable metadata and grid information were saved separately.

The raw tensors from the climate simulations are initially either 2D or 3D, depending on the variable. For 2D tensors, the dimensions represent time and horizontal location. While these variables actually depend on three physical dimensions (time and 2D space), since each location on the sphere is indexed along a single axis due to the climate model's unstructured horizontal grid, the apparent dimensionality is lower. Such variables include solar insolation, snow depth over land, surface energy fluxes, and surface precipitation rate. 3D tensors include the additional dimension representing altitude relative to the Earth's surface for height-varying state variables like temperature, humidity, and wind vector components. Separate files are used to store each timestep and variable. \dataset{} includes a total of 24 2D variables and 10 3D variables (see Table 1 in SI).

\textbf{Dataset Split:} The 10-year datasets are divided into the following splits: (a) a training and validation set spanning the first 8 years (0001-02 to 0009-01; YYYY-MM), excluding the first simulated month for numerical spin-up, and (b) a test set spanning the remaining two years (i.e., 0009-03 to 0011-02). A one-month gap is intentionally introduced between the two sets to prevent test set contamination via temporal correlation. Both sets are stored separately in our data repositories. 

It's important to note that optimizing offline metrics does not necessarily lead to optimized online performance \citep{ott2020,Wang2022a}. Small prediction errors at each time step can accumulate over time, potentially causing biases or drifts of the atmospheric states in the hybrid simulation when compared to a purely physical reference simulation. These accumulated errors can arise from the high degrees of freedom of the host model interacting with imperfections of its embedded ML parameterization. Among other conceivable pathologies, this interaction may lead to out-of-sample atmospheric states that further degrade ML predictions and destabilize the hybrid simulation, potentially causing model blowup.

\textbf{Energy use:} The computing and energy costs of generating ClimSim could be viewed as wasteful and having a negative consequence for society through associated emissions \citep{luccioni2023counting}. We emphasize that while it can appear large, the compute used is actually orders of magnitude less than what is consumed by operational climate prediction. Associated emissions are minimized given that our E3SM-MMF simulations were performed on energy-efficient GPU hardware. The cost must also be weighed against the potential social benefit of mitigating future energy consumption by eliminating end users' need for costly physics-based multi-scale climate simulations. Meanwhile, a large consortium of interested parties has helped agree on this dataset to help ensure it is not wasted.

% ======================================================
\section{Offline Experiments}
\label{sec:experiments}
% ======================================================

To guide ML practitioners using \dataset{}, we provide an example ML workflow using the low-resolution, real-geography dataset to train ML models to predict target outputs from the provided inputs. All but one of our baselines focuses on emulating the subset of total available input and target variables illustrated in Figure \ref{fig:preprocessdiagram}, with the following inputs $x\in\mathbb R^{d_i}$ of size $d_i=124$, and targets $y\in\mathbb R^{d_o}$ of size $d_o=128$ (Figure \ref{fig:preprocessdiagram}, Table \ref{tab:baselinevars}), chosen for its similarity to recent attempts in the literature.

\begin{table}[ht]
\small
\centering
\resizebox{\textwidth}{!}{\begin{tabular}{lclc}
\toprule
\textbf{Input} & Size & \textbf{Target} & Size \\
\midrule
Temperature [K] & 60 & Heating tendency, $dT/dt$ [K/s] & 60 \\
Specific humidity [kg/kg] & 60 & Moistening tendency, $dq/dt$ [kg/kg/s] & 60 \\
Surface pressure [Pa] & 1 & Net surface shortwave flux, NETSW [W/m$^\text{2}$] & 1 \\
Insolation [W/m$^\text{2}$] & 1 & Downward surface longwave flux, FLWDS [W/m$^\text{2}$] & 1 \\
Surface latent heat flux [W/m$^\text{2}$] & 1 & Snow rate, PRECSC [m/s] & 1 \\
Surface sensible heat flux [W/m$^\text{2}$] & 1 & Rain rate, PRECC [m/s] & 1 \\
  &   & Visible direct solar flux, SOLS [W/m$^\text{2}$] & 1 \\
  &   & Near-IR direct solar flux, SOLL [W/m$^\text{2}$] & 1 \\
  &   & Visible diffused solar flux, SOLSD [W/m$^\text{2}$] & 1 \\
  &   & Near-IR diffused solar flux, SOLLD [W/m$^\text{2}$] & 1 \\
\bottomrule
\end{tabular}}
\vspace{1mm}
\setlength{\belowcaptionskip}{-1em}%
\caption{The subset of input and target variables used in most of our experiments (Figure \ref{fig:preprocessdiagram}). Dimension length 60 corresponds to the total number of vertical levels (discretized altitudes) of the climate simulator.}
\label{tab:baselinevars}
\end{table}

\textbf{Training/Validation Split:} We divide the 8-year training/validation set into the first 7 years (i.e., 0001-02 to 0008-01 in the raw filenames' ``year-month'' notation) for training and the subsequent 1 year (0008-02 to 0009-01) for validation. This split was chosen somewhat arbitrarily, and we encourage users of this dataset to consider alternative splits. However, it is crucial to ensure that the validation period is separate from the training period by at least one month to avoid contamination due to temporal autocorrelation in the atmosphere. 

\textbf{Preprocessing Workflow:} Our preprocessing steps were (1) downsample in time by using every 7th sample, (2) collapse horizontal location and time into a single sample dimension, (3) normalize variables by subtracting the mean and dividing by the range, with these statistics calculated separately at each of the 60 vertical levels for the four variables with vertical dependence, and (4) concatenate variables into multi-variate input and output vectors for each sample (Figure \ref{fig:preprocessdiagram}). The heating tendency target $dT/dt$ (i.e., time rate of temperature $T$) was calculated from the raw climate simulator output  as $(T_\mathit{after} - T_\mathit{before}) / \Delta t$, where $\Delta t$  (= 1200 s) is the climate simulator's known macro-scale timestep. Likewise, the moisture tendency was calculated by taking the difference of humidity state variables recorded before versus after the convection and radiation calculations. This target variable transformation (i.e., state to tendency) is done to compare our baseline models' performance to that of previously published models that reported errors of emulated tendencies \citep{Mooers2021, Behrens2022}. Additionally, this transformation implicitly normalizes the target variables, leading to better convergence properties for ML algorithms. Given the domain-specific nature of the preprocessing workflow, we provide scripts in the GitHub repository for workflow reproduction.

% ===============================
\subsection{Baseline Architectures}
% ===============================

Six baseline models used in our experiment are briefly described here. Refer to SI for further details. 

\textbf{Convolutional Neural Network (CNN)} uses a 1D ResNet-style network. Each ResNet block contains two 1D convolutional layers and a skip connection. CNNs can learn spatial structure and have outperformed MLP and graph-based networks~\citep{Cachay2021}. The inputs and outputs for the CNN are stacked in the channel dimensions, such that the mapping is 60 $\times$ 6 $\rightarrow$ 60 $\times$ 10. Accordingly, global variables have been repeated along the vertical dimension. % The baseline CNN has a depth of 12, each convolutional layer has a width of 406, and a kernel size of 3. It was trained for 10 epochs.

\textbf{Encoder-Decoder (ED)} consists of an Encoder and a Decoder with 6 fully-connected hidden layers each \citep{Behrens2022}. The Encoder condenses the original dimensionality of the input variables down to only 5 nodes inside the latent space. This enhances the interpretability of ED and makes the model beneficial for advanced postprocessing of multivariate climate data~\citep{Behrens2022, mooers2023comparing}.

\textbf{Heteroskedastic Regression (HSR)} predicts a separate mean and standard deviation for each output variable, using a regularized MLP \citep{wongtoi2023understanding}.

\textbf{Multi-layer Perceptron (MLP)} is a fully connected, feed-forward neural network. The MLP architecture used for our experiments is optimized via an extensive hyperparameter search with 8,257 trials.

\textbf{Randomized Prior Network (RPN)} is an ensemble model \citep{Osband2018}. Each member of the RPN is built as the sum of a trainable and a non-trainable (so-called ``prior'') surrogate model; we used MLP for simplicity. Multiple replicas of the networks are constructed by independent and random sampling of both trainable and non-trainable parameters \citep{Yang2022, Bhouri2023}. RPNs also resort to data bootstrapping (e.g., subsampling and randomization) in order to mitigate the uncertainty collapse of the ensemble method when tested beyond the training data points \citep{Bhouri2023}.
%Data bootstrapping consists of sub-sampling and randomization of the data each network in the ensemble sees during training. Hyperparameters of individual MLPs did not need to be tuned from scratch, and were instead chosen based on the hyperparameter search done for the MLP. RPN ensembles of 128 networks were considered as justified in \cite{Yang2022}.

\textbf{Conditional Variational Autoencoder (cVAE)} uses amortized variational inference to fit a deep generative model that is conditioned on the input and can produce samples from a complex predictive distribution.

\begin{table}[ht]
\centering
\small
\resizebox{\textwidth}{!}{\begin{tabular}{l|cccccc|cccccc}
\toprule
\multicolumn{1}{c|}{\multirow{2}{*}{\textbf{Variable}}} & \multicolumn{6}{c|}{\textbf{MAE [W/m$^\text{2}$]}} & \multicolumn{6}{c}{\textbf{R$^\text{2}$}} \\
\cmidrule{2-13}
\multicolumn{1}{c|}{} 
& CNN & ED & HSR & MLP & RPN & cVAE & CNN & ED & HSR & MLP & RPN & cVAE \\
\midrule
$dT/dt$ & \textbf{2.585} & 2.864 & 2.845 & 2.683 & 2.685 & 2.732 & \textbf{0.627} & 0.542 & 0.568 & 0.589 & 0.617 & 0.590 \\
$dq/dt$ & \textbf{4.401} & 4.673 & 4.784 & 4.495 & 4.592 & 4.680 & -- & -- & -- & -- & -- & -- \\
NETSW & 18.85 & 14.968 & 19.82 & \textbf{13.36} & 18.88 & 19.73 & 0.944 & 0.980 & 0.959 & \textbf{0.983} & 0.968 & 0.957  \\
FLWDS & 8.598 & 6.894 & 6.267 & \textbf{5.224} & 6.018 & 6.588 & 0.828 & 0.802 & 0.904 & \textbf{0.924} & 0.912 & 0.883 \\
PRECSC & 3.364 & 3.046 & 3.511 & \textbf{2.684} & 3.328 & 3.322 & -- & -- & -- & -- & -- & -- \\
PRECC & 37.83 & 37.250 & 42.38 & \textbf{34.33} & 37.46 & 38.81 & \textbf{0.077} & -17.909 & -68.35 & -38.69 & -67.94 & -0.926 \\
SOLS & 10.83 & 8.554 & 11.31 & \textbf{7.971} & 10.36 & 10.94 & 0.927 & 0.960 & 0.929 & \textbf{0.961} & 0.943 & 0.929 \\
SOLL & 13.15 & 10.924 & 13.60 & \textbf{10.30} & 12.96 & 13.46  & 0.916 & 0.945 & 0.916 & \textbf{0.948} & 0.928 & 0.915  \\
SOLSD & 5.817 & 5.075 & 6.331 & \textbf{4.533} & 5.846 & 6.159 & 0.927 & 0.951 & 0.923 & \textbf{0.956} & 0.940 & 0.921 \\
SOLLD & 5.679 & 5.136 & 6.215 & \textbf{4.806} & 5.702 & 6.066 & 0.813 & 0.857 & 0.797 & \textbf{0.866} & 0.837 & 0.796 \\
\bottomrule
\end{tabular}}
\vspace{1mm}
\setlength{\belowcaptionskip}{-0.5em}%
\caption{MAE and R$^\text{2}$ for target variables averaged globally and temporally (from 0009-03 to 0011-02). Variables include heating tendency ($dT/dt$), moistening tendency ($dq/dt$), net surface shortwave flux (NETSW), downward surface longwave flux (FLWDS), snow rate (PRECSC), rain rate (PRECC), visible direct solar flux (SOLS), near-IR direct solar flux (SOLL), visible diffused solar flux (SOLSD), and near-IR diffused solar flux (SOLLD). Units of non-energy flux variables are converted to a common energy unit, W/m$^\text{2}$. Best model performance for each variable is bolded. For $dq/dt$ and PRECSC, global mean R$^\text{2}$ is not an ideal evaluation metric and not reported due to negligible variability in $dq/dt$ in the upper atmosphere and PRECSC in the tropics in the dataset.}
\label{tab:summarystats}
\end{table}

% ===============================
\subsection{Offline Skill Boost from Expanding Features and Targets}
% ===============================
\label{subsection:baseline_expanded}
We performed an ablation of our best-performing MLP baseline to demonstrate the added value of the expanded inputs and targets available in \dataset{}, i.e., using inputs $x$ of size $d_i=617$ and targets $y\in\mathbb R^{d_o}$ of size $d_o=368$; see Table 1 in SI for the full list of variables. We use the same transformation described in our preprocessing workflow to compute and add condensate (cloud liquid and cloud ice) and momentum (zonal and meridional winds) tendencies to the target vector. We conducted this ablation study with both the low-resolution and the high-resolution datasets (see Section 3.1 in SI for further details regarding these MLP variants). For common elements of the target vector, using all available variables leads to a uniform improvement in prediction accuracy, especially for precipitation, in both resolutions (Figures SI7, SI8 and Table SI4). The larger errors (e.g., MAE and RMSE) observed in the high-resolution emulators are anticipated due to the increased variance of higher-resolution data. Nevertheless, the similarity of their R$^\text{2}$ values to those of the corresponding low-resolution emulators confirms their adequate performance.

% ===============================
\subsection{Evaluation Metrics}
% ===============================
\label{subsection:offline_metrics}
Our evaluation metrics are computed separately for each variable in the output vector. The mean absolute error (MAE) and the coefficient of determination (R$^\text{2}$) are calculated independently at each horizontal and vertical location and then averaged horizontally and vertically to produce the summary statistics in Figure~\ref{fig:allmetricsplot}. For the vertically-varying fields, we first form a mass-weighting and then convert moistening and heating tendencies into common energy units in Watts per square meter as in \cite{beucler2024climate}. We also report continuous ranked probability scores (CRPS) for all considered models in SI.

% ===============================
\subsection {Baseline Model Results}
% ===============================

\begin{figure}[ht]
  \centering 
  \includegraphics[width=1.\textwidth, trim={1.5cm 0 1.5cm 1.cm},clip]{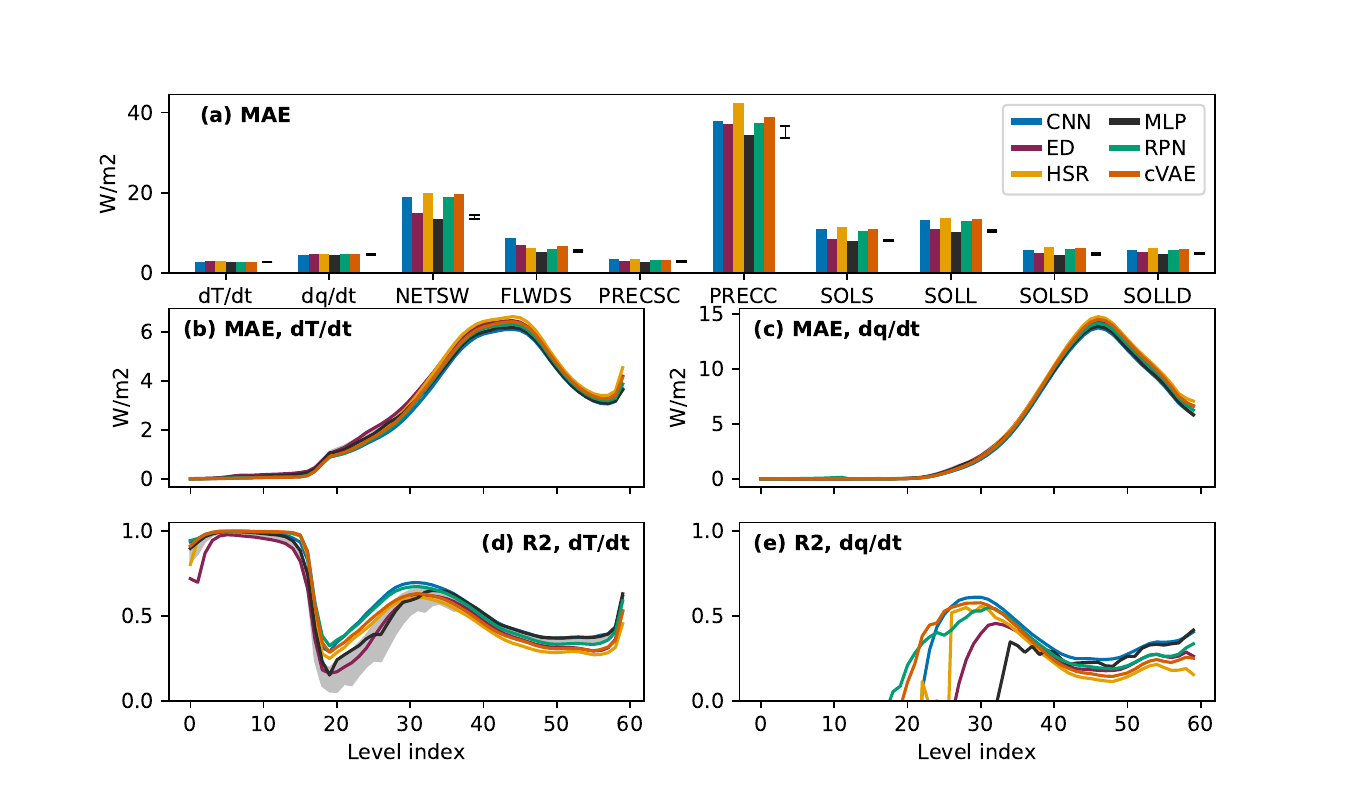}
  \setlength{\belowcaptionskip}{-0.25em}%
   \caption{(a) Summary, where $dT/dt$ and $dq/dt$ are the tendencies of temperature and specific humidity, respectively, and were vertically integrated with mass weighting. (b,c) retain the vertical structure of MAE and (d,e) R$^\text{2}$. Error bars and grey shadings show the the 5- to 95-percentile range of MLP. Refer to Table \ref{tab:baselinevars} for variable definitions.}
  \label{fig:allmetricsplot}
\end{figure}

Figure \ref{fig:allmetricsplot} summarizes the error characteristics. Whereas heating and moistening rates have comparable global mean MAE, behind a common background vertical structure (Figure \ref{fig:allmetricsplot} b,c) the coefficient of determination R$^\text{2}$ (d,e) reveals that certain architectures (RPN, HSR, cVAE, CNN) consistently perform better in the upper atmosphere (model level < 30) whereas the highly optimized MLP model outperforms in the lower atmosphere (model level > 30) and therefore the global mean (Table \ref{tab:summarystats}). For the global mean MAE, we see the largest averaged errors for PRECC and NETSW (mean MAE > 15 W/m², Figure \ref{fig:allmetricsplot} and Table \ref{tab:summarystats}), where MLP clearly has the best skill compared to all other benchmark models. For the other variables, the global mean MAE is considerably smaller, and the skill of the benchmarks model appears to be more similar in absolute numbers. While for the global mean R$^\text{2}$ we find the lowest measurable performance for dT/dt and PRECC (mean R$^\text{2} <$ 0.7) and in these cases, CNN gives the most skillful predictions. The other variables have larger R$^\text{2}$ of order 0.8 or higher, which suggests that these quantities are easier to deep-learn (Table \ref{tab:summarystats}). For dq/dt and PRECSC global mean R$^\text{2}$ is not an ideal evaluation metric due to negligible variability in dq/dt in the upper atmosphere and for PRECSC in the tropics in the dataset (Table \ref{tab:summarystats}). 

Additional tables and figures that reveal the geographic and vertical structure of these errors, fit quality, and analysis of stochastic metrics are included in SI (Sections 4.3, 8.1, and 8.2 in SI).

% ===============================
\subsection{Physics-Informed Guidance to Improve Generalizability and Online Performance}
\label{subsec:physguidance}
% ===============================

\textbf{Physical Constraints:} Mass and energy conservation are important criteria for Earth system simulation. If these terms are not conserved, errors in estimating sea level rise or temperature change over time may become as large as the signals we hope to measure. Enforcing conservation on the learned quantities helps constrain results to be physically plausible and reduce the potential for errors accumulating over long time scales. In addition to conservation laws, we demonstrate in Section \ref{subsec:online_results} that implementing cloud physics constraints can optimize online error and improve the stability of hybrid simulations. We discuss how to implement a range of conservation and cloud physics constraints, as well as enforce additional constraints, such as non-negativity for precipitation, condensate, and moisture variables, in the SI. 

\textbf{Stochasticity and Memory:} The results of the embedded convection calculations regulating $d_o$ are chaotic and thus worthy of stochastic architectures, as in our RPN, HSR, and cVAE baselines. These solutions are likewise sensitive to sub-grid initial state variables from an interior nested spatial dimension that has not been included in our data.  

\textbf{Temporal Locality:} Incorporating the previous timesteps' target or feature in the input vector inflation could be beneficial as it captures some information about this convective memory and utilizes temporal autocorrelations present in atmospheric data. This approach has been explored in previous studies \citep{Han2020,Wang2022a,han2023ensemble,behrens2024improving} and has been integrated into our model for online testing (see Section \ref{sec:online} in the main text and Section 6.3.3 in the SI).

\textbf{Causal Pruning:} A systematic and quantitative pruning of the input vector based on objectively assessed causal relationships to subsets of the target vector has been proposed as an attractive preprocessing strategy, as it helps remove spurious correlations due to confounding variables and optimize the ML algorithm \citep{Iglesias2023,kuhbacher2024towards}.

\textbf{Normalization:} Normalization that goes beyond removing vertical structure could be strategic, such as removing the geographic mean (e.g., latitudinal, land/sea structure) or composite seasonal variances (e.g., local smoothed annual cycle) present in the data. For variables exhibiting exponential variation and approaching zero at the highest level (e.g., metrics of moisture), log-normalization might be beneficial. 

\textbf{Expanded Resolution and Complete Inputs and Outputs:} Our baseline models have focused on the low-resolution dataset, for ease of data volume, and using only a subset of the available inputs and outputs. This illustrates the essence of the ML challenge. However, we show in our ablation study, using MLPs, that including all input variables generally yields an improved reproduction of the target variables in both the low-resolution and the high-resolution datasets (Figures SI7 and SI8 and Table SI4). Accordingly, we encourage users who discover competitive fits in this approachable limit to expand to all inputs/outputs in the high-resolution, real-geography dataset, for which successful fits become operationally relevant.

\textbf{Further ML Approaches:} Recent methods to capture multi-scale processes using neural operators that learn in a discretization-invariant manner and can predict at higher resolutions than available during training time \citep{Li2021} may be attractive. Their performance can be further enhanced by incorporating physics-informed losses at a higher resolution than available training data \citep{Li2023}. Ideas on ML modeling for sub-grid closures from adjacent fields like turbulent flow physics and reactive flows can also be leveraged for developing architectures with an inductive bias for known priors \citep{Ling2016}, easing prediction of stiff non-linear behavior \citep{MacArt2021, Xing2021, Brenner2019}, generative modeling with physical constraints \citep{Subramaniam2020, Kim2019} and for interpretability of the final trained models \citep{MacArt2021}.

% ======================================================
\section{Hybrid Testing and Online Performance Evaluation}
\label{sec:online}
% ======================================================

The primary objective of evaluating a machine learning model within hybrid climate simulations is to measure the online error \citep{Rasp2018,Wang2022a,kochkov2023neural,sanford2023improving}. This error assesses how well the hybrid simulation, which integrates the ML model with the rest of the climate simulator, reproduces the statistics of the original high-fidelity climate simulation. Optimizing offline metrics does not necessarily lead to optimized online performance \citep{ott2020,Wang2022a}, as small prediction errors at each time step can accumulate over year-long climate simulations (26,280 timesteps per simulated year). In this section, we describe how to integrate the ML model into the climate model and the process for evaluating the online performance of the hybrid ML-physics climate simulations. We also provide a case study from \cite{Hu2024} illustrating our experience in improving the online performance of the hybrid simulation. In this online task, the ML models predict the expanded targets $y \in \mathbb{R}^{d_o}$ of size $d_o=368$, similar to baseline models with expanded features and target (Section \ref{subsection:baseline_expanded}) (tendencies of temperature, moisture, cloud water, cloud ice, zonal wind and meridional wind, in addition to precipitation and radiative fluxes at the surface, see Table 1 in the SI). These expanded targets ensure that the ML model predicts all the necessary variables to update the atmospheric state and drive the rest of the climate simulator, thereby enabling complete coupling.

% ===============================
\subsection{Software to Integrate ML Models into Physical Climate Simulations}
\label{subsec:online_workflow}
% ===============================

\textbf{Pytorch-Fortran Coupling:} The original climate model, the E3SM-MMF multi-scale climate model (see Section 1.1 in SI for more details), is written in object-oriented, MPI-decomposed Fortran. To integrate a Python-based ML model into the climate model and replace the learned code subregion, we implemented a coupling workflow using an open-source library called Pytorch-Fortran \citep{dmitry_alexeev_2023_7851167}. This library simplifies the integration of PyTorch models with Fortran-based climate simulators and specifically supports TorchScript models. It provides straightforward interfaces for loading ML models, processing Fortran tensors in a zero-copy fashion, and performing efficient batch inference.

\textbf{TorchScript:} To utilize the PyTorch-Fortran bindings, it is necessary to first serialize PyTorch models using TorchScript \citep{torchscript}. TorchScript models can operate independently of Python and support flexible architecture design. Converting a PyTorch model into TorchScript is straightforward, as TorchScript is compatible with most PyTorch functions and many Python built-ins. Section 6.1 in the SI provides detailed instructions on writing PyTorch models that can be converted into TorchScript. We also include example code for converting a PyTorch model into TorchScript. Although we use the TorchScript interface in PyTorch-Fortran for best performance, PyTorch-Fortran also supports spawning a Python interpreter and run any Python code using other ML packages (e.g., JAX, scikit-learn, and TensorFlow), provided the input and output interface of the Python code is \texttt{torch.Tensor}.

\textbf{Cross-Platform Containerized Hybrid Testing Workflow: } Building flexible couplers to foster the progress from skillful ML parameterizations into skillful hybrid ML-Physics climate simulators is vital for the ML and climate community. Unfortunately, the complexity and nuance involved in performing climate simulations has meant many wasted graduate student hours and a dearth of online error results in the hybrid simulation literature \citep{lin2023systematic}.  We have implemented the first end-to-end containerized workflow for this purpose, enabling ML models to be integrated into our climate simulator for online testing. This container can be deployed on multiple platforms, including Linux-based laptops or workstations, high-performance computing (HPC) clusters, and cloud-based virtual machines. Once the container is set up, users can easily launch hybrid simulations by providing the trained ML model in TorchScript format and reproduce the results shown in Figure \ref{fig:online-ablation}.

% ===============================
\subsection{Metrics for Evaluating Online Errors in Hybrid Climate Simulations}
\label{subsec:online_metric}
% ===============================

\textbf{Root Mean Square Error:} Our online evaluation metrics are computed separately for each variable in the hybrid simulations. For a given month, the root mean square error (RMSE) for each variable is calculated as follows:

\begin{equation}
\text{RMSE} = \sqrt{\sum\limits_{i=1}^{S_m} w_i (\hat{y}_m - y_m)^2}
\end{equation}

where:
\begin{itemize}
  \item $S_m$ is the number of samples (each horizontal grid cell at a given time is one sample) across the entire globe,
  \item $\hat{y}_m$ represents the values from the hybrid simulation averaged over the entire month,
  \item $y_m$ represents the values from the reference simulation averaged over the entire month,
  \item $w_1, w_2, \ldots, w_{S_m}$ are weights that sum to 1 and are proportional to the air mass in each grid cell.
\end{itemize}

\textbf{Zonal Mean Bias:} Additionally, we evaluate the multi-year (26,280 timesteps per year) zonal mean bias, which measures the average difference between the hybrid simulation and the reference simulation across key atmospheric variables, such as temperature, moisture, wind, and cloud liquid and ice water. The zonal mean bias is derived by comparing variables averaged over time and longitudes. For a more detailed illustration of the zonal mean bias results, please refer to Section 6.4 in the SI. 

% ===============================
\subsection{Experiment Setup for Hybrid Online Testing}
\label{subsec:online_experiment}
% ===============================

The initial architecture we used for testing the online performance was the MLP model described in Section \ref{sec:experiments}. However, as we will show in the next section, the hybrid simulations with the MLP model were unstable and exhibited large online errors. The online failure of the MLP model motivated us to explore more expressive architectures to achieve stable hybrid simulations. \cite{Hu2024} explored a U-Net architecture, which is effective at capturing the atmospheric vertical structure and is also recommended by \cite{heuer2023interpretable}. This U-Net architecture achieved stable hybrid simulations with satisfying online performance (see Section \ref{subsec:online_results} and see Section 6.4 in the SI).

Here, we use the experiments and results from \cite{Hu2024} to illustrate the online evaluation process and highlight experiences and key factors for optimizing online performance. Three architecture designs are evaluated: a baseline MLP architecture, and two U-Net architectures with expanded input features that include information from previous time steps. One of the U-Net models further incorporates additional cloud physics constraints to enhance performance. It is worth noting that these MLP and U-Net models are trained with different input/output normalization strategies than those used in the baseline models in Section \ref{sec:experiments} \cite{Hu2024}.

To account for the variability in online performance not fully captured by offline skill, we tested three different checkpoints for each model. These checkpoints were obtained using varying loss functions and learning rate schedules, as different configurations were found to lead to differing online stability and error, even with similar offline performance \citep{lin2023systematic}.

Each checkpoint was used to run a one-year hybrid simulation with the same initial conditions. We evaluated the monthly RMSE evolution throughout the year compared to the reference E3SM-MMF simulation. Additionally, to estimate the inherent unpredictability of the atmospheric system, we ran three additional pure physical simulations with the same initial conditions. These pure physical simulations were implemented using parallel reductions and atomic operations that prevent bitwise reproducibility (see Section 6.3.5 in the SI). The accumulation of these rounding errors over time can lead to variations in the climate model outcomes, mimicking the chaotic nature of the atmosphere. These simulations serve as a baseline for the atmospheric unpredictability.

For more details on model architectures, hyperparameters, input variables, and cloud physics constraints, please refer to Section 6.3 in the SI.

% ===============================
\subsection{Results of Online Performance Testing}
\label{subsec:online_results}
% ===============================

\begin{figure}[ht]
  \centering 
  \includegraphics[width=1.\textwidth]{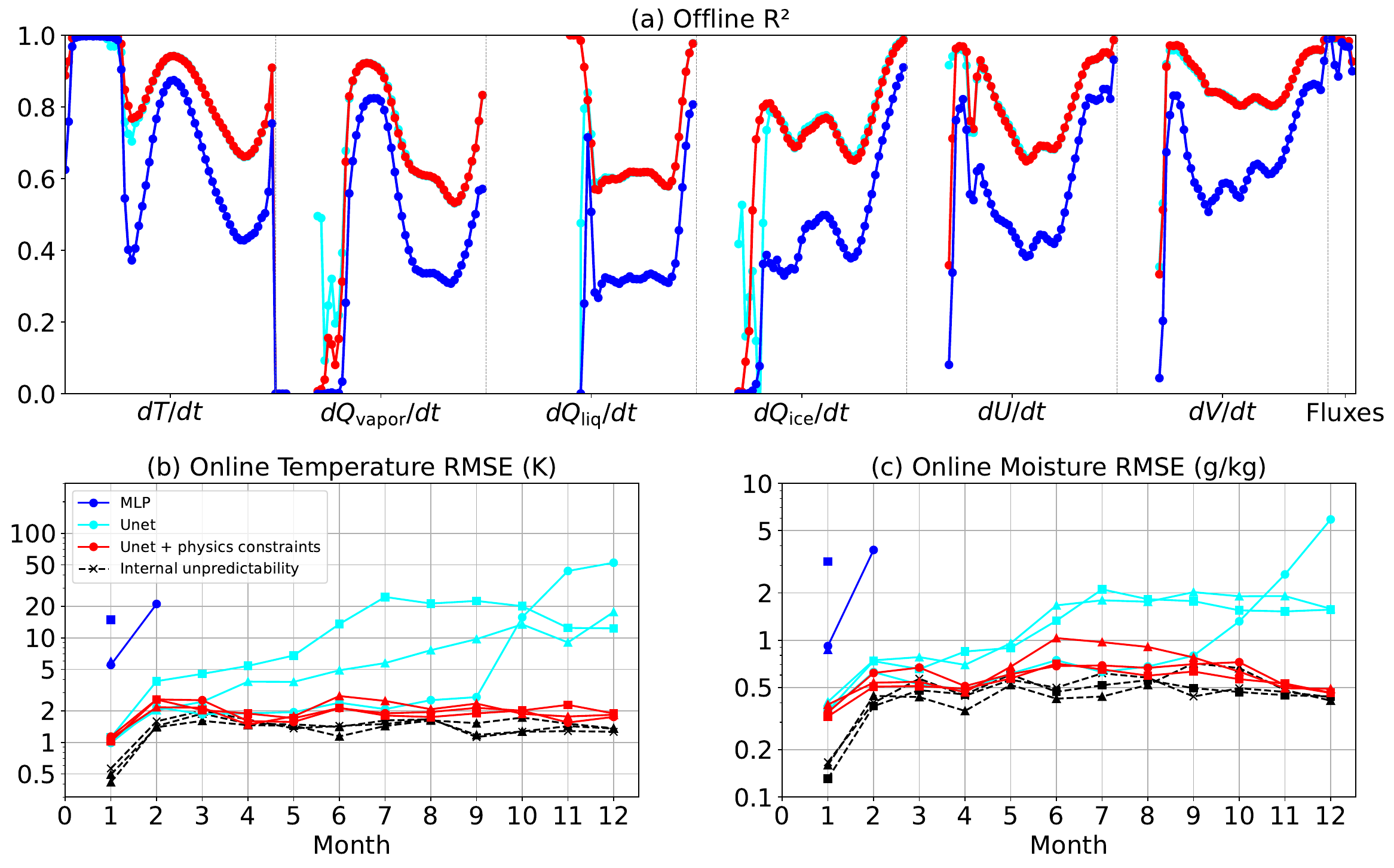}
  \setlength{\belowcaptionskip}{-0.25em}%
   \caption{(a) Offline R$^2$ scores across various variables for MLP, U-Net, and U-Net with physics constraints. Variables are the full target variables listed in Table S1, including temperature tendency ($\frac{dT}{dt}$), water vapor tendency ($\frac{dQ_{\text{v}}}{dt}$), liquid cloud mixing ratio tendency ($\frac{dQ_{\text{c}}}{dt}$), ice cloud mixing ratio tendency ($\frac{dQ_{\text{i}}}{dt}$), zonal wind tendency ($\frac{dU}{dt}$), meridional wind tendency ($\frac{dV}{dt}$), and eight flux variables. (b,c) Online monthly and globally averaged (both horizontally and vertically and weighted by mass in each grid) RMSE of temperature (K) and moisture (g/kg) over a one-year period, comparing baseline MLP, U-Net, and U-Net with physics constraints models against the reference E3SM-MMF simulation. Atmospheric unpredictability (black dashed lines) is estimated by running the reference E3SM-MMF simulations multiple times with the same initial condition while allowing for the chaotic growth of the random rounding errors.
}
  \label{fig:online-ablation}
\end{figure}

\textbf{Summary of Online Error and Offline Skill:} Figure \ref{fig:online-ablation} summarizes the online error for the one-year hybrid simulations along with the offline skill changes across our model choices. The hybrid simulations with the baseline MLP models were unstable, with all three instances crashing within the first two months of the simulations. In contrast, the more expressive U-Net architecture, which includes additional inputs from previous time steps, improved the R$^\text{2}$ score across all vertical levels and variables. In our case, this enhanced offline skill also translated to better online performance. The U-Net architecture allowed for more stable hybrid simulations, with all the U-Net simulations completing the full year. During the first month before the MLP simulations crashed, these U-Net simulations demonstrated significantly lower first-month RMSE for climatological temperature and moisture compared to the hybrid simulations that used the MLP models without the additional previous-time inputs.

\textbf{Impact of Cloud Physics Constraints:} Incorporating cloud physics constraints significantly improved stability and reduced error growth in the hybrid simulations. Without these constraints, the U-Net models developed increasing errors after a few months, leading to unrealistic cloud formations not represented in the training data, which potentially contributed to higher error growth (see \cite{Hu2024} for more details). The cloud physics constraints mitigated this issue by ensuring clouds formed appropriately relative to temperature, such as preventing liquid clouds at very low temperatures where condensate should be frozen. 

\textbf{Achieving State of the Art for Hybrid Error:}
With these constraints, the online RMSE stabilized after an initial rise within the first two months. The global temperature RMSE remained around 2K, and the global moisture RMSE stayed below 1 g/kg, during the entire span of the one-year simulations (red lines in Figure \ref{fig:online-ablation}). In the SI, we show that the U-Net model with cloud physics constraints can integrate stably for at least five years, maintaining a tropospheric zonal mean temperature bias below 2K and a moisture bias below 1 g/kg. While there is still room for improvement, these magnitudes outperform the state of the art results of \citep{han2023ensemble} within the context of analogous multi-scale hybrid climate simulations, despite our having included the full complexity of interactive condensate coupling and inclusion of radiative transfer within the full-physics ML parameterization, which was sidestepped in that work. A caveat in this comparison is that the work of \citep{han2023ensemble} used a higher resolution host climate model that has higher intrinsic variance, as well as a different software version for the multi-scale climate simulator. 

Beyond the context of the multiscale modeling framework, other state-of-the-art hybrid ML-physics climate simulators include those by \cite{kochkov2023neural} and \cite{sanford2023improving} demonstrate additional metrics of hybrid model performance. For example,\cite{sanford2023improving} reported a pattern RMSE of 1.2K for the annual mean temperature climatology at 200 hPa and 850 hPa and 2 mm for precipitable water compared to their reference dataset using a coarse-grained high-resolution global storm-resolving model. \cite{sanford2023improving} also showed an annual-mean zonal mean temperature bias below 1K over most of the troposphere and a zonal mean moisture bias within ~0.8 g/kg. \cite{kochkov2023neural} reported an annual-mean temperature RMSE of 0.61K at 850 hPa and 1.05K at 200 hPa, and an RMSE of 1.09 mm for annual precipitable water against their reference ERA5 dataset. It is also worth noting that such error metrics tend to be a function of climate model resolution and thus cannot be directly compared between these studies; the resolution in our hybrid model is approximately 10$^\circ$, which is much lower than the 2$^\circ$ used by \cite{sanford2023improving} and the 1.4$^\circ$ used by \cite{kochkov2023neural}.  

Further optimizing the remaining non-differentiable online bias is challenging and worthy of community effort. We discuss potential methods to optimize these online bias in Section 6.5 of the SI. 

% ======================================================
\section{Limitations and Other Applications}
\label{sec:limitations}
% ======================================================

\textbf{Idealizations:} A limitation of the multi-scale climate simulator used to produce \dataset{}  (E3SM-MMF) is that it assumes scale separation, i.e., that convection can be represented as laterally periodic within the grid size of the host simulator, and neglects sub-grid scale representations of topographic and land-surface variability. The configuration of the multi-scale climate simulator used to make \dataset{} also has no atmosphere-ocean coupling and ignores the radiative effects of aerosols. These are essential for simulating important climate phenomena like El Niño and the influence of aerosols on cloud properties, which are critical for realistic future climate projections. Despite these simplifications, the data adequately capture many historically challenging aspects of the ML parameterization problem, such as stochasticity, and complex nonlinear interactions across radiation, microphysics, and turbulence.

\textbf{Hybrid testing:} To maximize simplicity and scalability, our containerized pipeline for evaluating hybrid error uses a very low resolution hybrid climate simulation that can run on just a single cloud compute node and even a personal laptop, and our evaluation protocol contains only minimum viable integrative statistics of hybrid climate simulation errors. As methods to reliably achieve hybrid skill mature, this pipeline should be expanded. First, in its computational ambition, towards multi-node cloud-compatible configurations compatible with a full-resolution hybrid simulation. Second, to more fully evaluate a resulting historical simulation or a simulation under Atmospheric Modelling Intercomparison Project conditions \citep[AMIP,][]{gates1999overview} with a hybrid model with Earth observations. Testing large ensembles launched from multiple initial conditions as in \cite{kochkov2023neural} would be beneficial. Implementing diagnostics of the tendencies predicted by the ML versus physics model, while each is alternately coupled to the host dynamics, would be strategic as was found useful for making progress on microphysics ML parameterization in \citet{perkins2024emulation}. Finally, as hybrid simulators stabilize and begin to produce reasonable time-mean climate statistics, their variability behind the mean state becomes important to validate, such as by measuring intrinsic cyclogenesis frequency \citep{kochkov2023neural}. Community-developed open-source diagnostic tools such as the Earth System Model Evaluation Tool \citep{Eyring_2020} and the Model Diagnostics Task Force Framework \citep{neelin2023process} facilitate the  evaluation of climate simulations compared to observations and traditional climate models.

\textbf{Stochasticity:} One open problem that the dataset may allow assessing is understanding the role of stochasticity in hybrid-ML simulation. While primarily used as a dataset for regression it would be also interesting to assess and understand the degree to which different variables are better modeled as stochastic or deterministic, or if the dataset gives rise to heavy-tailed or even multi-modal conditional distributions that are important to capture. To date, these questions have been raised based on physical conjectures \citep[e.g.,][]{Lin2003} but remain to be addressed in the ML-based parameterization literature. For instance, precipitation distributions have long tails that are projected to lengthen under global climate change \citep{O'Gorman2015, Neelin2022}---and will thus tend to generate out-of-distribution extremes. \dataset{} could help construct optimal architectures to capture precipitation tails and other impactful climate variables such as surface temperature, and could be easily extended to a distributional regression benchmark. 

\textbf{Interpretability:} This dataset could also be utilized to discover physically interpretable models for atmospheric convection, radiation, boundary layer turbulence, and microphysics. A possible workflow would apply dimensionality reduction techniques to identify dominant predictors and vertical variations, followed by symbolic regression to recover analytic expressions \citep{Zanna2020, Grundner2023}.

\textbf{Generalizability:} Although the impacts of global climate change and inter-annual variability are absent in this initial version of \dataset{}, important questions surrounding climate-convection interactions can begin to be addressed. One strategy would involve partitioning the data such that the emulator is trained on cold columns, but validated on warm columns, where warmth could be measured by surface temperatures, as in \cite{beucler2024climate}. However, the results from this approach may also reflect the dependence of convection on the geographical distribution of surface temperatures in the current climate and should be interpreted with caution. To optimally engage ML researchers in solving the climate generalization problem, a multi-climate extension of \dataset{} should be developed that includes physical simulations that samples future climate states and more internal variability. 

%\textbf{Relevance determination and active learning:} While the climate simulator code offers data generation flexibility, guidance on ideal regimes to target for improved learning would benefit the domain scientists able to run it. This question can be addressed with the current data and metrics of interest provided. 

% ======================================================
\section{Conclusion and Future Work}
\label{sec:conclusion}
% ======================================================

We introduce \datasetonline{}, the most physically comprehensive dataset and framework yet published for training and testing ML-based parameterizations of atmospheric storms, clouds, turbulence, rainfall, and radiation for use in hybrid-ML climate simulations. It contains all inputs and outputs necessary for online coupling in a full-complexity multi-scale climate model. Additionally, it provides a containerized pipeline to integrate ML models into climate models, allowing for the evaluation of online performance in hybrid-ML climate model simulations. This containerized approach ensures reproducibility and accessibility, making it user-friendly for ML researchers without domain expertise and, unlike typical climate simulations, compatible with commonly available cloud and local computing environments. We conduct a series of experiments on a subset of these variables that demonstrate the degree to which climate data scientists have been able to fit the deterministic and stochastic components in the dataset. We also provide a hybrid-ML baseline model to showcase one example of improving the hybrid stability and online error, along with initial metrics for assessing it. This demonstrates how \datasetonline{} can be an operational pipeline to explore capabilities of novel models from the ML community in climate science.

We hope ML community engagement in \dataset{} will advance fundamental ML methodology and clarify the path to producing increasingly skillful subgrid-scale physics parameterizations that can be reliably used for operational climate simulation \citep{eyring24_hybridmodelling}. To facilitate two-way communications between ML practitioners and climate scientists, we incorporate many desired characteristics for an ideal benchmark dataset suggested in \cite{ebert2017vision,dueben2022challenges}. 
Such interdisciplinary collaboration will open up an exciting future in which the computational limits that currently constrain climate simulation can be reconsidered. We are already encouraged by several thousand global participants in a Kaggle ML competition based on the \dataset{} dataset that has attracted users from diverse domains and fostered innovation in the offline problem \cite{leap-atmospheric-physics-ai-climsim}.    
 
We plan to soon extend the \dataset{} dataset to include a sampling of multiple future climate states \cite{Clark2022-sr, Bhouri2023-pe}. We hope the lessons learned from our focus on multi-scale atmospheric simulations will apply to other sub-fields of Earth System Science, where computational constraints currently hinder explicit representations of more complex systems.

% \blindmathpaper

%  Here is a citation \cite{chow:68}.

% Acknowledgements and Disclosure of Funding should go at the end, before appendices and references

\acks{
This work is broadly supported across countries and agencies. Primary support is by NVIDIA, the National Science Foundation (NSF) Science and Technology Center (STC) Learning the Earth with Artificial Intelligence and Physics (LEAP; Award \# 2019625-STC), and the Exascale Computing Project (17-SC-20-SC), a collaborative effort of the U.S. Department of Energy (DOE) Office of Science (SC), the National Nuclear Security Administration, and the Energy Exascale Earth System Model project, funded by DOE grant DE-SC0022331. M.S.P, S.Y., L.P., A.M.J., J.L., N.L., and G.M. further acknowledge support from the DOE (DE-SC0023368) and NSF (AGS-1912134). R.Y, S.M, P.G, M.P. acknowledge funding from the DOE Advanced Scientific Computing Research (ASCR) program (DE-SC0022255). C.B. acknowledges the Paul G. Allen Family Foundation's funding of AI2. V.E., P.G., H.H., G.B., and F.I.-S. acknowledge funding from the European Research Council Synergy Grant (Agreement No. 855187) under the Horizon 2020 Research and Innovation Programme. E.A.B. was supported, in part, by NSF grant AGS-2210068. S.J. acknowledges funding from DOE ASRC under an Amalie Emmy Noether Fellowship Award in Applied Mathematics (B\&R \#KJ0401010). M.A.B acknowledges NSF funding from an AGS-PRF Fellowship Award (AGS-2218197). R.G. acknowledges funding from the NSF (DGE-2125913) and the U.S. Department of Defense (DOD). S.M. acknowledges support from an NSF CAREER Award and NSF grant IIS-2007719. P.-L. M. acknowledges support from the Enabling Aerosol–cloud interactions at GLobal convection-permitting scalES (EAGLES) project (No. 74358) sponsored by DOE SC, Office of Biological and Environmental Research (BER), Earth System Model Development (ESMD) program area. 
%SF 
L.Z. and N.L. received M$^2$LInES research funding by the generosity of Eric and Wendy Schmidt by recommendation of the Schmidt Futures program. 
% NERSC
This research used resources of the National Energy Research Scientific Computing Center (NERSC), a DOE SC User Facility operated under Contract No. DE-AC02-05CH11231.
% PNNL
The Pacific Northwest National Laboratory is operated by Battelle for the DOE under Contract DE-AC05-76RL01830.
% LLNL
This work was performed under the auspices of the DOE by Lawrence Livermore National Laboratory under Contract DE-AC52-07NA27344.
% Bridges2
This work used Bridges\-2 at the Pittsburgh Supercomputing Center through allocation ATM190002 from the Advanced Cyberinfrastructure Coordination Ecosystem: Services \& Support (ACCESS) program, which is supported by NSF grants \#2138259, \#2138286, \#2138307, \#2137603, and \#2138296.
This work also utilized the DOD High Performance Computing Modernization Program (HPCMP).
% Ack authors from the earlier manuscript:
We recognize the contributions of Anima Anandkumar, David C Bader, Peter Caldwell, Nicholas Geneva, Yilun Han, Karthik Kashinath, Marat Khairoutdinov, Thorsten Kurth, Griffin Mooers, Jaideep Pathak, David Randall, Mark A Taylor, and Carl Vondrick  to the earlier work that formed the basis of this updated manuscript.
}

% SY, 7/8/2024: turn off 'newpage' for preprint submission. (There was awkward white space due to the "preprint" option.)
% % Manual newpage inserted to improve layout of sample file - not
% % needed in general before appendices/bibliography.

% \newpage

% \vskip 0.2in
\bibliography{refs}

\begin{thebibliography}{107}
\providecommand{\natexlab}[1]{#1}
\providecommand{\url}[1]{\texttt{#1}}
\expandafter\ifx\csname urlstyle\endcsname\relax
  \providecommand{\doi}[1]{doi: #1}\else
  \providecommand{\doi}{doi: \begingroup \urlstyle{rm}\Url}\fi

\bibitem[Alexeev(2023)]{dmitry_alexeev_2023_7851167}
Dmitry Alexeev.
\newblock alexeedm/pytorch-fortran: Version v0.4, April 2023.
\newblock URL \url{https://doi.org/10.5281/zenodo.7851167}.

\bibitem[Behrens et~al.(2022)Behrens, Beucler, Gentine, Iglesias-Suarez, Pritchard, and Eyring]{Behrens2022}
Gunnar Behrens, Tom Beucler, Pierre Gentine, Fernando Iglesias-Suarez, Michael Pritchard, and Veronika Eyring.
\newblock Non-linear dimensionality reduction with a variational encoder decoder to understand convective processes in climate models.
\newblock \emph{J. Adv. Model. Earth Syst.}, 14\penalty0 (8):\penalty0 e2022MS003130, 2022.

\bibitem[Behrens et~al.(2024)Behrens, Beucler, Iglesias-Suarez, Yu, Gentine, Pritchard, Schwabe, and Eyring]{behrens2024improving}
Gunnar Behrens, Tom Beucler, Fernando Iglesias-Suarez, Sungduk Yu, Pierre Gentine, Michael Pritchard, Mierk Schwabe, and Veronika Eyring.
\newblock Improving atmospheric processes in earth system models with deep learning ensembles and stochastic parameterizations.
\newblock \emph{arXiv preprint arXiv:2402.03079}, 2024.

\bibitem[Benedict and Randall(2009)]{Benedict2009}
James~J Benedict and David~A Randall.
\newblock Structure of the madden--julian oscillation in the superparameterized cam.
\newblock \emph{J. Atmos. Sci.}, 66\penalty0 (11):\penalty0 3277--3296, 2009.

\bibitem[Beucler et~al.(2021)Beucler, Pritchard, Rasp, Ott, Baldi, and Gentine]{Beucler2021a}
Tom Beucler, Michael Pritchard, Stephan Rasp, Jordan Ott, Pierre Baldi, and Pierre Gentine.
\newblock Enforcing analytic constraints in neural networks emulating physical systems.
\newblock \emph{Phys. Rev. Lett.}, 126\penalty0 (9):\penalty0 098302, 2021.

\bibitem[Beucler et~al.(2024)Beucler, Gentine, Yuval, Gupta, Peng, Lin, Yu, Rasp, Ahmed, O’Gorman, et~al.]{beucler2024climate}
Tom Beucler, Pierre Gentine, Janni Yuval, Ankitesh Gupta, Liran Peng, Jerry Lin, Sungduk Yu, Stephan Rasp, Fiaz Ahmed, Paul~A O’Gorman, et~al.
\newblock Climate-invariant machine learning.
\newblock \emph{Science Advances}, 10\penalty0 (6):\penalty0 eadj7250, 2024.

\bibitem[Bhouri et~al.(2023{\natexlab{a}})Bhouri, Joly, Yu, Sarkar, and Perdikaris]{Bhouri2023}
Mohamed~Aziz Bhouri, Michael Joly, Robert Yu, Soumalya Sarkar, and Paris Perdikaris.
\newblock Scalable bayesian optimization with high-dimensional outputs using randomized prior networks, 2023{\natexlab{a}}.
\newblock arxiv:2302.07260.

\bibitem[Bhouri et~al.(2023{\natexlab{b}})Bhouri, Peng, Pritchard, and Gentine]{Bhouri2023-pe}
Mohamed~Aziz Bhouri, Liran Peng, Michael~S Pritchard, and Pierre Gentine.
\newblock Multi-fidelity climate model parameterization for better generalization and extrapolation.
\newblock \emph{arXiv.org}, 2023{\natexlab{b}}.

\bibitem[Bonev et~al.(2023)Bonev, Kurth, Hundt, Pathak, Baust, Kashinath, and Anandkumar]{Bonev2023}
Boris Bonev, Thorsten Kurth, Christian Hundt, Jaideep Pathak, Maximilian Baust, Karthik Kashinath, and Anima Anandkumar.
\newblock Spherical fourier neural operators: Learning stable dynamics on the sphere.
\newblock In \emph{Proc. ICLR}, 2023.

\bibitem[Brenner et~al.(2019)Brenner, Eldredge, and Freund]{Brenner2019}
M.~P. Brenner, J.~D. Eldredge, and J.~B. Freund.
\newblock Perspective on machine learning for advancing fluid mechanics.
\newblock \emph{Phys. Rev. Fluids}, 4:\penalty0 100501, 2019.

\bibitem[Brenowitz et~al.(2020)Brenowitz, Beucler, Pritchard, and Bretherton]{Brenowitz2020}
Noah~D Brenowitz, Tom Beucler, Michael Pritchard, and Christopher~S Bretherton.
\newblock Interpreting and stabilizing machine-learning parameterizations of convection.
\newblock \emph{J. Atmos. Sci.}, 77\penalty0 (12):\penalty0 4357--4375, 2020.

\bibitem[Bretherton et~al.(2022)Bretherton, Henn, Kwa, Brenowitz, Watt-Meyer, McGibbon, Perkins, Clark, and Harris]{Bretherton2022}
Christopher~S Bretherton, Brian Henn, Anna Kwa, Noah~D Brenowitz, Oliver Watt-Meyer, Jeremy McGibbon, W~Andre Perkins, Spencer~K Clark, and Lucas Harris.
\newblock Correcting coarse-grid weather and climate models by machine learning from global storm-resolving simulations.
\newblock \emph{J. Adv. Model. Earth Syst.}, 14\penalty0 (2):\penalty0 e2021MS002794, 2022.

\bibitem[Cachay et~al.(2021)Cachay, Ramesh, Cole, Barker, and Rolnick]{Cachay2021}
Salva~Rühling Cachay, Venkatesh Ramesh, Jason N.~S. Cole, Howard Barker, and David Rolnick.
\newblock Climart: A benchmark dataset for emulating atmospheric radiative transfer in weather and climate models, 2021.
\newblock arxiv:2111.14671.

\bibitem[Clark et~al.(2022{\natexlab{a}})Clark, Brenowitz, Henn, Kwa, McGibbon, Perkins, Watt-Meyer, Bretherton, and Harris]{Clark2022}
Spencer~K Clark, Noah~D Brenowitz, Brian Henn, Anna Kwa, Jeremy McGibbon, W~Andre Perkins, Oliver Watt-Meyer, Christopher~S Bretherton, and Lucas~M Harris.
\newblock Correcting a 200 km resolution climate model in multiple climates by machine learning from 25 km resolution simulations.
\newblock \emph{Journal of Advances in Modeling Earth Systems}, 14\penalty0 (9):\penalty0 e2022MS003219, 2022{\natexlab{a}}.

\bibitem[Clark et~al.(2022{\natexlab{b}})Clark, Brenowitz, Henn, Kwa, McGibbon, Perkins, Watt-Meyer, Bretherton, and Harris]{Clark2022-sr}
Spencer~K Clark, Noah~D Brenowitz, Brian Henn, Anna Kwa, Jeremy McGibbon, W~Andre Perkins, Oliver Watt-Meyer, Christopher~S Bretherton, and Lucas~M Harris.
\newblock Correcting a 200 km resolution climate model in multiple climates by machine learning from 25 km resolution simulations.
\newblock \emph{J. Adv. Model. Earth Syst.}, 14\penalty0 (9), September 2022{\natexlab{b}}.

\bibitem[Contributors(2024)]{torchscript}
PyTorch Contributors.
\newblock Torchscript documentation [software], 2024.
\newblock URL \url{https://pytorch.org/docs/stable/jit.html#pytorch-functions-and-modules}.

\bibitem[Davenport et~al.(2021)Davenport, Burke, and Diffenbaugh]{Davenport2021}
Frances~V. Davenport, Marshall Burke, and Noah~S. Diffenbaugh.
\newblock Contribution of historical precipitation change to us flood damages.
\newblock \emph{Proc. Natl. Acad. Sci. USA}, 118\penalty0 (4):\penalty0 e2017524118, 2021.

\bibitem[Dueben et~al.(2022)Dueben, Schultz, Chantry, Gagne, Hall, and McGovern]{dueben2022challenges}
Peter~D Dueben, Martin~G Schultz, Matthew Chantry, David~John Gagne, David~Matthew Hall, and Amy McGovern.
\newblock Challenges and benchmark datasets for machine learning in the atmospheric sciences: Definition, status, and outlook.
\newblock \emph{Artificial Intelligence for the Earth Systems}, 1\penalty0 (3):\penalty0 e210002, 2022.

\bibitem[Ebert-Uphoff et~al.(2017)Ebert-Uphoff, Thompson, Demir, Gel, Karpatne, Guereque, Kumar, Cabral-Cano, and Smyth]{ebert2017vision}
Imme Ebert-Uphoff, David~R Thompson, Ibrahim Demir, Yulia~R Gel, Anuj Karpatne, Mariana Guereque, Vipin Kumar, Enrique Cabral-Cano, and Padhraic Smyth.
\newblock A vision for the development of benchmarks to bridge geoscience and data science.
\newblock In \emph{17th International Workshop on Climate Informatics}, 2017.

\bibitem[Emanuel(1994)]{Emanuel1994}
Kerry~A Emanuel.
\newblock \emph{Atmospheric convection}.
\newblock 1994.

\bibitem[Eyring et~al.(2016)Eyring, Bony, Meehl, Senior, Stevens, Stouffer, and Taylor]{eyring2016overview}
Veronika Eyring, Sandrine Bony, Gerald~A Meehl, Catherine~A Senior, Bjorn Stevens, Ronald~J Stouffer, and Karl~E Taylor.
\newblock Overview of the coupled model intercomparison project phase 6 (cmip6) experimental design and organization.
\newblock \emph{Geoscientific Model Development}, 9\penalty0 (5):\penalty0 1937--1958, 2016.

\bibitem[Eyring et~al.(2020)Eyring, Bock, Lauer, Righi, Schlund, Andela, Arnone, Bellprat, Br\"{o}tz, Caron, Carvalhais, Cionni, Cortesi, Crezee, Davin, Davini, Debeire, de~Mora, Deser, Docquier, Earnshaw, Ehbrecht, Gier, Gonzalez-Reviriego, Goodman, Hagemann, Hardiman, Hassler, Hunter, Kadow, Kindermann, Koirala, Koldunov, Lejeune, Lembo, Lovato, Lucarini, Massonnet, M\"{u}ller, Pandde, P{\'{e}}rez-Zan{\'{o}}n, Phillips, Predoi, Russell, Sellar, Serva, Stacke, Swaminathan, Torralba, Vegas-Regidor, von Hardenberg, Weigel, and Zimmermann]{Eyring_2020}
Veronika Eyring, Lisa Bock, Axel Lauer, Mattia Righi, Manuel Schlund, Bouwe Andela, Enrico Arnone, Omar Bellprat, Bj\"{o}rn Br\"{o}tz, Louis-Philippe Caron, Nuno Carvalhais, Irene Cionni, Nicola Cortesi, Bas Crezee, Edouard~L. Davin, Paolo Davini, Kevin Debeire, Lee de~Mora, Clara Deser, David Docquier, Paul Earnshaw, Carsten Ehbrecht, Bettina~K. Gier, Nube Gonzalez-Reviriego, Paul Goodman, Stefan Hagemann, Steven Hardiman, Birgit Hassler, Alasdair Hunter, Christopher Kadow, Stephan Kindermann, Sujan Koirala, Nikolay Koldunov, Quentin Lejeune, Valerio Lembo, Tomas Lovato, Valerio Lucarini, Fran{\c{c}}ois Massonnet, Benjamin M\"{u}ller, Amarjiit Pandde, N{\'{u}}ria P{\'{e}}rez-Zan{\'{o}}n, Adam Phillips, Valeriu Predoi, Joellen Russell, Alistair Sellar, Federico Serva, Tobias Stacke, Ranjini Swaminathan, Ver{\'{o}}nica Torralba, Javier Vegas-Regidor, Jost von Hardenberg, Katja Weigel, and Klaus Zimmermann.
\newblock {Earth System Model Evaluation Tool ({ESMValTool}) v2.0 {\textendash} an extended set of large-scale diagnostics for quasi-operational and comprehensive evaluation of Earth system models in {CMIP}}.
\newblock \emph{Geoscientific Model Development}, 13\penalty0 (7):\penalty0 3383--3438, 2020.
\newblock \doi{10.5194/gmd-13-3383-2020}.
\newblock URL \url{https://doi.org/10.5194/gmd-13-3383-2020}.

\bibitem[Eyring et~al.(2021)Eyring, Mishra, Griffith, Chen, Keenan, Turetsky, Brown, Jotzo, Moore, and van~der Linden]{Eyring2021}
Veronika Eyring, Vimal Mishra, Gary~P. Griffith, Lei Chen, Trevor Keenan, Merritt~R. Turetsky, Sally Brown, Frank Jotzo, Frances~C. Moore, and Sander van~der Linden.
\newblock Reflections and projections on a decade of climate science.
\newblock \emph{Nat. Clim. Change}, 11\penalty0 (4):\penalty0 279--285, 2021.

\bibitem[Eyring et~al.(2024)Eyring, Collins, Gentine, Barnes, Barreiro, Beucler, Bocquet, Bretherton, Christensen, Gagne, Hall, Hammerling, Hoyer, Iglesias-Suarez, Lopez-Gomez, McGraw, Meehl, Molina, Monteleoni, Mueller, Pritchard, Rolnick, Runge, Stier, Watt-Meyer, Weigel, Yu, and Zanna]{eyring24natcc_AGCIMLClimate}
Veronika Eyring, William~D. Collins, Pierre Gentine, Elizabeth~A. Barnes, Marcelo Barreiro, Tom Beucler, Marc Bocquet, Christopher~S. Bretherton, Hannah~M. Christensen, David~John Gagne, David Hall, Dorit Hammerling, Stephan Hoyer, Fernando Iglesias-Suarez, Ignacio Lopez-Gomez, Marie~C. McGraw, Gerald~A. Meehl, Maria~J. Molina, Claire Monteleoni, Juliane Mueller, Michael~S. Pritchard, David Rolnick, Jakob Runge, Philip Stier, Oliver Watt-Meyer, Katja Weigel, Rose Yu, and Laure Zanna.
\newblock Pushing the frontiers in climate modeling and analysis with machine learning.
\newblock \emph{Nature Climate Change}, accepted, 2024.

\bibitem[Eyring et~al.(2025, accepted)Eyring, Gentine, Camps-Valls, Lawrence, and Reichstein]{eyring24_hybridmodelling}
Veronika Eyring, Pierre Gentine, Gustau Camps-Valls, David~M. Lawrence, and Markus Reichstein.
\newblock Ai-empowered next-generation multiscale climate modeling for mitigation and adaptation.
\newblock \emph{Nature Geoscience}, 2025, accepted.

\bibitem[Fischer et~al.(2007)Fischer, Seneviratne, Vidale, L{\"u}thi, and Sch{\"a}r]{Fischer2007}
Erich~M Fischer, Sonia~I Seneviratne, Pier~Luigi Vidale, Daniel L{\"u}thi, and Christoph Sch{\"a}r.
\newblock Soil moisture--atmosphere interactions during the 2003 european summer heat wave.
\newblock \emph{J. Clim.}, 20\penalty0 (20):\penalty0 5081--5099, 2007.

\bibitem[Gates et~al.(1999)Gates, Boyle, Covey, Dease, Doutriaux, Drach, Fiorino, Gleckler, Hnilo, Marlais, et~al.]{gates1999overview}
W~Lawrence Gates, James~S Boyle, Curt Covey, Clyde~G Dease, Charles~M Doutriaux, Robert~S Drach, Michael Fiorino, Peter~J Gleckler, Justin~J Hnilo, Susan~M Marlais, et~al.
\newblock An overview of the results of the atmospheric model intercomparison project (amip i).
\newblock \emph{Bulletin of the American Meteorological Society}, 80\penalty0 (1):\penalty0 29--56, 1999.

\bibitem[Gentine et~al.(2018)Gentine, Pritchard, Rasp, Reinaudi, and Yacalis]{Gentine2018}
P~Gentine, M~Pritchard, S~Rasp, G~Reinaudi, and G~Yacalis.
\newblock Could machine learning break the convection parameterization deadlock?
\newblock \emph{Geophys. Res. Lett.}, 45\penalty0 (11):\penalty0 5742--5751, 2018.

\bibitem[Gentine et~al.(2021)Gentine, Eyring, and Beucler]{Gentine2021}
Pierre Gentine, Veronika Eyring, and Tom Beucler.
\newblock Deep learning for the parametrization of subgrid processes in climate models.
\newblock In \emph{Deep Learning for the Earth Sciences}, pages 307--314. 2021.

\bibitem[Grabowski and Smolarkiewicz(1999)]{Grabowski1999}
Wojciech~W Grabowski and Piotr~K Smolarkiewicz.
\newblock Crcp: A cloud resolving convection parameterization for modeling the tropical convecting atmosphere.
\newblock \emph{Phys. D: Nonlinear Phenom.}, 133\penalty0 (1-4):\penalty0 171--178, 1999.

\bibitem[Grundner et~al.(2022)Grundner, Beucler, Gentine, Iglesias-Suarez, Giorgetta, and Eyring]{Grundner2022Cloud}
Arthur Grundner, Tom Beucler, Pierre Gentine, Fernando Iglesias-Suarez, Marco~A. Giorgetta, and Veronika Eyring.
\newblock {Deep Learning Based Cloud Cover Parameterization for ICON}.
\newblock \emph{Journal of Advances in Modeling Earth Systems}, 14\penalty0 (12):\penalty0 e2021MS002959, 2022.
\newblock \doi{https://doi.org/10.1029/2021MS002959}.

\bibitem[Grundner et~al.(2023)Grundner, Beucler, Gentine, and Eyring]{Grundner2023}
Arthur Grundner, Tom Beucler, Pierre Gentine, and Veronika Eyring.
\newblock Data-driven equation discovery of a cloud cover parameterization, 2023.
\newblock arxiv:2304.08063.

\bibitem[Guerreiro et~al.(2018)Guerreiro, Fowler, Barbero, Westra, Lenderink, Blenkinsop, Lewis, and Li]{Guerreiro2018}
Selma~B. Guerreiro, Hayley~J. Fowler, Renaud Barbero, Seth Westra, Geert Lenderink, Stephen Blenkinsop, Elizabeth Lewis, and Xiao~Feng Li.
\newblock Detection of continental-scale intensification of hourly rainfall extremes.
\newblock \emph{Nat. Clim. Change}, 8\penalty0 (9):\penalty0 803--807, 2018.

\bibitem[Han et~al.(2020)Han, Zhang, Huang, and Wang]{Han2020}
Yilun Han, Guang~J Zhang, Xiaomeng Huang, and Yong Wang.
\newblock A moist physics parameterization based on deep learning.
\newblock \emph{J. Adv. Model. Earth Syst.}, 12\penalty0 (9):\penalty0 e2020MS002076, 2020.

\bibitem[Han et~al.(2023)Han, Zhang, and Wang]{han2023ensemble}
Yilun Han, Guang~J Zhang, and Yong Wang.
\newblock An ensemble of neural networks for moist physics processes, its generalizability and stable integration.
\newblock \emph{Journal of Advances in Modeling Earth Systems}, 15\penalty0 (10):\penalty0 e2022MS003508, 2023.

\bibitem[Hannah et~al.(2020)Hannah, Jones, Hillman, Norman, Bader, Taylor, Leung, Pritchard, Branson, Lin, et~al.]{Hannah2020}
Walter~M Hannah, Christopher~R Jones, Benjamin~R Hillman, Matthew~R Norman, David~C Bader, Mark~A Taylor, LR~Leung, Michael~S Pritchard, Mark~D Branson, Guangxing Lin, et~al.
\newblock Initial results from the super-parameterized e3sm.
\newblock \emph{Journal of Advances in Modeling Earth Systems}, 12\penalty0 (1):\penalty0 e2019MS001863, 2020.

\bibitem[Hannah et~al.(2021)Hannah, Bradley, Guba, Tang, Golaz, and Wolfe]{Hannah2021}
Walter~M. Hannah, Andrew~M. Bradley, Oksana Guba, Qi~Tang, Jean-Christophe Golaz, and Jon Wolfe.
\newblock Separating physics and dynamics grids for improved computational efficiency in spectral element earth system models.
\newblock \emph{J. Adv. Model. Earth Syst.}, 13\penalty0 (7):\penalty0 e2020MS002419, 2021.

\bibitem[Hannah et~al.(2022)Hannah, Pressel, Ovchinnikov, and Elsaesser]{Hannah2022a}
Walter~M. Hannah, Kyle~G. Pressel, Mikhail Ovchinnikov, and Gregory~S. Elsaesser.
\newblock Checkerboard patterns in e3smv2 and e3sm-mmfv2.
\newblock \emph{Geosci. Model Dev.}, 15\penalty0 (9):\penalty0 6243--6257, 2022.

\bibitem[Hersbach et~al.(2020)Hersbach, Bell, Berrisford, Hirahara, Hor{\'a}nyi, Mu{\~n}oz-Sabater, Nicolas, Peubey, Radu, Schepers, et~al.]{hersbach2020era5}
Hans Hersbach, Bill Bell, Paul Berrisford, Shoji Hirahara, Andr{\'a}s Hor{\'a}nyi, Joaqu{\'\i}n Mu{\~n}oz-Sabater, Julien Nicolas, Carole Peubey, Raluca Radu, Dinand Schepers, et~al.
\newblock The era5 global reanalysis.
\newblock \emph{Quarterly Journal of the Royal Meteorological Society}, 146\penalty0 (730):\penalty0 1999--2049, 2020.

\bibitem[Heuer et~al.(2023)Heuer, Schwabe, Gentine, Giorgetta, and Eyring]{heuer2023interpretable}
Helge Heuer, Mierk Schwabe, Pierre Gentine, Marco~A Giorgetta, and Veronika Eyring.
\newblock Interpretable multiscale machine learning-based parameterizations of convection for icon.
\newblock \emph{arXiv preprint arXiv:2311.03251}, 2023.

\bibitem[Hohenegger et~al.(2023)Hohenegger, Korn, Linardakis, Redler, Schnur, Adamidis, Bao, Bastin, Behravesh, Bergemann, et~al.]{hohenegger2023icon}
Cathy Hohenegger, Peter Korn, Leonidas Linardakis, Ren{\'e} Redler, Reiner Schnur, Panagiotis Adamidis, Jiawei Bao, Swantje Bastin, Milad Behravesh, Martin Bergemann, et~al.
\newblock Icon-sapphire: simulating the components of the earth system and their interactions at kilometer and subkilometer scales.
\newblock \emph{Geoscientific Model Development}, 16\penalty0 (2):\penalty0 779--811, 2023.

\bibitem[Hu et~al.(2024)Hu, Subramaniam, Kuang, Lin, Yu, Hannah, Brenowitz, Romero, and Pritchard]{Hu2024}
Zeyuan Hu, Akshay Subramaniam, Zhiming Kuang, Jerry Lin, Sungduk Yu, Walter~M. Hannah, Noah~D. Brenowitz, Josh Romero, and Michael~S Pritchard.
\newblock Stable machine-learning parameterization of subgrid processes with real geography and full-physics emulation.
\newblock \emph{arXiv preprint arXiv:2407.00124}, 2024.

\bibitem[Iglesias-Suarez et~al.(2023)Iglesias-Suarez, Gentine, Solino-Fernandez, Beucler, Pritchard, Runge, and Eyring]{Iglesias2023}
Fernando Iglesias-Suarez, Pierre Gentine, Breixo Solino-Fernandez, Tom Beucler, Michael Pritchard, Jakob Runge, and Veronika Eyring.
\newblock Causally-informed deep learning to improve climate models and projections, 2023.
\newblock arxiv:2304.12952.

\bibitem[IPCC(2021)]{IPCC2021}
IPCC.
\newblock \emph{Climate Change 2021: The Physical Science Basis. Contribution of Working Group I to the Sixth Assessment Report of the Intergovernmental Panel on Climate Change}.
\newblock 2021.

\bibitem[Kaltenborn et~al.(2023)Kaltenborn, Lange, Ramesh, Brouillard, Gurwicz, Nagda, Runge, Nowack, and Rolnick]{kaltenborn2023climateset}
Julia Kaltenborn, Charlotte Lange, Venkatesh Ramesh, Philippe Brouillard, Yaniv Gurwicz, Chandni Nagda, Jakob Runge, Peer Nowack, and David Rolnick.
\newblock Climateset: A large-scale climate model dataset for machine learning.
\newblock \emph{Advances in Neural Information Processing Systems}, 36:\penalty0 21757--21792, 2023.

\bibitem[Khairoutdinov et~al.(2008)Khairoutdinov, DeMott, and Randall]{Khairoutdinov2008}
Marat Khairoutdinov, Charlotte DeMott, and David Randall.
\newblock Evaluation of the simulated interannual and subseasonal variability in an amip-style simulation using the csu multiscale modeling framework.
\newblock \emph{J. Clim.}, 21\penalty0 (3):\penalty0 413--431, 2008.

\bibitem[Kim et~al.(2019)Kim, Azevedo, Thuerey, Kim, Gross, and Solenthaler]{Kim2019}
Byungsoo Kim, Vinicius~C. Azevedo, Nils Thuerey, Theodore Kim, Markus Gross, and Barbara Solenthaler.
\newblock Deep fluids: A generative network for parameterized fluid simulations.
\newblock \emph{Comput. Graph. Forum}, 38\penalty0 (2):\penalty0 59--70, 2019.

\bibitem[Kochkov et~al.(2023)Kochkov, Yuval, Langmore, Norgaard, Smith, Mooers, Lottes, Rasp, D{\"u}ben, Kl{\"o}wer, et~al.]{kochkov2023neural}
Dmitrii Kochkov, Janni Yuval, Ian Langmore, Peter Norgaard, Jamie Smith, Griffin Mooers, James Lottes, Stephan Rasp, Peter D{\"u}ben, Milan Kl{\"o}wer, et~al.
\newblock Neural general circulation models.
\newblock \emph{arXiv preprint arXiv:2311.07222}, 2023.

\bibitem[K{\"u}hbacher et~al.(2024)K{\"u}hbacher, Iglesias-Suarez, Kilbertus, and Eyring]{kuhbacher2024towards}
Birgit K{\"u}hbacher, Fernando Iglesias-Suarez, Niki Kilbertus, and Veronika Eyring.
\newblock Towards physically consistent deep learning for climate model parameterizations.
\newblock \emph{arXiv preprint arXiv:2406.03920}, 2024.

\bibitem[Kwa et~al.(2023)Kwa, Clark, Henn, Brenowitz, McGibbon, Watt-Meyer, Perkins, Harris, and Bretherton]{Kwa2023}
Anna Kwa, Spencer~K Clark, Brian Henn, Noah~D Brenowitz, Jeremy McGibbon, Oliver Watt-Meyer, W~Andre Perkins, Lucas Harris, and Christopher~S Bretherton.
\newblock Machine-learned climate model corrections from a global storm-resolving model: Performance across the annual cycle.
\newblock \emph{J. Adv. Model. Earth Syst.}, 15\penalty0 (5):\penalty0 e2022MS003400, 2023.

\bibitem[Lam et~al.(2022)Lam, Sanchez-Gonzalez, Willson, Wirnsberger, Fortunato, Pritzel, Ravuri, Ewalds, Alet, Eaton-Rosen, Hu, Merose, Hoyer, Holland, Stott, Vinyals, Mohamed, and Battaglia]{Lam2022}
Remi Lam, Alvaro Sanchez-Gonzalez, Matthew Willson, Peter Wirnsberger, Meire Fortunato, Alexander Pritzel, Suman Ravuri, Timo Ewalds, Ferran Alet, Zach Eaton-Rosen, Weihua Hu, Alexander Merose, Stephan Hoyer, George Holland, Jacklynn Stott, Oriol Vinyals, Shakir Mohamed, and Peter Battaglia.
\newblock Graphcast: Learning skillful medium-range global weather forecasting, 2022.
\newblock arxiv:2212.12794.

\bibitem[Li et~al.(2021)Li, Kovachki, Azizzadenesheli, Liu, Bhattacharya, Stuart, and Anandkumar]{Li2021}
Zongyi Li, Nikola Kovachki, Kamyar Azizzadenesheli, Burigede Liu, Kaushik Bhattacharya, Andrew Stuart, and Anima Anandkumar.
\newblock Fourier neural operator for parametric partial differential equations, 2021.
\newblock arxiv:2010.08895.

\bibitem[Li et~al.(2023)Li, Zheng, Kovachki, Jin, Chen, Liu, Azizzadenesheli, and Anandkumar]{Li2023}
Zongyi Li, Hongkai Zheng, Nikola Kovachki, David Jin, Haoxuan Chen, Burigede Liu, Kamyar Azizzadenesheli, and Anima Anandkumar.
\newblock Physics-informed neural operator for learning partial differential equations, 2023.
\newblock arxiv:2111.03794.

\bibitem[Lin and Neelin(2000)]{Lin2000}
J.~W.-B. Lin and J.~D. Neelin.
\newblock Influence of a stochastic moist convective parameterization on tropical climate variability.
\newblock \emph{Geophys. Res. Lett.}, 27\penalty0 (22):\penalty0 3691--3694, 2000.

\bibitem[Lin and Neelin(2003)]{Lin2003}
J.~W.-B. Lin and J.~D. Neelin.
\newblock Toward stochastic moist convective parameterization in general circulation models.
\newblock \emph{Geophys. Res. Lett.}, 30 (4):\penalty0 1162, 2003.

\bibitem[Lin et~al.(2023)Lin, Yu, Beucler, Gentine, Walling, and Pritchard]{lin2023systematic}
Jerry Lin, Sungduk Yu, Tom Beucler, Pierre Gentine, David Walling, and Mike Pritchard.
\newblock Systematic sampling and validation of machine learning-parameterizations in climate models.
\newblock \emph{arXiv preprint arXiv:2309.16177}, 2023.

\bibitem[Lin et~al.(2024)Lin, Hu, Yu, Pritchard, Gupta, Zheng, Hannah, Mansfield, Qu, Geleta, Lopez, Rudolph, Chow, and Reade]{leap-atmospheric-physics-ai-climsim}
Jerry Lin, Zeyuan Hu, Sungduk Yu, Michael~S Pritchard, Ritwik Gupta, Tian Zheng, Walter Hannah, Laura Mansfield, Yongquan Qu, Margarita Geleta, Molly Lopez, Maja Rudolph, Ashley Chow, and Walter Reade.
\newblock Leap - atmospheric physics using ai (climsim), 2024.
\newblock URL \url{https://kaggle.com/competitions/leap-atmospheric-physics-ai-climsim}.

\bibitem[Ling et~al.(2016)Ling, Kurzawski, and Templeton]{Ling2016}
Julia Ling, Andrew Kurzawski, and Jeremy Templeton.
\newblock Reynolds averaged turbulence modelling using deep neural networks with embedded invariance.
\newblock \emph{J. Fluid Mech.}, 807:\penalty0 155--166, 2016.

\bibitem[Luccioni and Hernandez-Garcia(2023)]{luccioni2023counting}
Alexandra~Sasha Luccioni and Alex Hernandez-Garcia.
\newblock Counting carbon: A survey of factors influencing the emissions of machine learning.
\newblock \emph{arXiv preprint arXiv:2302.08476}, 2023.

\bibitem[Lütjens et~al.(2022)Lütjens, Crawford, Watson, Hill, and Newman]{Lütjens2022}
Björn Lütjens, Catherine~H. Crawford, Campbell~D Watson, Christopher Hill, and Dava Newman.
\newblock Multiscale neural operator: Learning fast and grid-independent pde solvers, 2022.
\newblock arxiv:2207.11417.

\bibitem[MacArt et~al.(2021)MacArt, Sirignano, and Freund]{MacArt2021}
Jonathan~F MacArt, Justin Sirignano, and Jonathan~B Freund.
\newblock Embedded training of neural-network subgrid-scale turbulence models.
\newblock \emph{Phys. Rev. Fluids}, 6\penalty0 (5):\penalty0 050502, 2021.

\bibitem[Martinez-Villalobos and Neelin(2023)]{Martinez-Villalobos2023}
Cristian Martinez-Villalobos and J.~David Neelin.
\newblock Regionally high risk increase for precipitation extreme events under global warming.
\newblock \emph{Sci. Rep.}, 13:\penalty0 5579, 2023.

\bibitem[Mooers et~al.(2021)Mooers, Pritchard, Beucler, Ott, Yacalis, Baldi, and Gentine]{Mooers2021}
Griffin Mooers, Michael Pritchard, Tom Beucler, Jordan Ott, Galen Yacalis, Pierre Baldi, and Pierre Gentine.
\newblock Assessing the potential of deep learning for emulating cloud superparameterization in climate models with real‐geography boundary conditions.
\newblock \emph{J. Adv. Model. Earth Syst.}, 13\penalty0 (5):\penalty0 e2020MS002385, 2021.

\bibitem[Mooers et~al.(2022)Mooers, Beucler, Pritchard, and Mandt]{mooers2022understanding}
Griffin Mooers, Tom Beucler, Mike Pritchard, and Stephan Mandt.
\newblock Understanding extreme precipitation changes through unsupervised machine learning.
\newblock \emph{arXiv preprint arXiv:2211.01613}, 2022.

\bibitem[Mooers et~al.(2023)Mooers, Pritchard, Beucler, Srivastava, Mangipudi, Peng, Gentine, and Mandt]{mooers2023comparing}
Griffin Mooers, Mike Pritchard, Tom Beucler, Prakhar Srivastava, Harshini Mangipudi, Liran Peng, Pierre Gentine, and Stephan Mandt.
\newblock Comparing storm resolving models and climates via unsupervised machine learning.
\newblock \emph{Scientific Reports}, 13\penalty0 (1):\penalty0 22365, 2023.

\bibitem[Neelin et~al.(2008)Neelin, Peters, Lin, Hales, and Holloway]{Neelin2008}
J.~D. Neelin, O.~Peters, J.~W.-B. Lin, K.~Hales, and C.~E. Holloway.
\newblock Rethinking convective quasi-equilibrium: observational constraints for stochastic convective schemes in climate models.
\newblock \emph{Phil. Trans. Royal Soc. A}, 366\penalty0 (1875):\penalty0 2581--2604, 2008.

\bibitem[Neelin et~al.(2022)Neelin, Martinez-Villalobos, Stechmann, Ahmed, Chen, Norris, Kuo, and Lenderink]{Neelin2022}
J~David Neelin, Cristian Martinez-Villalobos, Samuel~N Stechmann, Fiaz Ahmed, Gang Chen, Jesse~M Norris, Yi-Hung Kuo, and Geert Lenderink.
\newblock Precipitation extremes and water vapor: Relationships in current climate and implications for climate change.
\newblock \emph{Current Clim. Change Rep.}, 8\penalty0 (1):\penalty0 17--33, 2022.

\bibitem[Neelin et~al.(2023)Neelin, Krasting, Radhakrishnan, Liptak, Jackson, Ming, Dong, Gettelman, Coleman, Maloney, et~al.]{neelin2023process}
J~David Neelin, John~P Krasting, Aparna Radhakrishnan, Jessica Liptak, Thomas Jackson, Yi~Ming, Wenhao Dong, Andrew Gettelman, Danielle~R Coleman, Eric~D Maloney, et~al.
\newblock Process-oriented diagnostics: Principles, practice, community development, and common standards.
\newblock \emph{Bulletin of the American Meteorological Society}, 104\penalty0 (8):\penalty0 E1452--E1468, 2023.

\bibitem[Niu et~al.(2024)Niu, Wu, Kim, Ma, Watson-Parris, and Yu]{niu2024multi}
Ruijia Niu, Dongxia Wu, Kai Kim, Yi-An Ma, Duncan Watson-Parris, and Rose Yu.
\newblock Multi-fidelity residual neural processes for scalable surrogate modeling.
\newblock 2024.

\bibitem[Norman et~al.(2022)Norman, Bader, Eldred, Hannah, Hillman, Jones, Lee, Leung, Lyngaas, Pressel, et~al.]{Norman2022}
Matthew~R Norman, David~C Bader, Christopher Eldred, Walter~M Hannah, Benjamin~R Hillman, Christopher~R Jones, Jungmin~M Lee, LR~Leung, Isaac Lyngaas, Kyle~G Pressel, et~al.
\newblock Unprecedented cloud resolution in a gpu-enabled full-physics atmospheric climate simulation on olcf’s summit supercomputer.
\newblock \emph{Int. J. High Perform. Compu. Appl.}, 36\penalty0 (1):\penalty0 93--105, 2022.

\bibitem[Osband et~al.(2018)Osband, Aslanides, and Cassirer]{Osband2018}
Ian Osband, John Aslanides, and Albin Cassirer.
\newblock Randomized prior functions for deep reinforcement learning, 2018.
\newblock arxiv:1806.03335.

\bibitem[Ott et~al.(2020)Ott, Pritchard, Best, Linstead, Curcic, and Baldi]{ott2020}
Jordan Ott, Mike Pritchard, Natalie Best, Erik Linstead, Milan Curcic, and Pierre Baldi.
\newblock A fortran-keras deep learning bridge for scientific computing, 2020.
\newblock arxiv:2004.10652.

\bibitem[O’Gorman(2015)]{O'Gorman2015}
Paul~A O’Gorman.
\newblock Precipitation extremes under climate change.
\newblock \emph{Current Clim. Change Rep.}, 1:\penalty0 49--59, 2015.

\bibitem[Pall et~al.(2007)Pall, Allen, and Stone]{Pall2007}
P.~Pall, M.~R. Allen, and D.~A. Stone.
\newblock Testing the clausius – clapeyron constraint on changes in extreme precipitation under co2 warming.
\newblock \emph{Clim. Dyn.}, 28\penalty0 (4):\penalty0 351--363, 2007.

\bibitem[Pathak et~al.(2022)Pathak, Subramanian, Harrington, Raja, Chattopadhyay, Mardani, Kurth, Hall, Li, Azizzadenesheli, Hassanzadeh, Kashinath, and Anandkumar]{Pathak2022}
Jaideep Pathak, Shashank Subramanian, Peter Harrington, Sanjeev Raja, Ashesh Chattopadhyay, Morteza Mardani, Thorsten Kurth, David Hall, Zongyi Li, Kamyar Azizzadenesheli, Pedram Hassanzadeh, Karthik Kashinath, and Animashree Anandkumar.
\newblock Fourcastnet: A global data-driven high-resolution weather model using adaptive fourier neural operators, 2022.
\newblock arxiv:2202.11214.

\bibitem[Pendergrass and Hartmann(2014)]{Pendergrass2014}
Angeline~G. Pendergrass and Dennis~L. Hartmann.
\newblock Two modes of change of the distribution of rain.
\newblock \emph{J. Clim.}, 27\penalty0 (22):\penalty0 8357--8371, 2014.

\bibitem[Perkins et~al.(2024)Perkins, Brenowitz, Bretherton, and Nugent]{perkins2024emulation}
W~Andre Perkins, Noah~D Brenowitz, Christopher~S Bretherton, and Jacqueline~M Nugent.
\newblock Emulation of cloud microphysics in a climate model.
\newblock \emph{Journal of Advances in Modeling Earth Systems}, 16\penalty0 (4):\penalty0 e2023MS003851, 2024.

\bibitem[Randall(2012)]{Randall2012}
David Randall.
\newblock \emph{Atmosphere, clouds, and climate}, volume~6.
\newblock 2012.

\bibitem[Randall et~al.(2003)Randall, Khairoutdinov, Arakawa, and Grabowski]{Randall2003}
David Randall, Marat Khairoutdinov, Akio Arakawa, and Wojciech Grabowski.
\newblock Breaking the cloud parameterization deadlock.
\newblock \emph{Bull. Am. Meteorol. Soc.}, 84\penalty0 (11):\penalty0 1547--1564, 2003.

\bibitem[Randall(2013)]{Randall2013}
David~A Randall.
\newblock Beyond deadlock.
\newblock \emph{Geophys. Res. Lett.}, 40\penalty0 (22):\penalty0 5970--5976, 2013.

\bibitem[Rasp(2020)]{Rasp2020b}
Stephan Rasp.
\newblock Coupled online learning as a way to tackle instabilities and biases in neural network parameterizations: general algorithms and lorenz 96 case study (v1. 0).
\newblock \emph{Geosci. Model Dev.}, 13\penalty0 (5):\penalty0 2185--2196, 2020.

\bibitem[Rasp et~al.(2018)Rasp, Pritchard, and Gentine]{Rasp2018}
Stephan Rasp, Michael~S Pritchard, and Pierre Gentine.
\newblock Deep learning to represent subgrid processes in climate models.
\newblock \emph{Proc. Natl. Acad. Sci. USA}, 115\penalty0 (39):\penalty0 9684--9689, 2018.

\bibitem[Rasp et~al.(2024)Rasp, Hoyer, Merose, Langmore, Battaglia, Russell, Sanchez-Gonzalez, Yang, Carver, Agrawal, et~al.]{rasp2024weatherbench}
Stephan Rasp, Stephan Hoyer, Alexander Merose, Ian Langmore, Peter Battaglia, Tyler Russell, Alvaro Sanchez-Gonzalez, Vivian Yang, Rob Carver, Shreya Agrawal, et~al.
\newblock Weatherbench 2: A benchmark for the next generation of data-driven global weather models.
\newblock \emph{Journal of Advances in Modeling Earth Systems}, 16\penalty0 (6):\penalty0 e2023MS004019, 2024.

\bibitem[Reed et~al.(2023)Reed, Gupta, Li, Brockman, Funk, Clipp, Keutzer, Candido, Uyttendaele, and Darrell]{Reed2023}
Colorado~J. Reed, Ritwik Gupta, Shufan Li, Sarah Brockman, Christopher Funk, Brian Clipp, Kurt Keutzer, Salvatore Candido, Matt Uyttendaele, and Trevor Darrell.
\newblock Scale-mae: A scale-aware masked autoencoder for multiscale geospatial representation learning, 2023.
\newblock arxiv:2212.14532.

\bibitem[Ronchi et~al.(1996)Ronchi, Iacono, and Paolucci]{ronchi1996cubed}
Corrado Ronchi, Roberto Iacono, and Pier~S Paolucci.
\newblock The “cubed sphere”: A new method for the solution of partial differential equations in spherical geometry.
\newblock \emph{Journal of computational physics}, 124\penalty0 (1):\penalty0 93--114, 1996.

\bibitem[Ross et~al.(2023)Ross, Li, Perezhogin, Fernandez-Granda, and Zanna]{ross2023benchmarking}
Andrew Ross, Ziwei Li, Pavel Perezhogin, Carlos Fernandez-Granda, and Laure Zanna.
\newblock Benchmarking of machine learning ocean subgrid parameterizations in an idealized model.
\newblock \emph{Journal of Advances in Modeling Earth Systems}, 15\penalty0 (1), 2023.

\bibitem[Sanford et~al.(2023{\natexlab{a}})Sanford, Kwa, Watt-Meyer, Clark, Brenowitz, McGibbon, and Bretherton]{sanford2023improving}
Clayton Sanford, Anna Kwa, Oliver Watt-Meyer, Spencer~K Clark, Noah Brenowitz, Jeremy McGibbon, and Christopher Bretherton.
\newblock Improving the reliability of ml-corrected climate models with novelty detection.
\newblock \emph{Journal of Advances in Modeling Earth Systems}, 15\penalty0 (11):\penalty0 e2023MS003809, 2023{\natexlab{a}}.

\bibitem[Sanford et~al.(2023{\natexlab{b}})Sanford, Kwa, Watt-Meyer, Clark, Brenowitz, McGibbon, and Bretherton]{Sanford2023}
Clayton~Hendrick Sanford, Anna Kwa, Oliver Watt-Meyer, Spencer~Koncius Clark, Noah~Domino Brenowitz, Jeremy McGibbon, and Christopher~S Bretherton.
\newblock Improving the reliability of ml-corrected climate models with novelty detection.
\newblock \emph{Authorea Preprints}, 2023{\natexlab{b}}.

\bibitem[Schneider et~al.(2017)Schneider, Teixeira, Bretherton, Brient, Pressel, Sch{\"a}r, and Siebesma]{Schneider2017}
Tapio Schneider, Jo{\~a}o Teixeira, Christopher~S Bretherton, Florent Brient, Kyle~G Pressel, Christoph Sch{\"a}r, and A~Pier Siebesma.
\newblock Climate goals and computing the future of clouds.
\newblock \emph{Nat. Clim. Change}, 7\penalty0 (1):\penalty0 3--5, 2017.

\bibitem[Seneviratne et~al.(2021)Seneviratne, Zhang, Adnan, Badi, Dereczynski, Luca, Ghosh, Iskandar, Kossin, Lewis, Otto, Pinto, Satoh, Vicente-Serrano, Wehner, and Zhou]{seneviratne2021ipcc}
S.I. Seneviratne, X.~Zhang, M.~Adnan, W.~Badi, C.~Dereczynski, A.~Di Luca, S.~Ghosh, I.~Iskandar, J.~Kossin, S.~Lewis, F.~Otto, I.~Pinto, M.~Satoh, S.M. Vicente-Serrano, M.~Wehner, and B.~Zhou.
\newblock {Weather and Climate Extreme Events in a Changing Climate}.
\newblock In V.~Masson-Delmotte, P.~Zhai, A.~Pirani, S.L. Connors, C.~Péan, S.~Berger, N.~Caud, Y.~Chen, L.~Goldfarb, M.I. Gomis, M.~Huang, K.~Leitzell, E.~Lonnoy, J.B.R. Matthews, T.K. Maycock, T.~Waterfield, O.~Yelekçi, R.~Yu, , and B.~Zhou, editors, \emph{Climate Change 2021: The Physical Science Basis. Contribution of Working Group I to the Sixth Assessment Report of the Intergovernmental Panel on Climate Change}, chapter~11, page 1513–1766. Cambridge University Press, Cambridge, United Kingdom and New York, NY, USA, 2021.
\newblock URL \url{https://www.ipcc.ch/report/ar6/wg1/}.

\bibitem[Seneviratne et~al.(2010)Seneviratne, Corti, Davin, Hirschi, Jaeger, Lehner, Orlowsky, and Teuling]{Seneviratne2010}
Sonia~I Seneviratne, Thierry Corti, Edouard~L Davin, Martin Hirschi, Eric~B Jaeger, Irene Lehner, Boris Orlowsky, and Adriaan~J Teuling.
\newblock Investigating soil moisture--climate interactions in a changing climate: A review.
\newblock \emph{Earth-Sci. Rev.}, 99\penalty0 (3-4):\penalty0 125--161, 2010.

\bibitem[Sherwood et~al.(2020)Sherwood, Webb, Annan, Armour, Forster, Hargreaves, Hegerl, Klein, Marvel, Rohling, et~al.]{Sherwood2020}
SC~Sherwood, Mark~J Webb, James~D Annan, Kyle~C Armour, Piers~M Forster, Julia~C Hargreaves, Gabriele Hegerl, Stephen~A Klein, Kate~D Marvel, Eelco~J Rohling, et~al.
\newblock An assessment of earth's climate sensitivity using multiple lines of evidence.
\newblock \emph{Rev. Geophys.}, 58\penalty0 (4):\penalty0 e2019RG000678, 2020.

\bibitem[Siebesma et~al.(2020)Siebesma, Bony, Jakob, and Stevens]{Siebesma2020}
A~Pier Siebesma, Sandrine Bony, Christian Jakob, and Bjorn Stevens.
\newblock \emph{Clouds and climate: Climate science's greatest challenge}.
\newblock 2020.

\bibitem[Subramaniam et~al.(2020)Subramaniam, Wong, Borker, Nimmagadda, and Lele]{Subramaniam2020}
Akshay Subramaniam, Man~Long Wong, Raunak~D Borker, Sravya Nimmagadda, and Sanjiva~K Lele.
\newblock Turbulence enrichment using physics-informed generative adversarial networks, 2020.
\newblock arxiv:2003.01907.

\bibitem[Taylor et~al.(2023)Taylor, Caldwell, Bertagna, Clevenger, Donahue, Foucar, Guba, Hillman, Keen, Krishna, et~al.]{taylor2023simple}
Mark Taylor, Peter~M Caldwell, Luca Bertagna, Conrad Clevenger, Aaron Donahue, James Foucar, Oksana Guba, Benjamin Hillman, Noel Keen, Jayesh Krishna, et~al.
\newblock The simple cloud-resolving e3sm atmosphere model running on the frontier exascale system.
\newblock In \emph{Proceedings of the International Conference for High Performance Computing, Networking, Storage and Analysis}, pages 1--11, 2023.

\bibitem[Tebaldi et~al.(2021)Tebaldi, Debeire, Eyring, Fischer, Fyfe, Friedlingstein, Knutti, Lowe, O'Neill, Sanderson, van Vuuren, Riahi, Meinshausen, Nicholls, Tokarska, Hurtt, Kriegler, Lamarque, Meehl, Moss, Bauer, Boucher, Brovkin, Byun, Dix, Gualdi, Guo, John, Kharin, Kim, Koshiro, Ma, Olivi\'e, Panickal, Qiao, Rong, Rosenbloom, Schupfner, S\'ef\'erian, Sellar, Semmler, Shi, Song, Steger, Stouffer, Swart, Tachiiri, Tang, Tatebe, Voldoire, Volodin, Wyser, Xin, Yang, Yu, and Ziehn]{Tebaldi2021esd}
C.~Tebaldi, K.~Debeire, V.~Eyring, E.~Fischer, J.~Fyfe, P.~Friedlingstein, R.~Knutti, J.~Lowe, B.~O'Neill, B.~Sanderson, D.~van Vuuren, K.~Riahi, M.~Meinshausen, Z.~Nicholls, K.~B. Tokarska, G.~Hurtt, E.~Kriegler, J.-F. Lamarque, G.~Meehl, R.~Moss, S.~E. Bauer, O.~Boucher, V.~Brovkin, Y.-H. Byun, M.~Dix, S.~Gualdi, H.~Guo, J.~G. John, S.~Kharin, Y.~Kim, T.~Koshiro, L.~Ma, D.~Olivi\'e, S.~Panickal, F.~Qiao, X.~Rong, N.~Rosenbloom, M.~Schupfner, R.~S\'ef\'erian, A.~Sellar, T.~Semmler, X.~Shi, Z.~Song, C.~Steger, R.~Stouffer, N.~Swart, K.~Tachiiri, Q.~Tang, H.~Tatebe, A.~Voldoire, E.~Volodin, K.~Wyser, X.~Xin, S.~Yang, Y.~Yu, and T.~Ziehn.
\newblock {Climate model projections from the Scenario Model Intercomparison Project (ScenarioMIP) of CMIP6}.
\newblock \emph{Earth System Dynamics}, 12\penalty0 (1):\penalty0 253--293, 2021.
\newblock \doi{10.5194/esd-12-253-2021}.
\newblock URL \url{https://esd.copernicus.org/articles/12/253/2021/}.

\bibitem[Wang et~al.(2022{\natexlab{a}})Wang, Yuval, and O’Gorman]{Wang2022b}
Peidong Wang, Janni Yuval, and Paul~A O’Gorman.
\newblock Non-local parameterization of atmospheric subgrid processes with neural networks.
\newblock \emph{J. Adv. Model. Earth Syst.}, 14\penalty0 (10):\penalty0 e2022MS002984, 2022{\natexlab{a}}.

\bibitem[Wang et~al.(2022{\natexlab{b}})Wang, Han, Xue, Yang, and Zhang]{Wang2022a}
Xin Wang, Yilun Han, Wei Xue, Guangwen Yang, and Guang~J Zhang.
\newblock Stable climate simulations using a realistic general circulation model with neural network parameterizations for atmospheric moist physics and radiation processes.
\newblock \emph{Geosci. Model Dev.}, 15\penalty0 (9):\penalty0 3923--3940, 2022{\natexlab{b}}.

\bibitem[Watson-Parris et~al.(2022)Watson-Parris, Rao, Olivi{\'e}, Seland, Nowack, Camps-Valls, Stier, Bouabid, Dewey, Fons, Gonzalez, Harder, Jeggle, Lenhardt, Manshausen, Novitasari, Ricard, and Roesch]{Watson-Parris2022}
D~Watson-Parris, Y~Rao, D~Olivi{\'e}, {\O}~Seland, P~Nowack, G~Camps-Valls, P~Stier, S~Bouabid, M~Dewey, E~Fons, J~Gonzalez, P~Harder, K~Jeggle, J~Lenhardt, P~Manshausen, M~Novitasari, L~Ricard, and C~Roesch.
\newblock Climatebench v1.0: A benchmark for data-driven climate projections.
\newblock \emph{J. Adv. Model. Earth Syst.}, 14\penalty0 (10):\penalty0 e2021MS002954, 2022.

\bibitem[Watt-Meyer et~al.(2023)Watt-Meyer, Dresdner, McGibbon, Clark, Henn, Duncan, Brenowitz, Kashinath, Pritchard, Bonev, et~al.]{watt2023ace}
Oliver Watt-Meyer, Gideon Dresdner, Jeremy McGibbon, Spencer~K Clark, Brian Henn, James Duncan, Noah~D Brenowitz, Karthik Kashinath, Michael~S Pritchard, Boris Bonev, et~al.
\newblock Ace: A fast, skillful learned global atmospheric model for climate prediction.
\newblock \emph{arXiv preprint arXiv:2310.02074}, 2023.

\bibitem[Wong-Toi et~al.(2023)Wong-Toi, Boyd, Fortuin, and Mandt]{wongtoi2023understanding}
Eilliot Wong-Toi, Alex Boyd, Vincent Fortuin, and Stephan Mandt.
\newblock Understanding pathologies of deep heteroskedastic regression, 2023.
\newblock arxiv:2306.16717.

\bibitem[Xing et~al.(2021)Xing, Lapeyre, Jaravel, and Poinsot]{Xing2021}
Victor Xing, Corentin Lapeyre, Thomas Jaravel, and Thierry Poinsot.
\newblock Generalization capability of convolutional neural networks for progress variable variance and reaction rate subgrid-scale modeling.
\newblock \emph{Energies}, 14\penalty0 (16):\penalty0 5096, 2021.

\bibitem[Yang et~al.(2022)Yang, Kissas, and Perdikaris]{Yang2022}
Yibo Yang, Georgios Kissas, and Paris Perdikaris.
\newblock Scalable uncertainty quantification for deep operator networks using randomized priors.
\newblock \emph{Comput. Methods Appl. Mech. Eng.}, 399:\penalty0 115399, 2022.

\bibitem[Yu et~al.(2024)Yu, Hannah, Peng, Lin, Bhouri, Gupta, L{\"u}tjens, Will, Behrens, Busecke, et~al.]{yu2024climsim}
Sungduk Yu, Walter Hannah, Liran Peng, Jerry Lin, Mohamed~Aziz Bhouri, Ritwik Gupta, Bj{\"o}rn L{\"u}tjens, Justus~C Will, Gunnar Behrens, Julius Busecke, et~al.
\newblock Climsim: A large multi-scale dataset for hybrid physics-ml climate emulation.
\newblock \emph{Advances in Neural Information Processing Systems}, 36, 2024.

\bibitem[Yuval and O’Gorman(2020)]{Yuval2020}
Janni Yuval and Paul~A O’Gorman.
\newblock Stable machine-learning parameterization of subgrid processes for climate modeling at a range of resolutions.
\newblock \emph{Nature Comm.}, 11\penalty0 (1):\penalty0 3295, 2020.

\bibitem[Yuval et~al.(2021)Yuval, O'Gorman, and Hill]{Yuval2021}
Janni Yuval, Paul~A O'Gorman, and Chris~N Hill.
\newblock Use of neural networks for stable, accurate and physically consistent parameterization of subgrid atmospheric processes with good performance at reduced precision.
\newblock \emph{Geophys. Res. Lett.}, 48\penalty0 (6):\penalty0 e2020GL091363, 2021.

\bibitem[Zanna and Bolton(2020)]{Zanna2020}
Laure Zanna and Thomas Bolton.
\newblock Data-driven equation discovery of ocean mesoscale closures.
\newblock \emph{Geophys. Res. Lett.}, 47\penalty0 (17):\penalty0 e2020GL088376, 2020.

\end{thebibliography}

\end{document}

\title{\datasetonline: Supplementary Information}

\author{Sungduk Yu$^{1, 2}$\thanks{Equal contribution}, Zeyuan Hu$^{3,4}$$^*$\thanks{Correspondence: zeyuan\_hu@fas.harvard.edu}, Akshay Subramaniam$^{3}$, Walter Hannah$^5$, Liran Peng$^1$, Jerry Lin$^1$, Mohamed Aziz Bhouri$^6$, \textbf{Ritwik Gupta$^7$, Björn Lütjens$^{8}$, Justus C. Will$^1$, Gunnar Behrens$^9$, Julius J. M. Busecke$^6$, Nora Loose$^{10}$, Charles I Stern$^6$} \textbf{Tom Beucler$^{11}$, Bryce Harrop$^{12}$, Helge Heuer$^9$, Benjamin R Hillman$^{13}$, Andrea Jenney$^{14}$, Nana Liu$^{15}$, Alistair White$^{16, 17}$,} \textbf{Tian Zheng$^6$, Zhiming Kuang$^{4}$, Fiaz Ahmed$^{18}$, Elizabeth Barnes$^{19}$, Noah D. Brenowitz$^{3}$, Christopher Bretherton$^{20}$,} \textbf{Veronika Eyring$^9$, Savannah Ferretti$^{1}$, Nicholas Lutsko$^{21}$, Pierre Gentine$^6$,} \textbf{Stephan Mandt$^1$, J. David Neelin$^{18}$, Rose Yu$^{21}$, Laure Zanna$^{22}$,} \textbf{Nathan Urban$^{23}$, Janni Yuval$^{24}$, Ryan Abernathey$^6$, Pierre Baldi$^1$, Wayne Chuang$^{6}$,} \textbf{Yu Huang$^6$, Fernando Iglesias-Suarez$^{25}$, Sanket Jantre$^{23}$, Po-Lun Ma$^{12}$, Sara Shamekh$^{22}$, Guang Zhang$^{21}$,} \textbf{Michael Pritchard$^{1, 3}$}
\\
\\
$^1$UCI, $^{2}$Intel Labs, $^{3}$NVIDIA, $^{4}$Harvard, $^5$LLNL, $^6$Columbia, $^7$UCB, $^8$MIT, $^9$DLR, $^{10}$Princeton, $^{11}$UNIL, $^{12}$PNNL, $^{13}$SNL, $^{14}$OSU, $^{15}$CIWRO/NOAA, $^{16}$PIK, $^{17}$TUM, $^{18}$UCLA, $^{19}$CSU, $^{20}$Allen AI, $^{21}$UCSD, $^{22}$NYU, $^{23}$BNL, $^{24}$Google Research, $^{25}$PRED}

\editor{}

\maketitle

{\hypersetup{linkcolor=black}
\tableofcontents}

% ======================================================
\section{Climate Simulations}
% ======================================================

Climate models divide the Earth's atmosphere, land surface, and ocean into a 3D grid, creating a discretized representation of the planet. Somewhat like a virtual Lego construction of Earth, with each brick representing a small region (grid cell). Earth system models are made up of independent component models for the atmosphere, land surface, rivers, ocean, sea ice, and glaciers. Each of these component models is developed independently and can run by itself when provided with the appropriate input data. When running as a fully coupled system the ``component coupler'' handles the flow of data between the components. 

Within each grid cell of the component models, a series of complex calculations are performed to account for various physical processes, such as phase changes of water, radiative heat transfer, and dynamic transport (referred to as ``advection''). Each component model uses the discretized values of many quantities (such as temperature, humidity, and wind speed) as inputs to parameterizations and fluid solvers to output those same values for a future point in time.

The atmosphere and ocean components are the most expensive pieces of an Earth system model, which is largely due to the computation and inter-process communication associated with their fluid dynamics solvers. Furthermore, a significant portion of the overall cost is attributed to the atmospheric physics calculations that are performed locally within each grid column. It is important to note that atmospheric physics serves as a major source of uncertainty in climate projections, primarily stemming from the challenges associated with accurately representing cloud and aerosol processes.

% ===============================
\subsection{Model Description}
\label{model_description}
% ===============================

The data that comprise \dataset{} are from simulations with the Energy Exascale Earth System Model-Multiscale Modeling Framework version 2.1.0 (E3SM-MMF v2) \citep{E3SM2023}. Traditionally, global atmospheric models parameterize clouds and turbulence using crude, low-order models that attempt to represent the aggregate effects of these processes on larger scales. However, the complexity and nonlinearity of cloud and rainfall processes make them particularly challenging to represent accurately with parameterizations. The MMF approach replaces these conventional parameterizations with a cloud resolving model (CRM) in each cell of the global grid, so that cloud and turbulence can be explicitly represented. Each of these independent CRMs is spatially fixed and exchange coupling tendencies with a parent global grid column. This novel approach to representing clouds and turbulence can improve various aspects of the simulated climate, such as rainfall patterns \citep{Kooperman2016}.

The configuration of E3SM-MMF used here shares some details with E3SMv2. The dynamical code of E3SM uses a spectral element approach on a cubed-sphere geometry. Physics calculations are performed on an unstructured, finite-volume grid that is slightly coarser than the dynamics grid, following \cite{Hannah2021}, which is better aligned with the effective resolution of the dynamics grid. Cases with realistic topography include an active land model component that responds to atmospheric conditions with the appropriate fluxes of heat and momentum. 

The embedded CRM in E3SM-MMF is adapted from the System for Atmospheric Modeling (SAM) described by \cite{Khairoutdinov2003}. While the CRM does explicitly represent clouds and turbulence, it still cannot represent the smallest scales of turbulence and microphysics, and, therefore, these processes still need to be parameterized within each CRM grid cell. Microphysical processes are parameterized with a single-moment scheme, and sub-grid scale turbulent fluxes are parameterized using a diagnostic Smagorinsky-type closure. Convective momentum transport in the nested CRM is handled using the scalar momentum tracer approach of \cite{Tulich2015}. The CRM uses an internal timestep of 10 seconds, while the global calculations use a timestep of 20 minutes.

Despite recent efforts to accelerate E3SM-MMF with GPUs and algorithmic techniques \citep{Norman2022}, the CRM domain size strongly affects the computational throughput and limits the type of experiments that can be conducted. However, the MMF approach is quite flexible in how the CRM size is specified. E3SM-MMF is typically run with a 2D CRM that neglects one of the horizontal dimensions, and employs relatively coarse grid spacing that cannot represent small clouds. Increasing the size of this 2D domain by adding further columns (more CRM cells) generally improves the realism of the model solution. Reducing the model grid spacing can also improve the model to a certain degree, although the number of columns often needs to be increased to avoid the degradation associated with a small CRM. Ideally, the CRM would always be used in a 3D configuration to fully capture the complex, chaotic turbulence that dictates the life cycle of each individual cloud, but this approach is generally limited to special experiments that can justify the extra computational cost. The simulations for \dataset{} utilize a 2D CRM with 64 columns and 2 km horizontal grid spacing within each grid cell.

The atmospheric component of E3SM uses a hybrid vertical grid that is ``terrain-following'' near the surface, and transitions to be equivalent to pressure levels near the top (e.g., \url{https://www2.cesm.ucar.edu/models/atm-cam/docs/usersguide/node25.html}). The vertical levels are specified to be thin near the surface to help capture turbulent boundary layer processes, and are gradually stretched to be very coarse in the stratosphere. E3SM-MMF uses 60 levels for the global dynamics with a top level around 65 km. The CRM used for atmospheric physics uses 50 levels, ignoring the upper 10 levels, to avoid problems that arise from using the anelastic approximation with very low densities. This does not create any issues, because cloud processes are generally confined to the troposphere where the anelastic approach is valid. The hybrid grid can be converted to pressure levels using Equation \ref{eq:hybridpressure}, where $P_0=100,000$ $\mathrm{Pa}$ is a reference pressure, and $P_s (\mathbf{x},t)$ is the surface pressure which varies in location $\mathbf{x}$ and time $t$:
\begin{equation}
    P_k = A_k P_0 + B_k P_s
    \label{eq:hybridpressure}
\end{equation}
$A_k$ and $B_k$---where the subscript $k$ denotes the index of vertical coordinate---are the fixed, prescribed coefficients that define how the ``terrain-following'' and ``pure pressure'' coordinates are blended to define the hybrid coordinate at each vertical level. $A_k$ and $B_k$ are provided as a part of the dataset with variable names of \texttt{hyam} and \texttt{hyai} or \texttt{hybm} and \texttt{hybi}, depending on whether mid-level or interface values are needed. The third character of the variable names (``a'' and ``b'') in Equation \ref{eq:hybridpressure} denotes $A_k$ and $B_k$ coefficients, respectively. Note that the indexing of the vertical coordinate starts from the top of the atmosphere due to the construct of $A_k$ and $B_k$ coefficients, e.g., $k=0$ for the top and $k=59$ for the surface in E3SM-MMF.

In the E3SM-MMF framework, the sequencing of atmospheric processes can be conceptualized as follows. It starts with a surface coupling step that receives fluxes from the surface component models (i.e., land, ocean, and sea ice). This is followed by a set of relatively inexpensive physics parameterizations that handle processes such as airplane emissions, boundary layer mixing, and unresolved gravity waves. The global dynamics then takes over to evolve the winds and advect tracers on the global grid. Finally, there is another set of physics calculations to handle clouds, chemistry, and radiation, which are relatively expensive. This final physics section is where the embedded CRM of E3SM-MMF is used, and is the ideal target for surrogate model emulation due to its outsized computational expense. Accordingly, this step represents the target of \dataset{}.

One area where E3SM-MMF significantly differs from E3SMv2 is in the treatment of aerosols and chemistry. The embedded CRM in E3SM-MMF predicts the mass of water species (i.e., cloud and rain droplet mass mixing ratios) but does not predict the number concentration (i.e., number of drops per mass of air). One consequence of this limitation is that E3SM-MMF cannot represent complex cloud aerosol interactions that can impact droplet number concentrations and cloud radiative feedbacks. Therefore, E3SM-MMF cannot use the more sophisticated aerosol and chemistry package used by E3SMv2, and instead uses prescribed aerosol and ozone amounts to account for the direct radiative impact of these tracers. Current efforts are addressing this limitation for future versions of E3SM-MMF.

% ===============================
\subsection{Model Configurations}
% ===============================

The simulations used for \dataset{} were performed on the NERSC Perlmutter machine. E3SM-MMF is unique among climate models in that it can leverage hybrid CPU/GPU architectures on machines such as NERSC Perlmutter (\url{https://www.nersc.gov/systems/perlmutter}), which has 4 NVIDIA A100 GPUs per node. All simulations were configured to run with 4 MPI ranks and 16 OpenMP threads per node. The low-resolution (real geography and aquaplanet) cases used 2 nodes, and the high-resolution (real geography) case used 32 nodes. The throughput of these configurations was roughly 11.5 simulated years per day (sypd) for low-resolution cases and 3.3 sypd for the high-resolution case, averaged over multiple batch submissions. The total simulation length in all cases was 10 model years and 2 model months.

Boundary conditions over maritime regions are constrained by prescribed sea surface temperatures and sea ice amount. Various input data are needed for the cases with realistic topography, such as ozone concentrations and sea surface temperatures, which have been generated to reflect a climatological average of the 2005-2014 period. The aquaplanet configuration does not have a land component, but otherwise has similar input requirements using idealized data to produce a climate that is symmetric along lines of constant latitude.

% ======================================================
\section{Dataset and Code Access}
% ======================================================
% ===============================
\subsection{Code Access}
% ===============================

Following NeurIPS Dataset and Benchmark Track guidelines, we have uploaded our datasets to Hugging Face:
\begin{itemize}
    \item {E3SM-MMF High-Resolution Real Geography dataset: \newline \url{https://huggingface.co/datasets/LEAP/ClimSim_high-res}}
    \item {E3SM-MMF Low-Resolution Real Geography dataset: \newline \url{https://huggingface.co/datasets/LEAP/ClimSim_low-res}}
    \item {E3SM-MMF Low-Resolution Aquaplanet dataset: \newline \url{https://huggingface.co/datasets/LEAP/ClimSim_low-res_aqua-planet}}
\end{itemize}
We have documented all code (including the code to preprocess the data, create, train, and evaluate the baseline models, and visualize data and metrics) in an openly-available GitHub repository: \url{https://leap-stc.github.io/ClimSim}. The containerized workflow of running hybrid simulations can be found at \url{https://github.com/leap-stc/climsim-online/}.

% ===============================
\subsection{Variable List}
% ===============================

All variables included in our dataset are listed in Table \ref{tab:allvars}.

\begin{table}[!htbp]
\centering
\small
\resizebox{\textwidth}{!}{\begin{tabular}{cclccl}
\toprule
\textbf{In} & \textbf{Out} & \textbf{Variable} & \textbf{Dimensions} & \textbf{Units} & \textbf{Description} \\
\midrule
\midrule
$\times$ &  & pbuf\_SOLIN & ncol & W/m$^\text{2}$ & Solar insolation \\ \midrule
$\times$ &  & pbuf\_COSZRS & ncol & & Cosine of solar zenith angle \\ \midrule
$\times$ &  & pbuf\_LHFLX & ncol & W/m$^\text{2}$ & Surface latent heat flux \\ \midrule
$\times$ &  & pbuf\_SHFLX & ncol & W/m$^\text{2}$ & Surface sensible heat flux \\ \midrule
$\times$ &  & pbuf\_TAUX & ncol & W/m$^\text{2}$ & Zonal surface stress \\ \midrule
$\times$ &  & pbuf\_TAUY & ncol & W/m$^\text{2}$ & Meridional surface stress \\ \midrule
$\times$ &  & pbuf\_ozone & lev, ncol & mol/mol & Ozone volume mixing ratio \\ \midrule
$\times$ &  & pbuf\_N2O & lev, ncol & mol/mol & Nitrous oxide volume mixing ratio \\ \midrule
$\times$ &  & pbuf\_CH4 & lev, ncol & mol/mol & Methane volume mixing ratio \\ \midrule
$\times$ &  & state\_ps & ncol & Pa & Surface pressure \\ \midrule
$\times$ & $\times$ & state\_q0001 & lev, ncol & kg/kg & Specific humidity \\ \midrule
$\times$ & $\times$ & state\_q0002 & lev, ncol & kg/kg & Cloud liquid mixing ratio \\ \midrule
$\times$ & $\times$ & state\_q0003 & lev, ncol & kg/kg & Cloud ice mixing ratio \\ \midrule
$\times$ & $\times$ & state\_t & lev, ncol & K & Air temperature \\ \midrule
$\times$ & $\times$ & state\_u & lev, ncol & m/s & Zonal wind speed \\ \midrule
$\times$ & $\times$ & state\_v & lev, ncol & m/s & Meridional wind speed \\ \midrule
$\times$ &  & state\_pmid & lev, ncol & Pa & Mid-level pressure \\ \midrule
$\times$ &  & cam\_in\_ASDIR & ncol & & Albedo for direct shortwave radiation \\ \midrule
$\times$ &  & cam\_in\_ASDIF & ncol & & Albedo for diffuse shortwave radiation \\ \midrule
$\times$ &  & cam\_in\_ALDIR & ncol & & Albedo for direct longwave radiation \\ \midrule
$\times$ &  & cam\_in\_ALDIF & ncol & & Albedo for diffuse longwave radiation \\ \midrule
$\times$ &  & cam\_in\_LWUP & ncol & W/m$^\text{2}$ & Upward longwave flux \\ \midrule
$\times$ &  & cam\_in\_SNOWHLAND & ncol & m & Snow depth over land (liquid water equivalent) \\ \midrule
$\times$ &  & cam\_in\_SNOWHICE & ncol & m & Snow depth over ice \\ \midrule
$\times$ &  & cam\_in\_LANDFRAC & ncol & & Land area fraction \\ \midrule
$\times$ &  & cam\_in\_ICEFRAC & ncol & & Sea-ice area fraction \\ \midrule
  & $\times$ & cam\_out\_NETSW & ncol & W/m$^\text{2}$ & Net shortwave flux at surface \\ \midrule
  & $\times$ & cam\_out\_FLWDS & ncol & W/m$^\text{2}$ & Downward longwave flux at surface \\ \midrule
  & $\times$ & cam\_out\_PRECSC & ncol & m/s & Snow rate (liquid water equivalent) \\ \midrule
  & $\times$ & cam\_out\_PRECC & ncol & m/s & Rain rate \\ \midrule
  & $\times$ & cam\_out\_SOLS & ncol & W/m$^\text{2}$ & Downward visible direct solar flux to surface \\ \midrule
  & $\times$ & cam\_out\_SOLL & ncol & W/m$^\text{2}$ & Downward near-IR direct solar flux to surface \\ \midrule
  & $\times$ & cam\_out\_SOLSD & ncol & W/m$^\text{2}$ & Downward visible diffuse solar flux to surface \\ \midrule
  & $\times$ & cam\_out\_SOLLD & ncol & W/m$^\text{2}$ & Downward near-IR diffuse solar flux to surface \\
\bottomrule
\end{tabular}}
\vspace{1mm}
\caption{Overview of input variables (first column) and output variables (second column) of the E3SM-MMF physics calculations (including the CRM) that are stored in \dataset{}. The other columns indicate the variable name, dimensions, units, and a brief description. IR is short for infrared, which is also often referred to as ``longwave'' radiation among atmospheric scientists.}
\label{tab:allvars}
\end{table}

% ===============================
\subsection{Dataset Statistics}
% ===============================

Here, we present some distribution statistics to aid in understanding the dataset. Detailed distributions for all variables are provided in \url{https://github.com/leap-stc/ClimSim/tree/main/dataset_statistics}. These statistics are calculated for each vertical level individually for the vertically-resolved variables (e.g., \texttt{state\_t} and \texttt{state\_q0001}). For each variable (additionally, at each level for the vertically-resolved variables), a histogram is provided to visualize the distribution using 100 bins. Additionally, a text file accompanies each histogram, containing key statistical measures such as the mean, standard deviation, skewness, kurtosis, median, deciles, quartiles, minimum, maximum, and mode. The text file also includes the bin edges and the corresponding frequency values used to generate the histogram figures. This comprehensive approach allows for a detailed analysis of the dataset's distributions.

% ===============================
\subsection{Dataset Applications}
% ===============================

Our data can benefit a broader audience beyond climate modelers wishing to explore ML for sub-grid  parameterization. For climate studies, while high-frequency timestep-level outputs from simulations are rarely archived, they offer insights into convective extremes and diurnal variability. Such data opens the path to  explore multi-scale interactions between rapid dynamics and broader weather and climate fluctuations. This includes a detailed examination of variables needed to constrain vertically resolved energy and water budgets and understand their variability. For the machine learning community, this dataset addresses the scarcity of large-scale regression benchmarks, common in the sciences. Such benchmarks are less common compared to prevalent industrial datasets that emphasize classification, computer vision, and NLP tasks.

% ===============================
\subsection{Target Audiences}
% ===============================

In essence, this benchmark aims to democratize and expand access to advanced climate modeling. High-potential architectures will undergo testing in the superparameterized version of the DOE's primary climate model, E3SM. Successful integration would substantially reduce computational costs for the DOE when contemplating the deployment of MMF technology in climate prediction. E3SM's external user community, typically deterred by the extensive computational demands of superparameterized simulators, also stands to benefit. Currently, only a minority with substantial computing resources can engage with such models. A successful recipe for ClimSim could thus democratize the use of explicit convection for a broader user base. If performant architectures also prove effective in the NCAR Community Earth System Model (CESM) - the world’s most widely used open source climate simulator - the user base could expand significantly. Given its software similarities to E3SM, it is logical to expect that ClimSim's learnt parameterizations will be readily adaptable to CESM. Moreover, we anticipate that a successful hybrid machine learning climate simulator will bring benefits to a diverse range of industry sectors, including those vulnerable to climate risks (such as agriculture, energy, and tourism), as well as the climate risk industry itself (such as insurance and risk assessment). 
% ======================================================
\section{Baseline Models}
% ======================================================

This section offers a detailed depiction of six baseline models. Every facet of model designs, excluding the dimensions of the input and output layers, differs among the models. We recognize that while this approach maximizes the differentiation among baseline models, such extensive degrees of freedom complicate the complete isolation of the effects arising from optimization parameter choices and those originating from the model architecture itself. In future ClimSim releases, baseline models will share more constraints (including optimization parameters) to highlight the performance difference due to model architectures.

% ===============================
\subsection{Multilayer Perceptron (MLP)}
\label{MLP}
% ===============================

A multilayer perceptron (MLP) is a basic, densely connected artificial neural network. We used KerasTuner \citep{O'Malley2019} with a random search algorithm for hyperparameter optimization. The following hyperparameters were optimized: the number of hidden layers (\textit{N}$_\text{layers}$), the number of nodes per layer (\textit{N}$_\text{nodes}$), activation function, and batch size. The search domains were:
\begin{itemize}
    \item \textit{N}$_\text{layers}$: [3, 4, 5, 6, 7, 8, 9, 10, 11, 12, 13]
    \item \textit{N}$_\text{nodes}$: [128, 256, 384, 512, 640, 768, 896, 1024]
    \item Activation function: [ReLU, LeakyReLU ($\alpha = 0.15$), eLU ($\alpha = 1.0$)]
    \item Batch size: [48, 96, 192, 384, 768, 1152, 1536, 2304, 3072]
    \item Optimizer: [Adam, RAdam, RMSprop, SGD]
\end{itemize}
Note that \textit{N}$_\text{nodes}$ was selected independently for each hidden layer. For example, for \textit{N}$_\text{layers} = k$, \textit{N}$_\text{nodes}$ was drawn from the search domain $k$ times. The width of the last hidden layer was fixed at 128. The output layer utilized the linear activation function for the first 120 outputs (corresponding to the heating and moistening tendencies), and ReLU for the remaining 8 variables (corresponding positive-definite surface variables). The loss function was taken as the mean squared error (MSE), and the learning rate was defined using a cyclic scheduler, with an initial learning rate of 2.5 $\times$ 10$^\text{-4}$, maximum of 2.5 $\times$ 10$^\text{-3}$, and step size of 4 epochs.

\begin{figure}[!htbp]
    \centering
    \includegraphics[width=.7\textwidth]{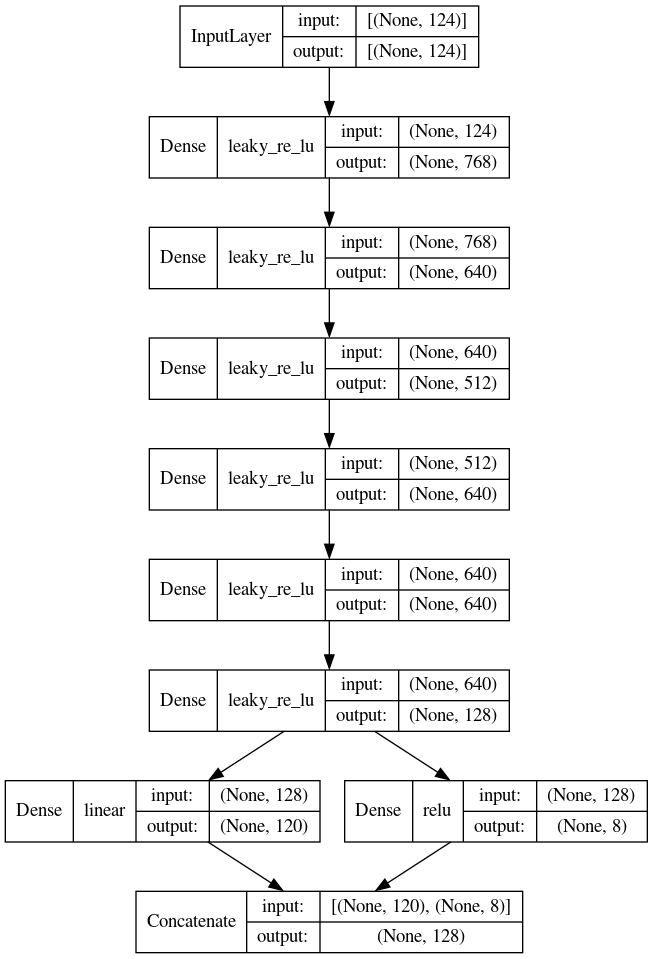}
    \caption{The architecture of the MLP baseline model.}
    \label{fig:mlpdiagram}
\end{figure}

Following \cite{Yu2023}, we conducted the hyperparameter search in two stages. In the first stage, a total of 8,257 randomly-drawn hyperparameter configurations were trained and evaluated with a tiny subset of the full training set, sub-sampled in the time dimension with a stride of 37. In the second stage, the top 0.2\% candidates (160 hyperparameter configurations) were re-trained with a larger fraction of the full training set (sub-sampled with a stride of 7), and then evaluated for our MLP baseline. After this two-step search process, the best hyperparameter configuration was identified as: \textit{N}$_\text{layers} =$ 5, \textit{N}$_\text{nodes} =$ [768, 640, 512, 640, 640], LeakyReLU activation, a batch size of 3,072, and RAdam optimizer. The MLP baseline has approximately 1.75 million parameters and executes 3.50 MFlops on one data point, the architecture of which is summarized in Figure \ref{fig:mlpdiagram}.

To provide some context on the amount of variance in model performance that can be attributed to random effects of optimization, the top 160 models were selected from our pool of 8,257 trials and scored on the validation set; the 5th to 95th percentile range of this ensemble is shown by the error bars in Figures 2a and SI3, and by the grey shading in Figures 2b-e, SI4, and SI5.

\textbf{MLP with expanded features and targets:} 
We built MLP with an expanded set of input and output variables, as elaborated in Section 4.2 of the main text. For the sake of clarity, we designate an MLP model employing the subset of available variables (outlined in Section 4 of the main text) as "MLPv1," while an MLP model utilizing the expanded variables is referred to as "MLPv2." The hyperparameter optimization for MLPv2 followed a similar process as MLPv1, with the exception that the search domain of batch size was defined as [2700, 5400, 10800, 21600, 43200, 64800, 86400, 129600, 172800]. After 11,851 search trials, the best hyperparameter configuration was identified as: \textit{N}$_\text{layers} =$ 3, \textit{N}$_\text{nodes} =$ [384, 1024, 640], ReLU activation, a batch size of 2,304, and Adam optimizer. The MLPv2 baseline has approximately 1.59 million parameters and executes 3.17 MFlops on one data point.

\textbf{MLP with the high-resolution dataset:}
In conjunction with the MLP featuring expanded features and targets, we also constructed MLP models using the high-resolution dataset for both MLPv1 and MLPv2. To differentiate these models from those constructed with the low-resolution dataset, we add the suffix "-ne30" to their names. The hyperparameters for MLPv1-ne30 and MLPv2-ne30 were optimized using the same methodology as was applied to their low-resolution counterparts. For MLPv1-ne30, after 10,296 search trials, the best hyperparameter configuration was identified as: \textit{N}$_\text{layers} =$ 4, \textit{N}$_\text{nodes} =$ [1024, 128, 128, 768], leaky ReLU activation, a batch size of 5,400, and Adam optimizer. The MLPv1-ne30 baseline has approximately 0.49 million parameters and executes 0.98 MFlops on one data point. For MLPv2-ne30, after 10,440 search trials, the best hyperparameter configuration was identified as: \textit{N}$_\text{layers} =$ 3, \textit{N}$_\text{nodes} =$ [640, 128, 1024], ReLU activation, a batch size of 2,700, and Adam optimizer. The MLPv2-ne30 baseline has approximately 1.00 million parameters and executes 2.00 MFlops on one data point.

The model performance comparison between MLPv1, MLPv2, MLPv1-ne30, and MLPv2-ne30 is presented in SI Section 8.1.

% ===============================
\subsection{Randomized Prior Network (RPN)}
% ===============================

A randomized prior network (RPN) is an ensemble model \citep{Osband2018}. Each member of the RPN is built as the sum of a trainable and a non-trainable (so-called ``prior'') surrogate model; we used MLP for simplicity. Multiple replicas of the networks are constructed by independent and random sampling of both trainable and non-trainable parameters \citep{Yang2022, Bhouri2023}. RPNs also resort to data bootstrapping in order to mitigate the uncertainty collapse of the ensemble method when tested beyond the training data points \citep{Bhouri2023}. Data bootstrapping consists of sub-sampling and randomization of the data each network in the ensemble sees during training. Hyperparameters of individual MLPs (i.e., \textit{N}$_\text{layers}$, \textit{N}$_\text{nodes}$, batch size) did not need to be tuned from scratch, and were instead chosen based on the hyperparameter search mentioned in Section \ref{MLP}. RPN ensembles of 128 networks were considered justified \citep{Yang2022}.

In particular, individual MLPs forming the RPN were considered as fully connected neural networks with \textit{N}$_\text{layers} =$ 5, \textit{N}$_\text{nodes} =$ [768, 640, 512, 640, 640], and a batch size of 3,072, as in Section \ref{MLP}. We utilized ReLU activation (with a negative slope of 0.15) for all layers except for the output layer, where the linear activation function was used.

The MLPs were trained for a total of 13,140 stochastic gradient descent (SGD) steps using the Adam optimizer. The learning rate was initialized at 5 $\times$ 10$^\text{-4}$ with an exponential decay at a rate of 0.99 for every 1,000 steps. The RPN baseline has approximately 222.3 million parameters ($\sim$1.74 million per MLP) and executes 0.89 GFlops on one data point. 

% ===============================
\subsection{Convolutional Neural Network (CNN)}
% ===============================

The convolutional neural network (CNN) used is a modified version of a residual network (ResNet). Each ResNet block is composed of two, 1D convolutions (Conv1D) with a 3 $\times$ 3 kernel using ``same'' padding, and an output feature map size of 406. Each Conv1D is followed by ReLU activation and dropout (with rate = 0.175). Residuals were also 1D convolved using a 1 $\times$ 1 kernel, and added back to the output of the main ResNet block.

The CNN composes 12 such ResNet blocks, followed by ``flattening'' of the feature map via a 1 $\times$ 1 convolution and eLU activation. Two separate Dense layers (and their corresponding activations) map the output feature map to their respective co-domains: one to $(-\infty, \infty)$ assuming that vertically-resolved variables have no defined range, and another to $[0, \infty)$ for all globally-resolved variables. These were concatenated as the output of the network.

A hyperparameter search was conducted on depth, width, kernel size, activation functions, loss functions, and normalization types using the Hyperband~\citep{Li2018} strategy with the KerasTuner \citep{O'Malley2019} framework. The search domains were:
\begin{itemize}
    \item Model depth/number of ResNet blocks: [2, 15]
    \item Model width: [32, 512]
    \item Kernel width: [3, 5, 7, 9]
    \item Activation function: [GeLU, eLU, ReLU, Swish]
    \item Layer normalization: [True, False]
    \item Dropout: [0.0, 0.5]
    \item Optimizer: [SGD, Adam]
\end{itemize}
The CNN was trained for 10 epochs with an Adam optimizer with standard hyperparameters ($\beta_1 = 0.9$, $\beta_2 = 0.999$, $\epsilon = 1 \times 10^{-7}$). The learning rate was defined using a cyclic scheduler, with an initial learning rate of 1 $\times$ 10$^\text{-4}$, a maximum of 1 $\times$ 10$^\text{-3}$, and a step size of 2 $\times \lfloor \frac{\text{10,091,520}}{\text{12}}\rfloor$. A scaling function of $\frac{1}{2.0^{x - 1}}$ was applied to the scheduler per step $x$.

The hyperparameter search was conducted for 12 hours on 8 NVIDIA Tesla V100 32GB cards, with one model executing on each card. A weighted mean absolute error (MAE) was used as the loss function for optimization. We down-weighted the standard MAE loss to de-emphasize repeated scalar values provided to the network as input. The weighted MAE function is defined below:
\begin{python}
def mae_adjusted(y_true, y_pred):
    abs_error = K.abs(y_pred - y_true)
    vertical_weights = K.mean(abs_error[:,:,0:2])*(120/128)
    scalar_weights   = K.mean(abs_error[:,:,2:10])*(8/128)
    return vertical_weights + scalar_weights
\end{python}
The CNN baseline has approximately 13.2 million parameters and executes 1.59 GFlops on one data point. The architecture is visualized below in Figure \ref{fig:cnndiagram}.

\begin{figure}[!htbp]
    \centering
    \includegraphics[width=\textwidth]{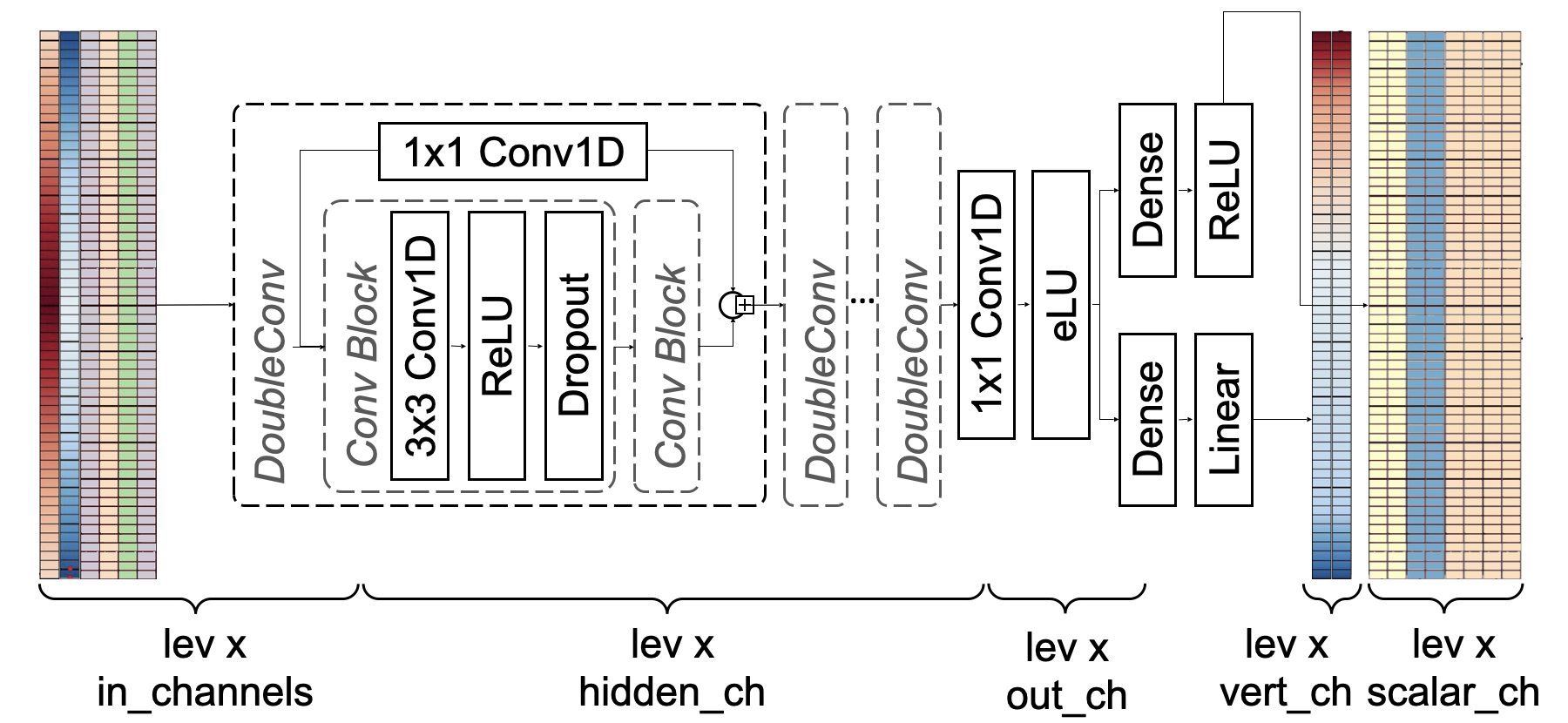}
    \caption{The ResNet-style CNN baseline is comprised of multiple ResNet blocks (i.e., DoubleConv), and applies different activation to the outputs for vertically-resolved and global variables. The channel dimensions are [in\_channels, hidden\_ch, out\_ch, vert\_ch, scalar\_ch] = [6, 406, 10, 8, 2].}
    \label{fig:cnndiagram}
\end{figure}

% ===============================
\subsection{Heteroskedastic Regression (HSR)}
% ===============================

\newcommand{\E}{\mathbb{E}}
\newcommand{\x}{\mathbb{\mathbf{x}}}
\newcommand{\y}{\mathbb{\mathbf{y}}}
\newcommand{\z}{\mathbb{\mathbf{z}}}
\newcommand{\norm}[1]{\left\Vert #1 \right\Vert}

We quantified the inherent stochasticity in the data $\mathcal{D} = \{(\x_1, \y_1), \dots, (\x_n, \y_n)\}$, and the uncertainty in our prediction by providing a distributional prediction instead of a point estimate. In hetereoskedastic regression (HSR), this predictive distribution is modeled explicitly; here as independent Gaussians with unique mean $\mu_k$ and precision (inverse variance) $\tau_k$ for each variable. We assumed
\begin{align*}
    \y_i \,|\, \x_i &\sim \mathcal{N}(\mu(\x_i), \mathrm{Diag}(\tau(\x_i)^{-1})),
\end{align*}
and parameterized both $\mu$ and $\tau$ as over-parameterized feed-forward neural networks (i.e., MLPs) $\hat{\mu}_\theta(\x)$ and $\hat{\tau}_\phi(\x)$, respectively. This yielded the corresponding predictive distribution
\begin{align*}
    \hat{\y}_i \,|\, \x_i \sim \mathcal{N}(\hat{\mu}_\theta(\x_i), \mathrm{Diag}(\hat{\tau}_\theta(\x_i)^{-1})),
\end{align*}
which was fitted with maximum likelihood estimation (MLE) by minimizing the objective
\begin{align*}
    \mathcal{L}(\theta, \phi) = \frac{1}{2n} \sum_{i=1}^n \left[ \norm{\hat{\tau}_\phi(\x_i) \left(\y_i - \hat{\mu}_\theta(\x_i)\right)}_2^2
    - \mathbf{1}^T \log \left(\hat{\mu}_\theta(\x_i) \right) \right].
\end{align*}
Note that, due to the flexibility of the neural networks, this formulation is ill-posed. It may lead to cases of extreme overfitting where $\hat{\tau}_\phi(\x_i) \approx \y_i$, $\hat{\tau}_\phi(\x_i) \approx \mathbf{0}$, thus making $\mathcal{L}(\theta, \phi)$ completely unstable. Hence, we instead minimized a modified objective that included L2-regularization via
\begin{align*}
    \mathcal{L}_{\rho, \gamma}(\theta, \phi) \coloneqq \rho \mathcal{L}(\theta, \phi) + (1 - \rho) \left[\gamma \norm{\theta}_2^2 + (1-\gamma) \norm{\phi}_2^2 \right],
\end{align*}
where $\rho, \gamma \in (0, 1)$ determines the trade-off between MLE estimation, mean regularization, and precision regularization. We follow \cite{wongtoi2023understanding} and set $\rho = 1 - \gamma$ to reduce the hyperparameter search domain. 

Specifically, we used two MLPs with layer normalization and ReLU activation, and trained them with gradient-based stochastic optimization. To improve stability, the first third of training was spent on exclusively training $\hat{\mu}_\theta(\x_i)$ with an MSE loss. To optimize hyperparameters, we selected a configuration from 300 trials with a random number of \textit{N}$_\text{layers} =$ [2, 3, 4], \textit{N}$_\text{nodes} =$ [256, 512, 1,024, 2,048], $\gamma$ (log-uniform in [0.001, 0.1]), optimizer $=$ [SGD, Adam] with hyperparameters ($\beta_1 = 0.9$, $\beta_2 = 0.999$), learning rate $\lambda$ (log-uniform in [10$^\text{-6}$, 10$^\text{-3}$]), and batch size = [1024, 2048, 4096, 8192, 16384]. Each run was trained for 12 epochs total on one NVIDIA GeForce RTX 4080 16GB. We chose the run with the lowest CRPS on the validation data, yielding \textit{N}$_\text{layers} =$ 4, \textit{N}$_\text{nodes} =$ 1,024, $\gamma = 2.2 \times 10^{-2}$, $\lambda = 7 \times 10^{-6}$, and a batch size of 16,384, trained with Adam. The HSR baseline has approximately 6.63 million parameters and executes 6.85 MFlops per data point.

% ===============================
\subsection{Conditional Variational Autoencoder (cVAE)}
% ===============================

A conditional generative latent variable model first samples---from a prior $p(\z)$---a point $\z$ in a low-dimensional latent space, which then informs a conditional distribution $p_\theta(\y|\z, \x)$ over the target domain. This allows for a complex and flexible predictive distribution. In our case, we used feed-forward neural networks (i.e., MLPs) $\mu_\theta(\z, \x)$ and $\sigma_\theta(\z, \x)$ with combined parameters $\theta$ and model:
\begin{equation}
    \begin{aligned}
        \z &\sim \mathcal{N}(\mathbf{0}, \mathcal{I}) \\
        \y|\z,\x &\sim \mathcal{N}\bigl(\mu_\theta(\z,\x), \mathrm{Diag}(\sigma_\theta(\x)^2)\bigr)
        \label{eq:cvae}
    \end{aligned}
\end{equation}
To fit the model to data $\mathcal{D} = \{(\x_1, \y_1), \dots, (\x_n, \y_n)\}$, we minimized the negative evidence lower bound (NELBO) $\mathcal{L}_\theta(\mathbf{q})$ that bounds the intractable negative marginal likelihood from above via
\begin{align*}
    \mathcal{L}_\theta(q) \coloneqq - \mathbb{E}_{\z_i \sim q} \left[\log\frac{p_\theta(\y_i, \z_i | \x_i)}{q(\z_i|\x_i)}\right] = - \log p_\theta(\y_i|\x_i) +\, \underbrace{\mathrm{KL}\bigl(q \,\|\, p_\theta(\z_i|\y_i, \x_i)\bigr)}_{\geq 0},
\end{align*}
using an approximation $q$ to the posterior $p_\theta(\z_i|\y_i, \x_i)$. The conditional variational autoencoder (cVAE) \citep{Kingma2014} uses amortized variational inference to optimize $\theta$ and $q$ jointly by approximating the latter with e.g., $q_\psi(\z_i) = \mathcal{N}\bigl(g_\psi(\x_i), \mathrm{Diag}(h_\psi(\x_i)^2)\bigr)$, where we again chose $g_\psi(\x_i)$ and $h_\psi(\x_i)$ to be MLPs. This allowed us to optimize for $\theta$ and $\psi$ by minimizing
\begin{align*}
    \mathcal{L}_\theta(q) &\stackrel{\beta=1}{=} %\E_{z\sim q}[-\log p_\theta(y|z, x)] + \mathrm{KL}(q \,\|\, p_\theta(z|x)) \\
    \mathbb{E}_{\z_i\sim q_\psi}\left[ \frac{1}{2} \norm{\frac{\y_i - \mu_\theta(\z_i, \x_i)}{\sigma_\theta(\z_i, \x_i)}}_2^2
    + \mathbf{1}^T \log \left(\sigma_\theta(\z_i, \x_i)\right)\right] + \beta\mathrm{KL}(q_\psi(\z_i) \,\|\, p(\z_i)) + \mathrm{const}
\end{align*}
with a Monte Carlo approximation by first sampling $\z_i$ (once) from the variational encoder $q_\psi(\z_i)$. After which, we decoded the predictive mean and standard deviation with the variational decoder $\mu_\theta(\z, \x)$ and $\sigma_\theta(\z, \x)$. We then computed NELBO as a sum of a reconstruction term and a KL term that regularizes the latent space, averaged over all samples, and back-propagated the gradients. By letting $\beta$ be a hyperparameter, we manually determined the trade-off between reconstruction quality and latent space structure. Finally, at inference time, we used Equation \ref{eq:cvae} to sample from the predictive distribution
\begin{align*}
    p_\theta(\hat{\y}|\x) = \int p_\theta(\hat{\y}|\x, \z)p(\z)\,d\z.
\end{align*}
For both the variational encoder and decoder, we used an MLP with layer normalization, ReLU activation, dropout with $p = 0.05$, and two branching final layers that produced the mean and standard deviation, respectively. We trained both MLPs jointly---with gradient-based stochastic optimization---on the objective described above.

To optimize hyperparameters, we ran 300 trials with a random number of hidden layers \textit{N}$_\text{layers} =$ [2, 3, 4], \textit{N}$_\text{nodes} =$ [256, 512, 1024, 2048], size of the latent space $=$ [4, 8, 16, 32], $\beta$ (log-uniform in [0.01, 10]), optimizer = [SGD, Adam] with ($\beta_1 = 0.9$, $\beta_2 = 0.0999$), learning rate $\lambda$ (log-uniform in [10$^\text{-6}$, 10$^\text{-3}$]), L2 regularization $\alpha$ (log-uniform in [10$^\text{-6}$, 10$^\text{-3}$]), and batch size $=$ [1024, 2048, 4096, 8192, 16384]. Each run was trained for 5 epochs total on one NVIDIA GeForce RTX 4080 16GB. The run with the lowest CRPS on the validation data yielded \textit{N}$_\text{layers} =$ 3, \textit{N}$_\text{nodes} =$ 1,024, and a batch size of 4,096, trained with Adam. In a second step, we fixed these hyperparameters and further fine-tuned $\beta$, $\lambda$, and $\alpha$ by training for 20 epochs every time, for 10 trials. We found the best model with $\beta = 0.5$, $\lambda = 5 \times 10^{-5}$, $\alpha = 10^{-3}$. The cVAE baseline has approximately 4.9 million parameters and executes 4.88 MFlops per data point.

% ===============================
\subsection{Encoder-Decoder (ED)}
% ===============================

The Encoder-Decoder (ED) is an adjusted version of the ED presented in \cite{Behrens2022}. We keep all tuneable hyperparameters except for the learning rate and the node sizes of input and output layer of ED fixed to the original values that were optimized with a detailed hyperparameter search for the superparameterization of the Community Atmosphere Model version 3 in an aquaplanet setup \citep{Behrens2022}. The Encoder consists of 6 hidden fully-connected layers. The Encoder decreases progressively the dimensionality of the input variables down to 5 nodes in the latent space of the network. These 5 latent nodes are the only input to the decoding part of ED. The Decoder maps the information from the latent space back to 128 nodes in the output layer through 6 progressively wider fully-connected hidden layers \citep{Behrens2022}. We train ED over 40 epochs with a learning rate step after each 7$^{\text{th}}$ epoch, which reduces the learning rate by factor 5 \citep{Behrens2022}. The adjusted initial learning rate has a value of 1 $\times$ 10$^{-4}$. The batch size has a value of 714 samples. As activation functions of all hidden layers we use ReLU and the output layer of the Decoder is ELU-activated \citep{Behrens2022}. As an optimizer during training we use Adam. As a loss function of ED we use a MSE loss and as additional metric the MAE during training. The following list summarizes the key hyperparameters of ED: 
\begin{itemize}
    \item Learning rate: 1 $\times$ 10$^{-4}$, learning rate decrease after every 7$^{\text{th}}$ epoch
    \item Batch size: 714
    \item Latent space width: 5 Nodes
    \item Encoder node size: [124, 463, 463, 232, 116, 58, 29, 5]
    \item Decoder node size: [5, 29, 58, 116, 232, 463, 463, 128] 
    \item Encoder activation functions: [Input, ReLU, ReLU, ReLU, ReLU, ReLU,
ReLU, ReLU]
    \item Decoder activation functions: [Input, ReLU, ReLU, ReLU, ReLU, ReLU,
ReLU, ELU]
    \item Optimizer: Adam
\end{itemize}
To prevent overfitting we shuffle the training data set before each epoch. ED baseline has approximately 832,000 parameters, with 415,000 parameters in the Encoder and 417,000 parameters in the Decoder. In total, ED executes 1.66 MFlops per data point, with 829 kFLops per data point for the Encoder and 832 kFlops per data point for the Decoder.

% ======================================================
\subsection{Inference Cost}
% ======================================================

\begin{table}[ht]
\centering
\small
\resizebox{\textwidth}{!}{\begin{tabular}{lcccccc}
\toprule
 & CNN & ED & HSR & MLP & RPN & cVAE \\
\midrule
\midrule
Number of Parameters & 13,200,000 & 832,000 & 6,630,000 & 1,750,000 & 222,300,000 & 4,900,000 \\
MFlops Per Data Point & 1590 & 1.66 & 6.85 & 3.50 & 890 &  4.88 \\
\bottomrule
\end{tabular}}
\vspace{1mm}
\setlength{\belowcaptionskip}{-0.5em}%
\caption{The number of learnable parameters and Megaflops (MFlops) per data point for each of the six baseline models.}
\label{tab:costs}
\end{table}

% ======================================================
\section{Baseline Model Evaluations}
% ======================================================
% ===============================
\subsection{Metrics}
% ===============================
% ===============================
\subsubsection{Deterministic Metrics}
% ===============================

Mean Absolute Error (MAE):
\begin{equation}
    \text{MAE} = \frac{1}{n}\sum_{i=1}^{n}|X_i - y|
    \label{eq:mae}
\end{equation}
Root Mean Squared Error (RMSE):
\begin{equation}
    \text{RMSE} = \sqrt{\frac{1}{n}\sum_{i=1}^{n}(X_i - y)^2}
    \label{eq:rmse}
\end{equation}
Coefficient of Determination (R$^\text{2}$):
\begin{equation}
    \text{R$^\text{2}$} = 1 - \frac{\sum_{i=1}^{n}(X_i - y)^2}{\sum_{i=1}^{n}(X_i - \bar{X})^2}
    \label{eq:r2}
\end{equation}
In Equations \ref{eq:mae}--\ref{eq:r2}, $X_i$ and $y$ represent the true and predicted values, respectively. The mean of the true values of the dependent variable is denoted by $\bar{X}$.

% ===============================
\subsubsection{Stochastic Metric (CRPS)}
% ===============================

The continuous ranked probability score (CRPS) is a generalization of the MAE for distributional predictions. CRPS penalizes over-confidence in addition to inaccuracy in ensemble predictions---a lower CRPS is better. For each variable, it compares the ground truth target $y$ with the cumulative distribution function (CDF) $F$ of the prediction via
\begin{align*}
    \mathrm{CRPS}(F, y) &\coloneqq \int \left( F(x) - \mathbf{1}_{\{x \geq y\}} \right)^2 dx \nonumber\\
    &= \mathbb{E}[|X - y|] - \frac{1}{2} \mathbb{E}[|X - X'|],
\end{align*}
where $X, X' \sim F$ are independent and identically distributed ($iid$) samples from the distributional prediction. We use the non-parametric ``fair estimate to the CRPS'' \citep{Ferro2014}, estimating $F$ with the empirical CDF of $n = 32\;iid$ samples $X_i \sim F$:
\begin{align}
    \hat{\mathrm{CRPS}}(\mathbf{X}, y) &\coloneqq \frac{1}{n} \sum_{i=1}^n |X_i - y| - \frac{1}{2 n (n-1)} \sum_{i=1}^n \sum_{j=1}^n |X_i - X_j|
    \label{eq:crps}
\end{align}
The first term in Equation \ref{eq:crps} is the MAE between the target and samples of the predictive distribution, while the second term is small for small predictive variances, vanishing completely for point estimates. Note that this definition extends to ensemble models, where we take the prediction of each ensemble member as a sample of an implicit predictive distribution. 

% \clearpage
\begin{table}[!htbp]
\centering
\small
\resizebox{\textwidth}{!}{\begin{tabular}{l|cccccc|cccccc}
\toprule
\multicolumn{1}{c|}{\multirow{2}{*}{\textbf{Variable}}} & \multicolumn{6}{c|}{\textbf{RMSE [W/m$^\text{2}$]}} & \multicolumn{6}{c}{\textbf{CRPS [W/m$^\text{2}$]}} \\
\cmidrule{2-13}
\multicolumn{1}{c|}{} & CNN & ED & HSR & MLP & RPN & cVAE & CNN & ED & HSR & MLP & RPN & cVAE \\
\midrule
$dT/dt$ & \textbf{4.369} & 4.696 & 4.825 & 4.421 & 4.482 & 4.721 & -- & -- & \textbf{2.158} & -- & 2.305 & 2.708 \\
$dq/dt$ & \textbf{7.284} & 7.643 & 7.896 & 7.322 & 7.518 & 7.780 & -- & -- & \textbf{3.645} & -- & 4.100 & 4.565 \\
NETSW & 36.91 & 28.537 & 37.77 & \textbf{26.71} & 33.60 & 38.36 & -- & -- & \textbf{14.62} & -- & 14.82 & 20.53 \\
FLWDS & 10.86 & 9.070 & 8.220 & \textbf{6.969} & 7.914 & 8.530 & -- & -- & 4.561 & -- & \textbf{4.430} & 6.732 \\
PRECSC & 6.001 & 5.078 & 6.095 & \textbf{4.734} & 5.511 & 6.182 & -- & -- & 2.905 & -- & \textbf{2.729} & 3.513 \\
PRECC & 85.31 & 76.682 & 90.64 & \textbf{72.88} & 76.58 & 88.71 & -- & -- & 34.30 & -- & \textbf{30.08} & 40.17 \\
SOLS & 22.92 & 17.999 & 23.61 & \textbf{17.40} & 20.61 & 23.27 & -- & -- & 8.369 & -- & \textbf{8.309} & 11.91 \\
SOLL & 27.25 & 22.540 & 27.78 & \textbf{21.95} & 25.22 & 27.81 & -- & -- & \textbf{10.14} & -- & 10.49 & 14.42 \\
SOLSD & 12.13 & 9.917 & 12.40 & \textbf{9.420} & 11.00 & 12.64 & -- & -- & 4.773 & -- & \textbf{4.649} & 5.945 \\
SOLLD & 12.10 & 10.417 & 12.47 & \textbf{10.12} & 11.25 & 12.63 & -- & -- & \textbf{4.599} & -- & 4.682 & 5.925 \\
\bottomrule
\end{tabular}}
\vspace{1mm}
\setlength{\belowcaptionskip}{-0.25em}%
\caption{Globally-averaged RMSE and CRPS. Each metric is calculated at each grid point, then horizontally-averaged and (for $dT/dt$ and $dq/dt$) vertically-averaged. The units of non-energy flux variables are converted to a common energy unit, W/m$^\text{2}$, following Section \ref{subsec:weighting}. Best model performance for each variable is highlighted in bold.}
\label{tab:summarystats}
\end{table}

% ===============================
\subsection{Results}
% ===============================

MAE and R$^\text{2}$ of the baseline models are presented in the main text (e.g., Table 1 and Figure 2 in the main text). Here, we show RMSE and CRPS in Table \ref{tab:summarystats} and Figures \ref{tab:summarystats}, \ref{fig:ttendmetrics}, and \ref{fig:qtendmetrics}.

\begin{figure}[!htbp]
    \centering
    \includegraphics[width=0.9\textwidth]{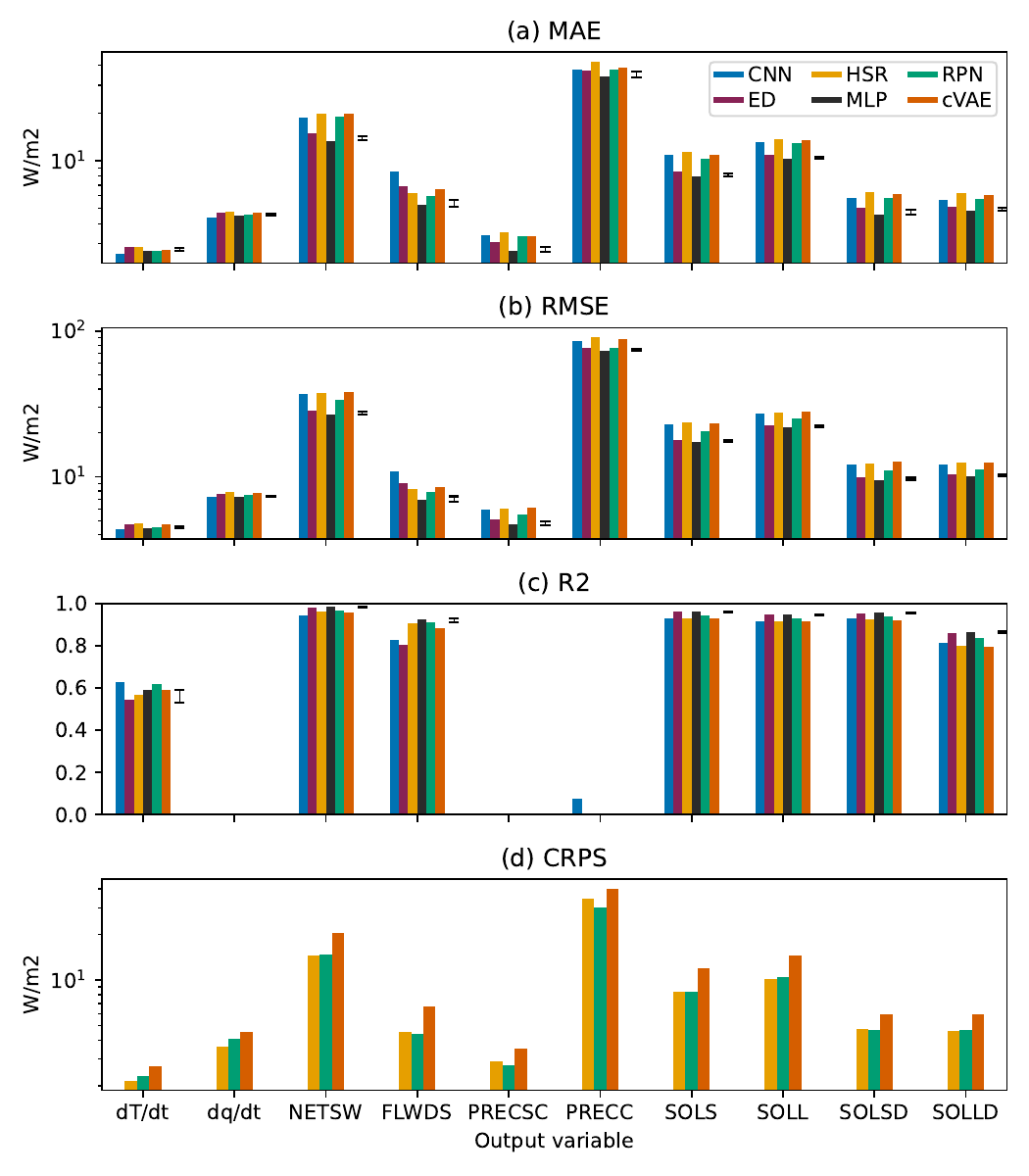}
    \setlength{\belowcaptionskip}{-10em}%
    \caption{Averaged (a) MAE, (b) RMSE, (c) R$^\text{2}$, and (d) CRPS. Each metric is calculated at each grid point, then horizontally-averaged and (for $dT/dt$ and $dq/dt$) vertically-averaged. For MAE, RMSE, and CRPS, the units of non-energy flux variables are converted to a common energy unit, W/m$^\text{2}$, following Section \ref{subsec:weighting}. Negative values are not shown for R$^\text{2}$. Error bars show the 5- to 95-percentile range of MLP.}
    \label{fig:metricsbarplot}
\end{figure}

\begin{figure}[!htbp]
    \centering
    \includegraphics[width=0.85\textwidth, trim={0 .1cm 0 .1cm},clip]{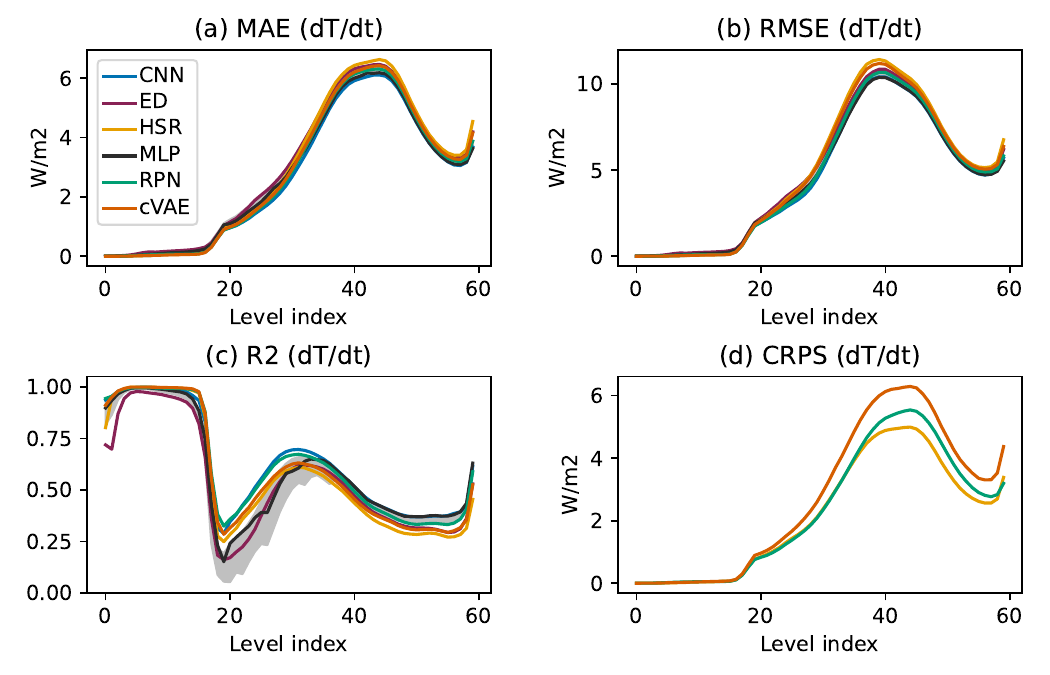}
    \setlength{\belowcaptionskip}{-10em}%
    \caption{Vertical structures of horizontally-averaged (a) MAE, (b) RMSE, (c) R$^\text{2}$, and (d) CRPS of $dT/dt$. For MAE, RMSE, and CRPS, the units of non-energy flux variables are converted to a common energy unit, W/m$^\text{2}$, following Section \ref{subsec:weighting}. Negative values are not shown for R$^\text{2}$. Grey shadings show the 5- to 95-percentile range of MLP.}
    \label{fig:ttendmetrics}
\end{figure}

\begin{figure}[!htbp]
    \centering
    \includegraphics[width=0.85\textwidth, trim={0 .1cm 0 .1cm},clip]{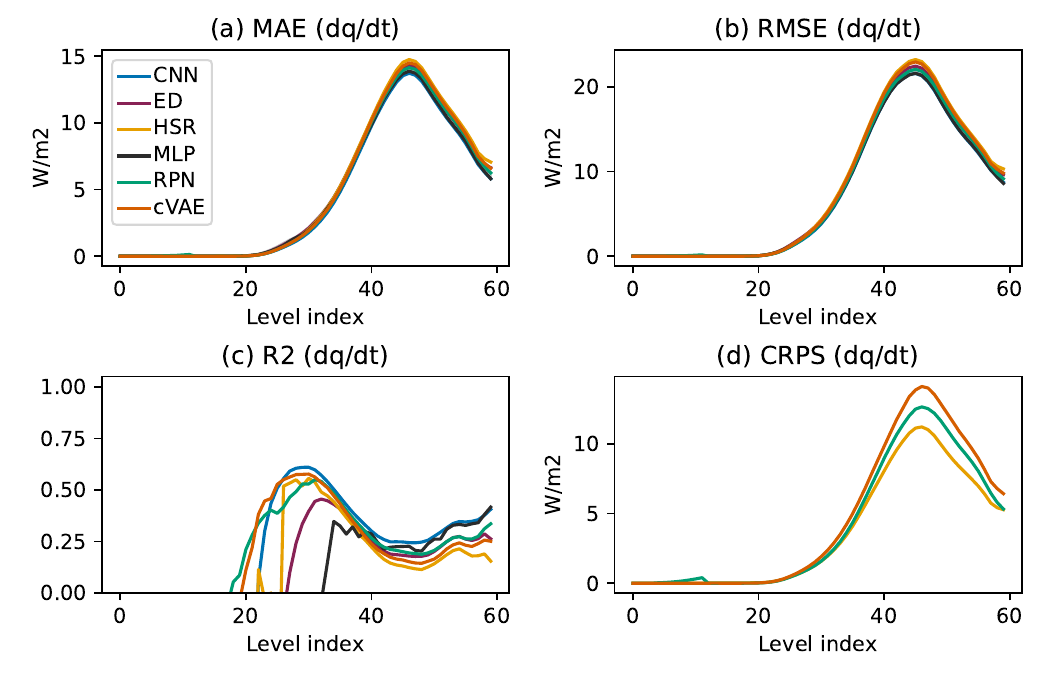}
    \setlength{\belowcaptionskip}{-1em}%
    \caption{Similar to Figure \ref{fig:ttendmetrics}, but for $dq/dt$.}
    \label{fig:qtendmetrics}
\end{figure}

\begin{figure}[!htbp]
    \centering
    \includegraphics[width=.9\textwidth, trim={0 0 0 0},clip]{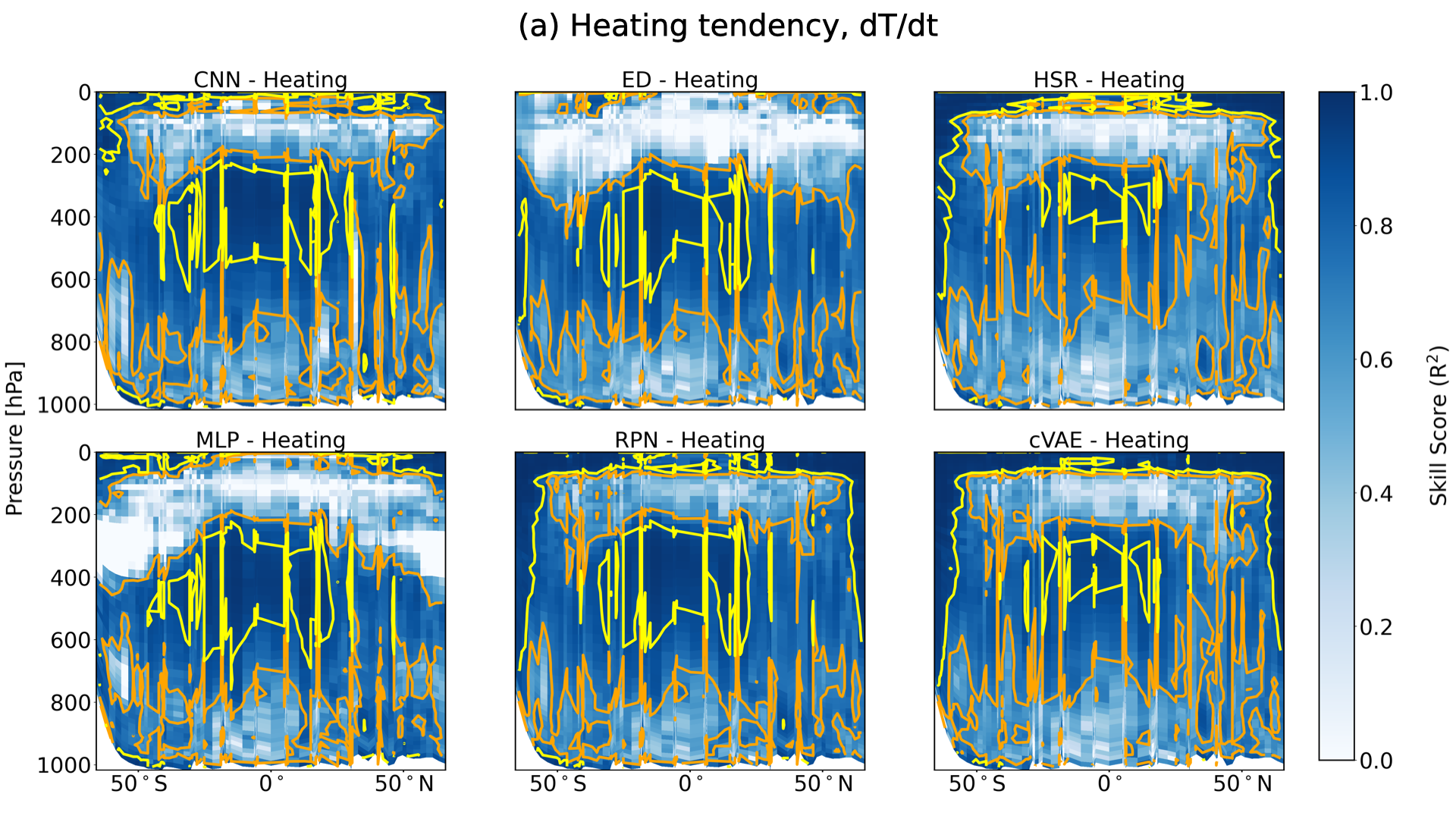}
    \includegraphics[width=.9\textwidth, trim={0 0 0 0},clip]{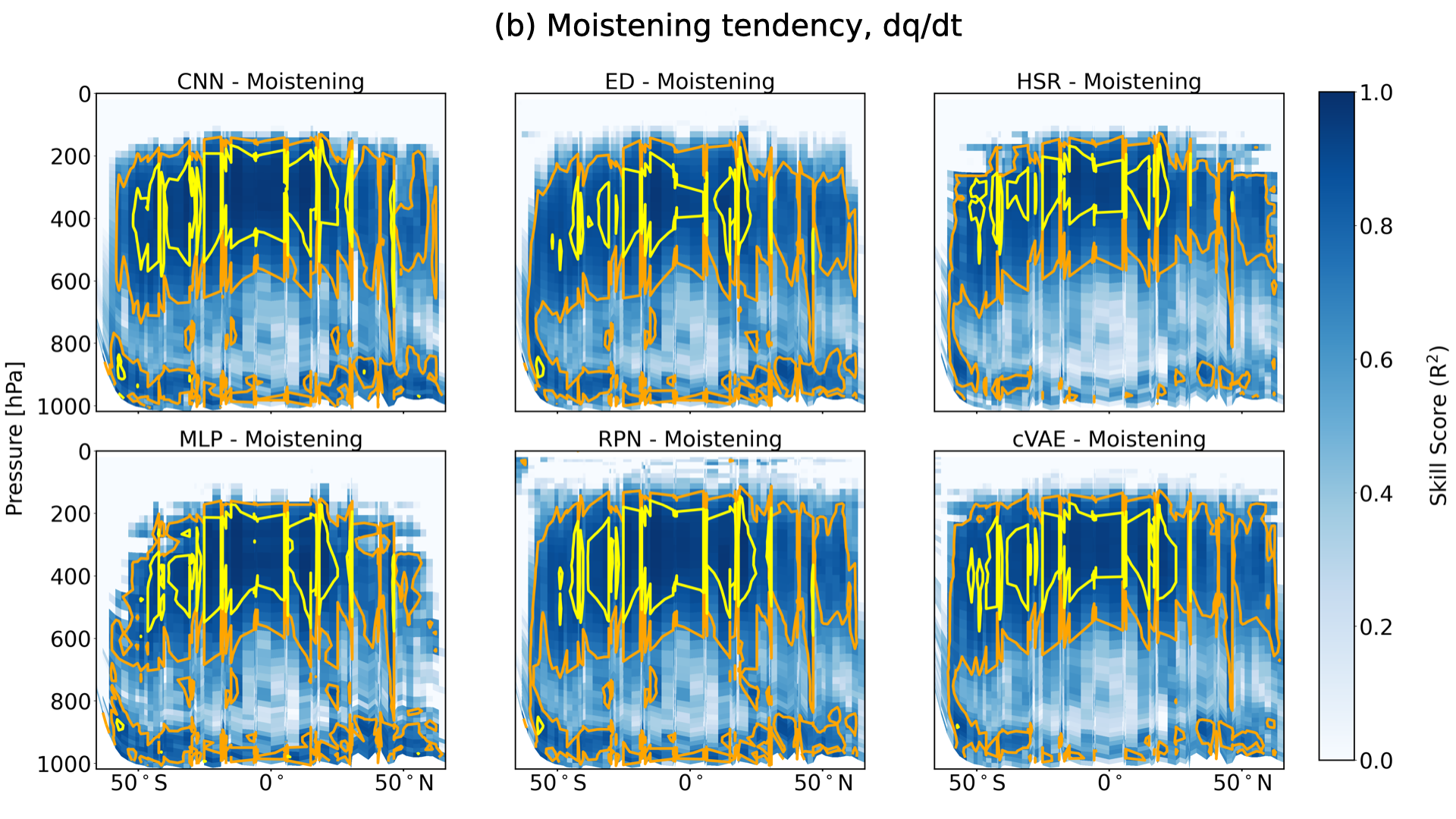}
    \setlength{\belowcaptionskip}{-0.75em}%
    \caption{R$^\text{2}$ of daily-mean, zonal-mean (a) heating tendency and (b) moistening tendency. Yellow contours surround regions of $> .9 R^2$ while orange contours surround regions of $> .7 R^2$. Negative values are not plotted (white). Sin(latitude) is used for x-axis to account for the curvature of Earth. The pressure levels on Y-axis are approximated values. }
    \label{fig:tendskillplot}
\end{figure}

We also present the spatial structure of the metrics. Figure \ref{fig:tendskillplot} shows the latitude-height structure of R$^\text{2}$.

% ===============================
\subsection{Fit Quality}
% ===============================

 Scatter plots of truth versus prediction are shown in this section (Figures SI\ref{fig:SI_Scatter_2D_part1} to SI\ref{fig:SI_Scatter_3D_part4} in SI Section \ref{sec:ExtraFigures}). While many variables exhibit consistent fit quality, some show notable variability between baselines, as seen with snow precipitation rate predictions. The performance of our optimized deterministic baseline (MLP) suggests these issues are avoidable. However, note that our prediction problem has a multi-variate and multi-dimensional nature.

% ======================================================
\section{Guidance}
% ======================================================
% ===============================
\subsection{Physical Constraints}
% ===============================

Mass and energy conservation are important criteria for Earth system modeling. If these terms are not conserved, errors in estimating sea level rise or temperature change over time may become as large as the signals we hope to measure. Enforcing conservation on emulated results helps constrain results to be physically plausible and reduce the potential for errors accumulating over long time scales.

In the atmospheric component of the E3SM climate model, mass is composed of ``dry air'' (i.e., well-mixed gases such as molecular nitrogen and oxygen) and water vapor. During the physics parameterizations we seek to emulate, there is no lateral exchange of mass across columns of the host model, and the model assumes that the total mass in each column and level remains unchanged. Thus, while surface pressure (\texttt{state\_ps}) is part of the state structure we seek to emulate, that surface pressure component must be held fixed. The water mass, however, is not held fixed, requiring fictitious sources and sinks of dry air, which are corrected later in the model---outside of the ``emulated'' part of the code---and is not addressed within the emulator.

Changes in column water mass should balance the sources and sinks of water into and out of the column through surface fluxes. The surface source of water is an input to the emulator via the \texttt{cam\_in} structure. The surface sink of water is generated by the model, and hence emulated in our case. The net surface water flux (source minus sink) should be equal to the tendency of water mass within the column (\ref{eq:waterconservation}). The mass of water is held in five separate terms within the \texttt{state} structure: water vapor ($q_v$), cloud liquid condensate ($q_l$), cloud ice ($q_i$), rain ($q_r$), and snow ($q_s$). These terms are held as ratios of their mass to the sum of dry air plus water vapor (referred to as specific humidity). The ``$\delta$'' refers to the difference (after minus before computation) in each quantity owing to the CRM physics. The layer mass (sum of dry air and water vapor) of level $k$ is equal to the pressure thickness of that layer $\Delta p_i$ (the difference between top and bottom interface pressure for level $i$) divided by the gravitational acceleration $g$ (assumed constant). The timestep length is $\delta t$. In addition to conserving water mass, we required each individual water constituent to remain greater than or equal to zero in every layer within the column. In Equation \ref{eq:waterconservation}, $E$ is the surface source of water (evapotranspiration) and $P$ is the surface sink of water (precipitation):
\begin{equation}
    \sum_i \left( \delta q_v + \delta q_l + \delta q_i + \delta q_r + \delta q_s \right) \frac{\Delta p_i}{g \delta t} = E - P
    \label{eq:waterconservation}
\end{equation}
For the portion of the code that we try to emulate, the water source $E$ is not applied such that the only surface flux to account for when constraining water conservation is the precipitation flux ($P$, \texttt{cam\_out\_PRECC}). Unfortunately, only the input and output state variables for water vapor (\texttt{state\_q0001}), cloud liquid (\texttt{state\_q0002}), and cloud ice (\texttt{state\_q0003}) are available. Additional storage terms related to precipitating water that have not exited the column over the course of a model timestep are unavailable in the current output. Therefore, we are unable to exactly enforce water conservation. Estimates show relative errors of a couple percent resulting from the lack of these precipitation mixing ratios. We can still require that the relative error be small. To accomplish this, we compared the ``expected'' total water, based on the combination of the input and surface fluxes, to the predicted total water. In the equations below, superscript $o$ denotes output and superscript $i$ denotes input:
\begin{align*}
    \mathrm{Total\;Water\;(Actual)} &= \sum_i \left( \delta q_v^o + \delta q_l^o + \delta q_i^o\right) \frac{\Delta p_i}{g} \\
    \mathrm{Total\;Water\;(Expected)} &= \sum_i \left( \delta q_v^i + \delta q_l^i + \delta q_i^i\right) \frac{\Delta p_i}{g} - P\delta t \\
    \mathrm{Relative\;Error} &= \frac{\mathrm{Total\;Water\;(Expected)} - \mathrm{Total\;Water\;(Actual)}}{\mathrm{Total\;Water\;(Actual)}} \\
\end{align*}
We required the model to keep the relative error small (e.g., below 5\%). Anything further is beyond the limit of the current data.

Like mass conservation, energy conservation can generally be enforced by requiring that the total change within the column is exactly balanced by the fluxes into and out of that column. Because the emulator does not predict upwelling radiative fluxes at the model top (a sink term for energy), we do not have the boundary conditions necessary to constrain column energy tendencies. However, we still required certain criteria be met for physical consistency. First, the downwelling surface shortwave radiative flux cannot exceed the downwelling shortwave flux at the model top (prescribed input \texttt{pbuf\_SOLIN}). Likewise, the net surface shortwave flux should also be bounded between zero (100\% reflection) and the surface downwelling shortwave flux (100\% absorption). Additionally, the downwelling longwave flux should not exceed the blackbody radiative flux from the warmest temperature in the column.

In addition to conservation laws, cloud physics are found to be useful to constrain model predictions in ways that significantly benefit online error and stability in hybrid ML-physic climate simulations. Cloud formation is partially governed by the thermal state of the atmosphere. See Section 6.3.4 for discussions and implementations of some cloud physics constraints and approximations that are beneficial.
% ===============================
\subsection{Unit Conversion and Weighting for Interpretable Evaluation}
\label{subsec:weighting}
% ===============================

To facilitate the objective evaluation of the model's prediction, we provided a weight tensor of shape $\left(d_{o},N_{\boldsymbol{x}}\right)$ to convert raw outputs to area-weighted outputs with consistent energy flux units [$\mathrm{W/m^{2}}$]. More details are given below. 

To ensure that our evaluation takes the Earth's spherical geometry into account, we designed an area weighting factor $a$ that depends on the horizontal position $\boldsymbol{x}$:
\begin{align*}
    a\left(\boldsymbol{x}\right)={\cal A}_{\mathrm{col}}\left(\boldsymbol{x}\right)/\left\langle {\cal A}_{\mathrm{col}}\right\rangle _{\boldsymbol{x}}
\end{align*}
where ${\cal A}_{\mathrm{col}}$ is the area of an atmospheric column and $\left\langle {\cal A}_{\mathrm{col}}\right\rangle _{\boldsymbol{x}}$ the horizontal average of all atmospheric columns' areas. This formula gives more weight to outputs if their grid cell has a larger horizontal area. To ensure that our evaluation is physically-consistent, we convert all predicted variables to energy flux units $\left[\mathrm{W/m^{2}}\right]$ (power per unit area). This has to be done for each variable separately. 
\begin{itemize}
    \item For the heating tendency $\dot{T}\left[\mathrm{K/s}\right]$, which depends on the horizontal position $\boldsymbol{x}$ and vertical level $\mathrm{lev}$, this was done using the specific heat capacity at constant pressure $c_{p}\left[1004.64\,\mathrm{J/\left(K\times kg\right)}\right]$, where $\Delta p_{i}\left[\mathrm{Pa}\right]$ is the layer's pressure thickness, calculated as the difference between the pressure at the layer's top and bottom interfaces:
    \begin{align*}
        \dot{T}\left[\mathrm{W/m^{2}}\right]=\frac{c_{p}}{g}\times a\left(\boldsymbol{x}\right)\times\Delta p_{i}\left(\mathrm{lev}\right)\times\dot{T}\left[\mathrm{K/s}\right]
    \end{align*}
    \item For the water concentration tendency $\dot{q}\left[\mathrm{s^{-1}}\right]$, which also depends on $\boldsymbol{x}$ and $\mathrm{lev}$, this was done using the latent heat of vaporization of water vapor at constant pressure $L_{v}\left[2.50\times10^6\,\mathrm{J/kg}\right]$:
    \begin{align*}
        \dot{q}\left[\mathrm{W/m^{2}}\right]=\frac{L_{v}}{g}\times a\left(\boldsymbol{x}\right)\times\Delta p_{i}\left(\mathrm{lev}\right)\times\dot{q}\left[\mathrm{s^{-1}}\right]
    \end{align*}
    Note that there is some level of arbitrariness, as the exact latent heat depends on which water phase is assumed to calculate the energy transfer. Here, we chose to weigh all phases using $L_{v}$ to give them comparable weights in the evaluation metrics.
    \item For momentum tendencies $\dot{u}\left[\mathrm{m/s^{2}}\right]$, which also depend on $\boldsymbol{x}$ and $\mathrm{lev}$, we used a characteristic wind magnitude $\left|\boldsymbol{U}\right|\left[\mathrm{m/s}\right]$ to convert these tendencies into turbulent kinetic energy fluxes, in units $\mathrm{W/m^{2}}$, making them comparable to $\dot{T}\left[\mathrm{W/m^{2}}\right]$ and $\dot{q}\left[\mathrm{W/m^{2}}\right]$:
    \begin{align*}
        \dot{u}\left[\mathrm{W/m^{2}}\right]=\frac{\left|\boldsymbol{U}\right|}{g}\times a\left(\boldsymbol{x}\right)\times\Delta p_{i}\left(\mathrm{lev}\right)\times\dot{u}\left[\mathrm{m/s}^{2}\right]
    \end{align*}
    Note that there is some level of arbitrariness in the choice of $\left|\boldsymbol{U}\right|\left[\mathrm{m/s}\right]$, which could e.g., be chosen so that the variances of $\dot{u}\left[\mathrm{W/m^{2}}\right]$ and $\dot{T}\left[\mathrm{W/m^{2}}\right]$ are comparable.
    \item Precipitation rate variables $P\left[\mathrm{m}/\mathrm{s}\right]$ were also be converted to energy fluxes using $L_{v}$ and the density of liquid water $\rho_{w}$ $\left[\mathrm{kg/m^{3}}\right]$ (or the density of snow/ice for solid precipitation), though they do not require vertical integration:
    \begin{align*}
        P\left[\mathrm{W/m^{2}}\right]=L_{v}\times \rho_{w} \times a\left(\boldsymbol{x}\right)\times P\left[\mathrm{m}/\mathrm{s}\right]
    \end{align*}
    \item Finally, surface energy fluxes ${\cal F}\left[\mathrm{W/m^{2}}\right]$ were simply multiplied by $a\left(\boldsymbol{x}\right)$ to account for area-weighting. 
\end{itemize}
Note that while these choices ensured unit consistency, facilitating the physical interpretation of our evaluation metrics, we recommend tailoring the exact choice of physical constants to the application of interest. 

% ===============================
\subsection{Additional Guidance}
% ===============================

\textbf{Stochasticity and Memory:} The results of the embedded convection calculations regulating $d_o$ come from a chaotic dynamical system and thus could be worthy of architectures and metrics beyond the deterministic baselines in this paper. These solutions are likewise sensitive to sub-grid initial state variables from an interior nested spatial dimension that have not been included in our data. 

\textbf{Temporal Locality:} Incorporating the previous timesteps' target or feature in the input vector inflation could be beneficial as it captures some information about this convection memory and utilizes temporal autocorrelations present in atmospheric data. This approach has been explored in previous studies \citep{Han2020,Wang2022a,han2023ensemble,lin2023systematic,behrens2024improving} and has been integrated into our model for online testing (see Section 5 in the main text and Section 6.3.3 in SI).

\textbf{Causal Pruning}: A systematic and quantitative pruning of the input vector based on objectively assessed causal relationships to subsets of the target vector has been proposed as an attractive preprocessing strategy, as it helps remove spurious correlations due to confounding variables and optimize the machine learning (ML) algorithm \citep{Iglesias2023}.

\textbf{Normalization:} Normalization that goes beyond removing vertical structure could be strategic, such as removing the geographical mean (e.g., latitudinal, land/sea structure) or composite seasonal variances (e.g., local smoothed annual cycle) present in the data. For variables exhibiting exponential variation and approaching zero at the highest level (e.g., metrics of moisture), log-normalization might be beneficial. 

% ======================================================
\section{Online Evaluation Pipeline and Experiment Details}
% ======================================================

This section provides a detailed example of the online evaluation pipeline for hybrid simulations. First, we describe the integration of the Python-based machine learning (ML) model into the Fortran-based climate simulator using the Pytorch-Fortran bindings library \citep{dmitry_alexeev_2023_7851167}. This integration requires the ML model to be in the TorchScript format, which we will explain in the following subsection, including examples and notes on converting a PyTorch model into TorchScript. Next, we present online results, citing our work in \cite{Hu2024} to illustrate methods for evaluating online errors, share insights for optimizing stability and performance, and showcase the best online errors we have achieved so far along with potential pathways for further improvements.

% ======================================================
\subsection{Converting PyTorch Models to TorchScript}
% ======================================================
Converting a PyTorch model to TorchScript is straightforward. However, it is essential to ensure that the model does not use operations unsupported by TorchScript. Most standard PyTorch operations are supported, but some advanced or less common operations may not be. Documentation for supported operations can be found at \url{https://pytorch.org/docs/stable/jit_builtin_functions.html#builtin-functions}, and unsupported constructs are listed at \url{https://pytorch.org/docs/stable/jit_unsupported.html#jit-unsupported}.

TorchScript requires more static typing compared to regular Python. Ensure your functions have clear type annotations and avoid using Python-only constructs unsupported by TorchScript. For detailed instructions, refer to the official TorchScript documentation: \url{https://pytorch.org/docs/stable/jit.html}.

Below is an example of how to convert a PyTorch MLP model to TorchScript:

\begin{python}[language=Python, caption=Converting a PyTorch MLP model to TorchScript]
import torch
import torch.nn as nn
import torch.jit

# Define the MLP model
class MLP(nn.Module):
    def __init__(self, in_dims: int = 512, out_dims: int = 368):
        super(MLP, self).__init__()
        self.linear1 = nn.Sequential(
            nn.Linear(in_dims, 512),
            nn.ReLU()
        )
        self.final_linear = nn.Linear(512, out_dims)

    def forward(self, x: torch.Tensor) -> torch.Tensor:
        x = self.linear1(x)
        x = self.final_linear(x)
        return x

# Create an instance of the model
model = MLP()

# Convert the model to TorchScript using scripting
scripted_model = torch.jit.script(model)

# Set the model to evaluation mode
scripted_model = scripted_model.eval()

# Save the TorchScript model to a file
scripted_model.save('model.pt')
\end{python}

\subsection{Metrics for Evaluating Hybrid Climate Simulators}

\textbf{Root Mean Square Error:} Our online evaluation metrics are computed separately for each variable in the hybrid simulations. The goal is to measure the error in simulated climate by analyzing state variables that are sufficiently averaged in space and time. For a given month, RMSE for each variable is calculated as follows:

$$\text{RMSE} = \sqrt{\sum\limits_{i=1}^{S_m} w_i (\hat{y}_m - y_m)^2}$$

where:
\begin{itemize}
  \item $S_m$ is the number of samples (each grid cell is one sample) across the entire globe, both horizontally and vertically.
  \item $\hat{y}_m$ represents the values from the hybrid simulation averaged over the entire month.
  \item $y_m$ represents the values from the reference simulation averaged over the entire month.
  \item $w_1, w_2, \ldots, w_{S_m}$ are mass-weights that sum to 1, proportional to the air mass in each grid cell.
\end{itemize}

\textbf{Zonal Mean Bias:} Additionally, we evaluate the long-term zonal mean bias, which measures the average difference between the hybrid simulation and the reference simulation across various atmospheric variables, such as temperature, moisture, wind, cloud water, and cloud ice. The zonal mean bias is derived by comparing variables averaged over time and longitudes. The E3SM-MMF climate simulator outputs variables on unstructured grids instead of regular latitude-longitude grids. To average a variable over all longitudes, we first define latitude bins. For the low-resolution version of the \dataset{} dataset, we choose 10-degree intervals from 90°S to 90°N. Within each bin, we count all the grid columns that fall into the bin and calculate the horizontal average of those columns weighted by their area.

% ======================================================
\subsection{Experiment Setup for Online Testing}

\subsubsection{Multilayer Perceptron (MLP)}

We used the architecture parameters recommended from the hyperparameter search in Section \ref{MLP}, with \textit{N}$_\text{layers} =$ 3, \textit{N}$_\text{nodes} =$ [384, 1024, 640], and ReLU activation. We trained these MLP models with a batch size of 1024 using the Adam optimizer on 4 GPUs. The training utilized the full low-resolution training data without subsampling.

\subsubsection{U-Net}

\begin{figure}[!htbp]
    \centering
    \includegraphics[width=1.0\textwidth]{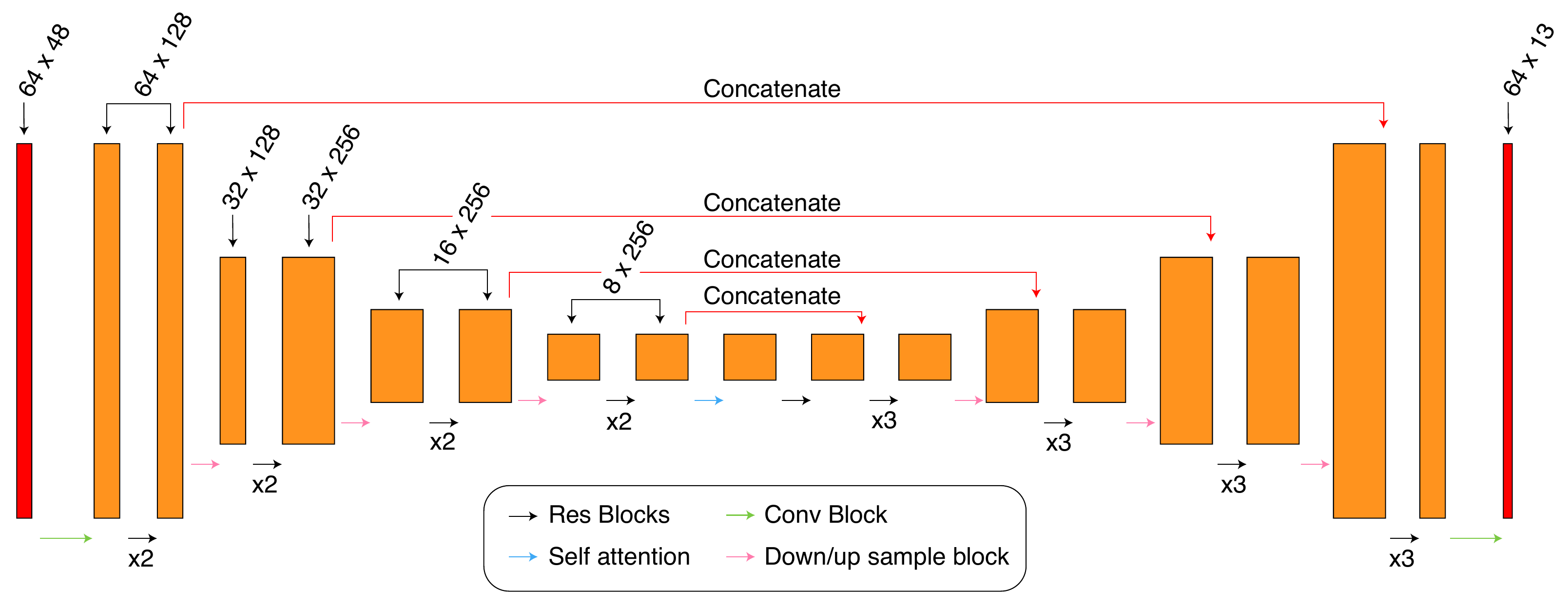}
    \setlength{\belowcaptionskip}{-1em}%
    \caption{The schematic of the U-Net architecture. The U-Net architecture is comprised of multiple ResBlocks, with 4 layers in depth and latent dimension sizes of [128, 256, 256, 256]. The model requires preparing the input and output as sequence data, with different features as the channels of the sequence and the sequence as the vertical dimension. See \cite{Hu2024} for more details.}
    \label{fig:unet}
\end{figure}

Since the MLP model turned out not to succeed on the downtream task, we now describe a considerably more sophisticated approach that proved more successful. \cite{Hu2024} adapted the 2D U-Net model from \cite{song2019generative} into a 1D version for column-to-column prediction. A U-Net model can efficiently learn and use the vertical structures in the atmosphere. The architecture schematic is shown in Figure \ref{fig:unet}. Each ResBlock consists of the following operations:
\begin{align*}
y & = \mathrm{Conv1D}(\mathrm{GM}(\mathrm{Conv1D}(\mathrm{silu}(\mathrm{GM}(x))))) + x
\end{align*}

This series of operations includes group normalization (GM), sigmoid linear unit (silu) activation function, 1D convolution (Conv1D) with a kernel size of 3, and a residual connection. The U-Net has 4 layers in depth with latent feature dimensions in each layer \textit{N}$_\text{latent} =$ [128, 256, 256, 256]. The model comprises 13 million parameters. Like the MLP models, we trained these U-Net models with a batch size of 1024 using the Adam optimizer on 4 GPUs, utilizing the full low-resolution training data without subsampling. More training details, e.g. input and output normalization, can be found in \cite{Hu2024}.

\subsubsection{Additional Inputs for the U-Net Model}

The U-Net model utilized additional input variables listed in Table \ref{tab:additionalvars}. These inputs include large-scale forcing at the current and previous time steps, and the convection memory (i.e., the target) at two previous time steps. Large-scale forcing and convection memory are also used in \cite{Han2020, Wang2022a, han2023ensemble}. Cosine and sine of latitudes are included as well. When cloud physics constraints are enabled (see Section \ref{cloud-constraints}), liquid cloud and ice cloud inputs are replaced with the corresponding total cloud condensate inputs (liquid plus ice), along with a diagnosed fraction of liquid cloud based on temperature. All these inputs can be retrieved from the existing \dataset{} dataset. Retrieving large-scale forcing and convection memory requires using continuous time series data without gaps.

\textbf{Dynamical Forcing:} Dynamical forcing can be calculated as follows:

\lstset{basicstyle=\ttfamily, columns=fullflexible, mathescape=true}
\begin{lstlisting}
state_t_dyn(t=0)       = (state_t_in(t=0)       - state_t_out(t=-1)) / 1200
state_q0001_dyn(t=0)   = (state_q0001_in(t=0)   - state_q0001_out(t=-1)) / 1200
state_q0002_dyn(t=0)   = (state_q0002_in(t=0)   - state_q0002_out(t=-1)) / 1200
state_q0003_dyn(t=0)   = (state_q0003_in(t=0)   - state_q0003_out(t=-1)) / 1200
state_q0_dyn           =  state_q0001_dyn + state_q0002_dyn + state_q0003_dyn
state_u_dyn(t=0)       = (state_u_in(t=0)       - state_u_out(t=-1)) / 1200
\end{lstlisting}

Here, the subscripts '\_in' and '\_out' refer to the variables in the input and output files, respectively. The current model time step is denoted as $t=0$, and $t=-1$ represents the previous time step. Dynamical forcing represents the rate of change of a variable between two time checkpoints over one time step (1200 seconds). The first checkpoint is before the call to the embedded cloud resolving model (CRM) in the multi-scale climate simulator (see Section \ref{model_description}) at the current time step, and the second is immediately after the call to the CRM at the previous time step. The changes in temperature, water, and wind fields during this period are due to model physics outside the CRM, primarily advection from the host dynamical model's planetary scale fluid solver. These dynamical tendencies of temperature, total water (vapor plus cloud), and zonal winds are applied to the CRM grids during the CRM calculation, making them useful input features for training.

\textbf{Convection Memory:} Convection memory refers to the residual effects of convective processes in the CRM model that might not be fully captured by the values in the coarser grids of the host climate simulator. We utilize the CRM tendencies from two previous time steps to represent this convection memory, as suggested by \cite{Han2020, Wang2022a, han2023ensemble}. At each time step, the CRM tendencies can be retrieved as follows:

\lstset{basicstyle=\ttfamily, columns=fullflexible, mathescape=true}
\begin{lstlisting}
state_t_prvphy     = (state_t_out     - state_t_in)     / 1200
state_q0001_prvphy = (state_q0001_out - state_q0001_in) / 1200
state_q0002_prvphy = (state_q0002_out - state_q0002_in) / 1200
state_q0003_prvphy = (state_q0003_out - state_q0003_in) / 1200
state_u_prvphy     = (state_u_out     - state_u_in)     / 1200
\end{lstlisting}

During the hybrid simulation where an ML emulator replaces the CRM, the previous steps' ML predictions are used as the convection memory. Because this simulation requires information from previous steps as input features, the hybrid simulation must start by calling the CRM for during the initial a few host simulator's time steps before switching to using only the ML emulator. When launching a brand new hybrid simulation, the atmospheric fields in the CRM typically start from an unrealistic initial condition and require some host-model time steps to form realistic clouds and storms that are consistent with the training data. It is important to fully spin up the CRMs before switching to using only the ML emulator. For a brand new simulation, we recommend a spin-up period of at least one simulation day. In our container workflow, we can launch the hybrid simulation by restarting from the previous simulation where the atmosphere is already spun up. When this restarting configuration is used, we only need to run the CRM for a few steps, depending on how many previous step input features are required.

\begin{table}[!htbp]
\centering
\small
\resizebox{\textwidth}{!}{\begin{tabular}{cclccl}
\toprule
\textbf{In} & \textbf{Out} & \textbf{Variable} & \textbf{Dimensions} & \textbf{Units} & \textbf{Description} \\
\midrule
\midrule
$\times$ &  & state\_t\_dyn & lev, ncol, t=0,-1 & K/s & Large-scale forcing of temperature \\ \midrule
$\times$ &  & state\_q0\_dyn & lev, ncol, t=0,-1 & kg/kg/s & Large-scale forcing of total water \\ \midrule
$\times$ &  & state\_u\_dyn & lev, ncol, t=0,-1 & m/s\textsuperscript{2} & Large-scale forcing of zonal wind \\ \midrule
$\times$ &  & state\_t\_prvphy & lev, ncol, t=-1,-2 & K/s & Temperature tendency at previous steps \\ \midrule
$\times$ &  & state\_q0001\_prvphy & lev, ncol, t=-1,-2 & kg/kg/s & Water vapor tendency at previous steps \\ \midrule
$\times$ &  & state\_q0002\_prvphy & lev, ncol, t=-1,-2 & kg/kg/s & Liquid cloud tendency at previous steps \\ \midrule
$\times$ &  & state\_q0003\_prvphy & lev, ncol, t=-1,-2 & kg/kg/s & Ice cloud tendency at previous steps \\ \midrule
$\times$ &  & state\_u\_prvphy & lev, ncol, t=-1,-2 & m/s\textsuperscript{2} & Zonal wind tendency at previous steps \\ \midrule
$\times$ &  & clat & ncol &  & Cosine of latitude \\ \midrule
$\times$ &  & slat & ncol &  & Sine of latitude \\ \midrule
$\times$ &  & liq\_partition & lev, ncol &  & Fraction of liquid cloud \\ \midrule
\bottomrule
\end{tabular}}
\vspace{1mm}
\caption{Additional input variables used in the U-Net model}
\label{tab:additionalvars}
\end{table}

\subsubsection{Implementing Cloud Physics Constraints}
\label{cloud-constraints}
In this and the following section we describe two physical constraints helpful to achieving reasonable performance on the online task related to cloud condensate. 

\textbf{Liquid-Ice Cloud Partition:} The formulation of cloud microphysics in the CRM uses temperature to determine the mass of liquid and ice clouds on each grid, as a function of a more fundamental total non-precipitating water prognostic variable. Liquid clouds are only allowed to exist when the temperature is above 273.16K, and ice clouds are only allowed below 253.16K. In between these temperatures, the fraction of liquid cloud over the total cloud mixing ratio follows a linear function of temperature. \cite{Hu2024} demonstrates that this temperature-based partition relationship between liquid and ice clouds holds well on the grid of the host E3SM model. To incorporate this constraint, the U-Net model predicts only the total cloud (liquid plus ice) change and then diagnoses the liquid and ice clouds based on the updated temperature.

\textbf{Cloud Top Capped by the Tropopause Layer:} The tropopause is the boundary between the lower atmosphere (troposphere) and the upper atmosphere (stratosphere). It acts like a ceiling for weather systems. When a storm's rising air reaches this boundary, it encounters a very stable layer that prevents further upward movement, effectively capping cloud formation below a dynamical barrier. To identify the tropopause, we look for the lowest altitude where the pressure is below 400 hPa and the potential temperature vertical gradient is steep (greater than 10 K/km). In our simulations, any clouds that form above the tropopause (either from strong storms or being moved up by other atmospheric processes) are quickly removed by falling back down or evaporating. \cite{Hu2024} showed that residual clouds above the tropopause after CRM are rare (but not zero) and have low water content (less than $10^{-3}$ g/kg). The empirical conditions and thresholds used to detect the tropopause could be refined further. To improve the stability of the hybrid simulator, we include an approximate constraint that removes all clouds above the tropopause. This helps prevent unrealistic cloud accumulation in the upper atmosphere and reduces error growth in hybrid simulations. This constraint is implemented directly in the hybrid simulator and does not alter the model architecture and training.

\subsubsection{Sources of Online Uncertainty in Hybrid Simulations}

\textbf{Uncertainty from Sampling Model Checkpoints:} Previous studies, such as \cite{ott2020,Wang2022a}, have shown that improved offline skill can benefit online error, but the online error is not fully constrained by offline skill. This means different checkpoints with very similar offline skill can exhibit varying online stability and error. A trial-and-error approach is often used to identify the checkpoint with the best online performance. For subtle sensitivities, hundreds of trials can become important to detect downtream signals of hybrid climate error from upstream ML noise \citep{lin2023systematic}. 

To account for some of this online uncertainty, we trained three MLP models with different loss functions and learning rate schedules:
\begin{itemize}
  \item The first model used Mean Absolute Error (MAE) loss, with an initial learning rate of \(1 \times 10^{-3}\), reduced by a factor of 0.3162 every 7 epochs for a total of 28 epochs.
  \item The second model used the same learning rate schedule but a standard Huber loss with \(\delta=1\).
  \item The third model also used Huber loss but with a different learning rate schedule: starting at \(1 \times 10^{-3}\), using a ReduceOnPlateau scheduler with a patience of 3 epochs and a reduction factor of 0.3162 for a total of 20 epochs. This was followed by manually reducing the learning rate to \(1 \times 10^{-4}\) for fine-tuning over 12 more epochs with a patience of 0 epochs and a reduction factor of 0.5.
\end{itemize}

Similarly, we trained three versions of the U-Net model with cloud physics constraints and three versions of the U-Net models without the constraints:
\begin{itemize}
  \item The first used MAE loss, with an initial learning rate of \(1 \times 10^{-4}\), reduced by a factor of 0.5 every 3 epochs for a total of 16 epochs.
  \item The second model used the same learning rate schedule but a standard Huber loss with \(\delta=1\).
  \item The third used Huber loss with a different learning rate schedule: starting at \(1 \times 10^{-4}\) with a ReduceOnPlateau scheduler with a patience of 3 epochs and a reduction factor of 0.3162 for a total of 20 epochs, followed by manually reducing the learning rate to \(5 \times 10^{-5}\) for fine-tuning over 8 more epochs with a patience of 0 epochs and a reduction factor of 0.5.
\end{itemize}

\textbf{Uncertainty Due to Atmospheric Stochasticity:} The atmosphere is a chaotic system where small random perturbations can grow over time and lead to different weather patterns at weekly timescale. For online evaluation, we do not expect our ML emulator hybrid simulations to reproduce daily patterns but rather to match monthly or yearly mean patterns -- i.e. climate, not weather. However, the reference E3SM-MMF simulation's climate also contains some inherent uncertainty due to the way it is implemented and its stochastic nature. The E3SM-MMF simulator is not bit-for-bit reproducible due to the use of atomic operations, particularly during dimension reductions such as horizontal summation. These operations add numbers in a different random order each time, resulting in round-off errors that lead to non-reproducible outcomes. Tiny numerical differences in the calculations can grow, contributing to small but non-zero uncertainty in monthly atmospheric states. To estimate this atmospheric unpredictability, we ran the climate model three additional times with the same initial conditions and compared their differences to the reference simulation as a baseline uncertainty, which can be viewed as an error floor that the ML emulator simulations should not be expected to ever overcome.

\subsection{Diagnosis of Online Performance}

In Section 5.4 of the main text, we presented the global RMSE evolution for 1-year hybrid simulations using the MLP and U-Net models. The hybrid runs using baseline MLP models crashed within 2 months. In contrast, the more advanced U-Net architecture, with its expanded input features and microphysics constraints, significantly reduced the online error and achieved hybrid simulations that operated stably for years. 

The purpose is to clearly demonstrate the connection between the offline task around which \dataset{} is defined and its ultimate online impact -- achieving skillful hybrid climate runs, by showing two approaches that have considerably different online skill, and to illustrate how to appropriately measure hybrid climate errors.

\begin{figure}[!htbp]
    \centering
    \includegraphics[width=1.0\textwidth]{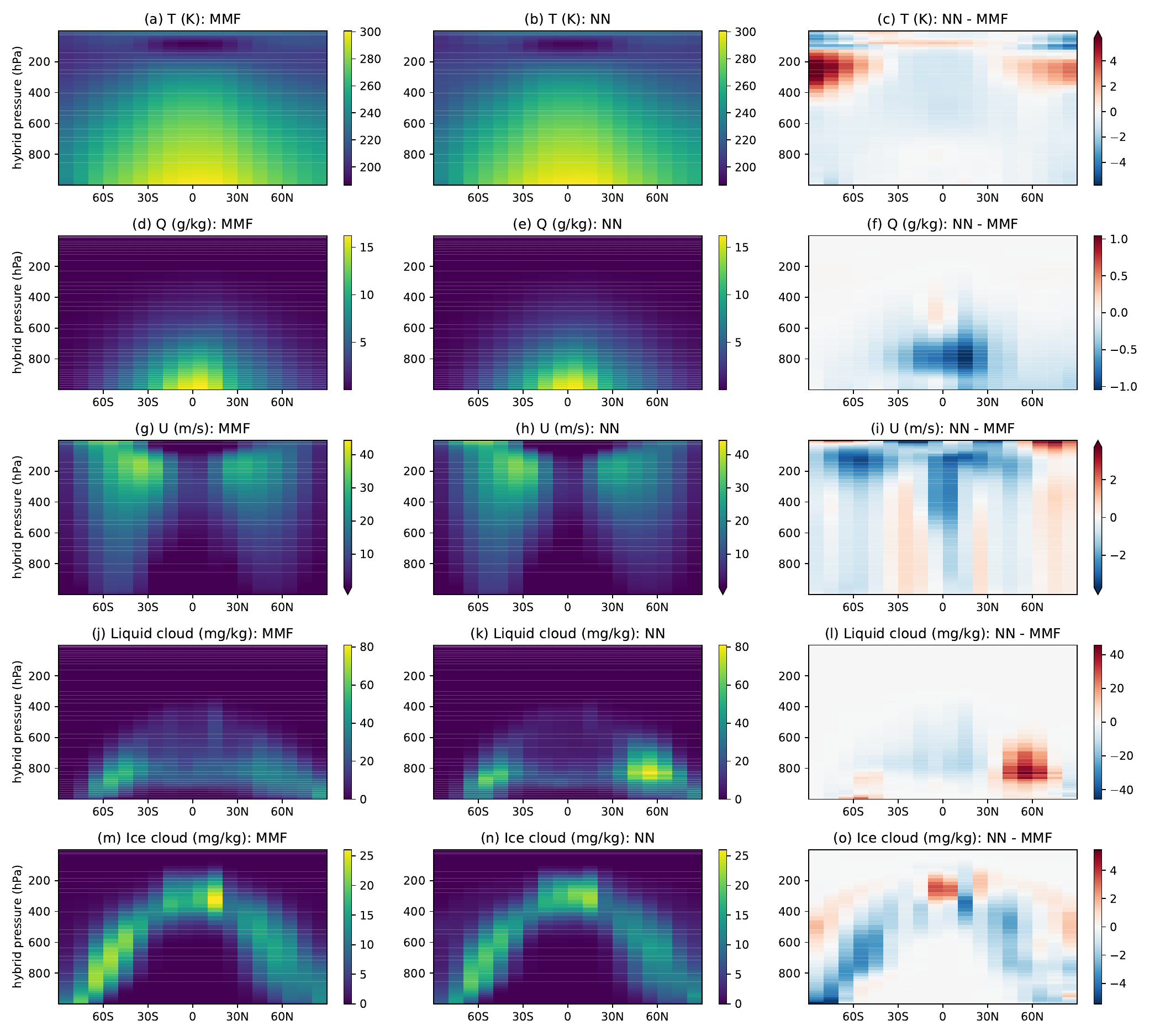}
    \setlength{\belowcaptionskip}{-1em}%
    \caption{Five-year zonal mean atmospheric state in the reference E3SM-MMF simulation (left) and in the hybrid simulation (middle). The right column shows the zonal mean bias as the mean state from the hybrid simulation minus that of the reference simulation. The five rows show temperature, water vapor, zonal wind, liquid cloud, and ice cloud.}
    \label{fig:zonal-mean}
\end{figure}

As a final illustration, we extended the simulation of one of the constrained U-Net models to 5 years. At this level of sampling geographic structures of the simulated climate are statistically stable and worth revealing. pressure-vs-latitude structure of the zonal mean bias of the 5-year mean state in this hybrid simulation. For comparison, we also ran a reference 5-year E3SM-MMF simulation with the same initial conditions. In the lower atmosphere, the zonal mean temperature bias was approximately within 2K, and the moisture bias was around 1 g/kg (Figure \ref{fig:zonal-mean}c and f). The zonal mean structures of wind and cloud distribution also compared favorably with the reference simulation. 

To our knowledge, this represents state-of-the-art performance in online bias for hybrid climate simulation tests with real-geography and full subgrid physics emulation within the context of multi-scale modeling framework. Beyond the context of multi-scale modeling framework, other hybrid models have achieved better online bias \citep{kochkov2023neural,sanford2023improving}. See Section 5.3 in the main text for some reference numbers. \cite{kochkov2023neural} used an end-to-end training framework, developing a fully differentiable hybrid model and directly optimizing multi-step rollout loss. Nevertheless, making existing Fortran-based climate models differentiable is challenging.

\subsection{Challenges and Future Directions in Online Error Optimization}

\textbf{Checkpoint Sensitivity and Persistent Error Patterns:} Optimizing the online error, as shown in Figure \ref{fig:zonal-mean} and in the Section 5.4 of the main text, presents significant challenges. Unlike offline errors, online errors are non-differentiable and cannot be optimized directly using gradient descent. While we have identified some methods for improving offline error, such as enhancing model architecture and adding physics constraints, can help reduce online errors, these methods are not sufficient on their own. Once the architecture is fixed, we currently rely on checkpoint searches to identify the checkpoint that yields the best online performance. Some error patterns, such as zonal-mean cloud bias at latitudes greater than 30N or 30S, are sensitive to the choice of checkpoint. For example, we observed checkpoints that do not exhibit the significant positive liquid cloud bias near 60N seen in Figure \ref{fig:zonal-mean}l. However, other error patterns, such as the stratospheric dipole temperature bias in high latitudes (Figure \ref{fig:zonal-mean}c) and drying vapor and liquid cloud biases in the tropics (Figure \ref{fig:zonal-mean}f and l), are persistent and seem so far immune to checkpoint search.

\textbf{Exploring Gradient-Free and Differentiable Methods:}  Fully optimizing these online errors remains an open question and important challenge for the community. Promising approaches might include gradient-free methods, such as imitation learning and online learning, and it would be interested to explore whether these could further optimize these bias patterns \citep{ross2011reduction, Rasp2020b, kelp2022online, lopez2022training, pahlavan2024explainable}. Another approach involves making the online error differentiable by either training a differentiable emulator for all the physics outside the CRM or rewriting the entire numerical climate simulator in a differentiable manner, as demonstrated by \cite{kochkov2023neural}. With a model that can be integrated in time in a differentiable way, it would be possible to optimize multi-step losses using gradient descent. However this is a nontrivial effort for a fully-featured climate simulator.

% ======================================================

% ======================================================
\section{Other Related Work}
% ======================================================

Several benchmark datasets have been developed to facilitate AI tasks in weather and climate. ClimateNet \citep{Prabhat2021} and Extremeweather \citep{Racah2017} were both designed for AI-based feature detection of extreme weather events in forecasts of Earth's future climate made using conventional climate models. WeatherBench 2 \citep{rasp2024weatherbench} provides data specifically designed for data-driven weather forecasting, focusing on periods ranging from 3 to 5 days into the future. PDEBench \citep{Takamoto2023} provides data from numerical simulations of several partial differential equations (PDEs) for benchmarking AI PDE emulators. ClimateBench \citep{Watson-Parris2022} was designed for emulators that produce annual mean global predictions of temperature and precipitation given greenhouse gas concentrations and emissions. The ClimateBench provides data from onyl one climate model, while ClimateSet \citep{kaltenborn2023climateset} expands on the ML tasks in ClimateBench by providing a large-scale dataset with inputs and outputs from 36 climate models. ClimART \citep{Cachay2021} was designed for the development of radiative energy transfer parameterization emulators for use in weather and climate modeling. These benchmark datasets play a vital role in advancing AI and ML research within the weather and climate domains. 

\dataset{}, a dataset for parameterization emulators trained on high-resolution data from small-scale embedded models, is unique compared to other benchmark datasets designed for emulators in climate simulation (ClimateBench, ClilmateSet, ClimART, and PDEBench). While PDEBench provides data for developing AI emulators of the same PDEs commonly used in climate simulation, \dataset{} is uniquely tailored to address the challenging task of replacing a sophisticated parameterization for the combined effects of clouds, rain, radiation, and storms. Specifically, models trained using \dataset{} will learn to emulate the nonlinear effect of clouds, rain, and storms resolved on the 1 km (20 s) space (time) scale, which is a collection of hundreds of equations rather than one, to represent their upscale impacts on the 100 km (30 min) scale. Hybrid simulation is also the goal of ClimART, which is designed specifically for the narrower and less computationally costly task of radiative energy transfer parameterization, rather than cloud and rain emulators. ClimateBench, on the other hand, is not an attempt at hybrid simulation, but rather for ``whole-model'' emulators that reproduce the annual mean global predictions of climate that a conventional climate model would simulate given unseen greenhouse gas concentrations and emissions. This does not attempt to sidestep Moore's Law or admit previously unattainable resolution, i.e., any error or bias related to the parameterizations used to create the training data are part of what is learned by the emulator. 

In contrast, the goal of \dataset{} is to develop an emulator for the \textit{explicitly resolved} effect of clouds and storms on climate, so that, down the road, the emulator can be used to replace parameterizations in a climate model, enabling more realistic climate simulation without the typical computational overhead. \dataset{} builds off work by a few climate scientists who have been exploring since 2017 to apply ML for hybrid multi-scale climate modeling. \cite{Gentine2018} first demonstrated that using simple ML models, and a simple atmosphere test-bed, certain atmospheric patterns of convective heating and moistening could be effectively predicted, particularly in the tropics and mid-latitude storm tracks. However, when these models were integrated into broader climate simulations, except for lucky fits that demonstrated the exciting potential for success \citep{Rasp2018}, issues related to stability arose, a common problem when constructing hybrid climate models. Various methods were tried to improve the stability, such as coupling multiple models together and searching for better model architectures \citep{Brenowitz2020, ott2020}. These efforts led to improved error rates in the predictions. More recently, researchers have expanded this work into real-world settings, using more advanced ML architectures \citep{Han2020, Mooers2021, Wang2022a, han2023ensemble}. \cite{Wang2022b} and \cite{han2023ensemble} even managed to create a deep-learning model that showed hybrid stability over 5 years to a decade under real-world conditions. While these hybrid models had a few biases, they were successful in capturing some aspects of climate variability. Additionally, work has been done to compress input data to avoid causal confounders while maintaining accuracy \citep{Iglesias2023}, use latent representations that account for stochasticity \citep{Behrens2022}, and enforce physical constraints within these models \citep{Beucler2021a}, all of which could potentially improve their reliability.

% \clearpage
\FloatBarrier
% ======================================================
\section{Extra Figures and Tables}
\label{sec:ExtraFigures}
% ======================================================
\subsection{MLP with Expanded Target Variables}

\begin{table}[H]
\centering
{\fontsize{9}{10.3}\selectfont
\begin{tabular}{llcccc}
\toprule
& (Variables) & MLPv1 & MLPv2 & MLPv1-ne30 & MLPv2-ne30\\
\midrule
\midrule
\multirow{14}{*}{\textbf{MAE}}
& dT/dt & 2.688 & 2.305 & 2.799 & 2.886 \\
& dq/dt & 4.503 & 4.030 & 4.231 & 4.068 \\
& dq$_\text{l}$/dt & N/A & 0.689 & N/A & 0.697 \\
& dq$_\text{i}$/dt & N/A & 0.384 & N/A & 0.330 \\
& du/dt & N/A & 1.34E-04 & N/A & 2.68E-04 \\
& dv/dt & N/A & 1.09E-04 & N/A & 2.66E-04 \\
& NETSW & 13.47 & 8.339 & 15.47 & 11.04 \\
& FLWDS & 5.118 & 4.134 & 5.318 & 4.891 \\
& PRECSC & 2.645 & 1.539 & 3.115 & 3.009 \\
& PRECC & 33.89 & 23.74 & 42.49 & 29.62 \\
& SOLS & 7.942 & 5.774 & 8.484 & 6.866 \\
& SOLL & 10.30 & 8.190 & 10.582 & 8.993 \\
& SOLSD & 4.587 & 3.230 & 5.056 & 4.360 \\
& SOLLD & 4.834 & 3.977 & 4.963 & 4.553 \\
\midrule
\midrule
\multirow{14}{*}{\textbf{R2}}
& dT/dt & 0.590 & 0.663 & 0.626 & 0.536 \\
& dq/dt & - & - & - & - \\
& dq$_\text{l}$/dt & N/A & - & N/A & - \\
& dq$_\text{i}$/dt & N/A & - & N/A & - \\
& du/dt & N/A & - & N/A & - \\
& dv/dt & N/A & - & N/A & - \\
& NETSW & 0.982 & 0.993 & 0.977 & 0.988 \\
& FLWDS & 0.927 & 0.945 & 0.914 & 0.924 \\
& PRECSC & - & - & -0.117 & -0.117 \\
& PRECC & -1.494 & 0.833 & -0.115 & -0.115 \\
& SOLS & 0.962 & 0.978 & 0.963 & 0.976 \\
& SOLL & 0.948 & 0.964 & 0.953 & 0.965 \\
& SOLSD & 0.955 & 0.976 & 0.950 & 0.965 \\
& SOLLD & 0.866 & 0.905 & 0.874 & 0.899 \\
\midrule
\midrule
\multirow{14}{*}{\textbf{RMSE}}
& dT/dt & 4.437 & 3.756 & 5.199 & 4.958 \\
& dq/dt & 7.337 & 6.521 & 7.550 & 7.135 \\
& dq$_\text{l}$/dt &  & 1.192 &  & 1.489 \\
& dq$_\text{i}$/dt &  & 0.812 &  & 0.940 \\
& du/dt &  & 2.80E-04 &  & 6.45E-04 \\
& dv/dt &  & 2.25E-04 &  & 6.72E-04 \\
& NETSW & 26.95 & 17.24 & 30.48 & 21.18 \\
& FLWDS & 6.803 & 5.532 & 7.136 & 6.540 \\
& PRECSC & 4.656 & 2.955 & 7.791 & 7.509 \\
& PRECC & 73.16 & 53.47 & 119.8 & 83.22 \\
& SOLS & 17.39 & 12.84 & 18.51 & 14.74 \\
& SOLL & 21.96 & 17.89 & 22.71 & 19.27 \\
& SOLSD & 9.474 & 6.837 & 10.42 & 8.724 \\
& SOLLD & 10.14 & 8.486 & 10.62 & 9.526 \\
\bottomrule
\end{tabular}
}

\caption{Similar to Table 2 in the main text but for comparing MAR, R2, and RMSE of different MLP models: MLP v1 (subset emulation) and the MLP v2 (full vector emulation) built with the low-resolution (ne4) and the high-resolution datasets (ne30). dq$_\text{l}$/dt, dq$_\text{i}$/dt, du/dt, and dv/dt correspond to the tendencies of state\_q0002, state\_q0003, state\_u, and state\_v, respectively, in Table SI1.}
\label{tab:MLP_MAE}
\end{table}

\begin{figure}[H]
    \centering
    \includegraphics[width=.85\textwidth]{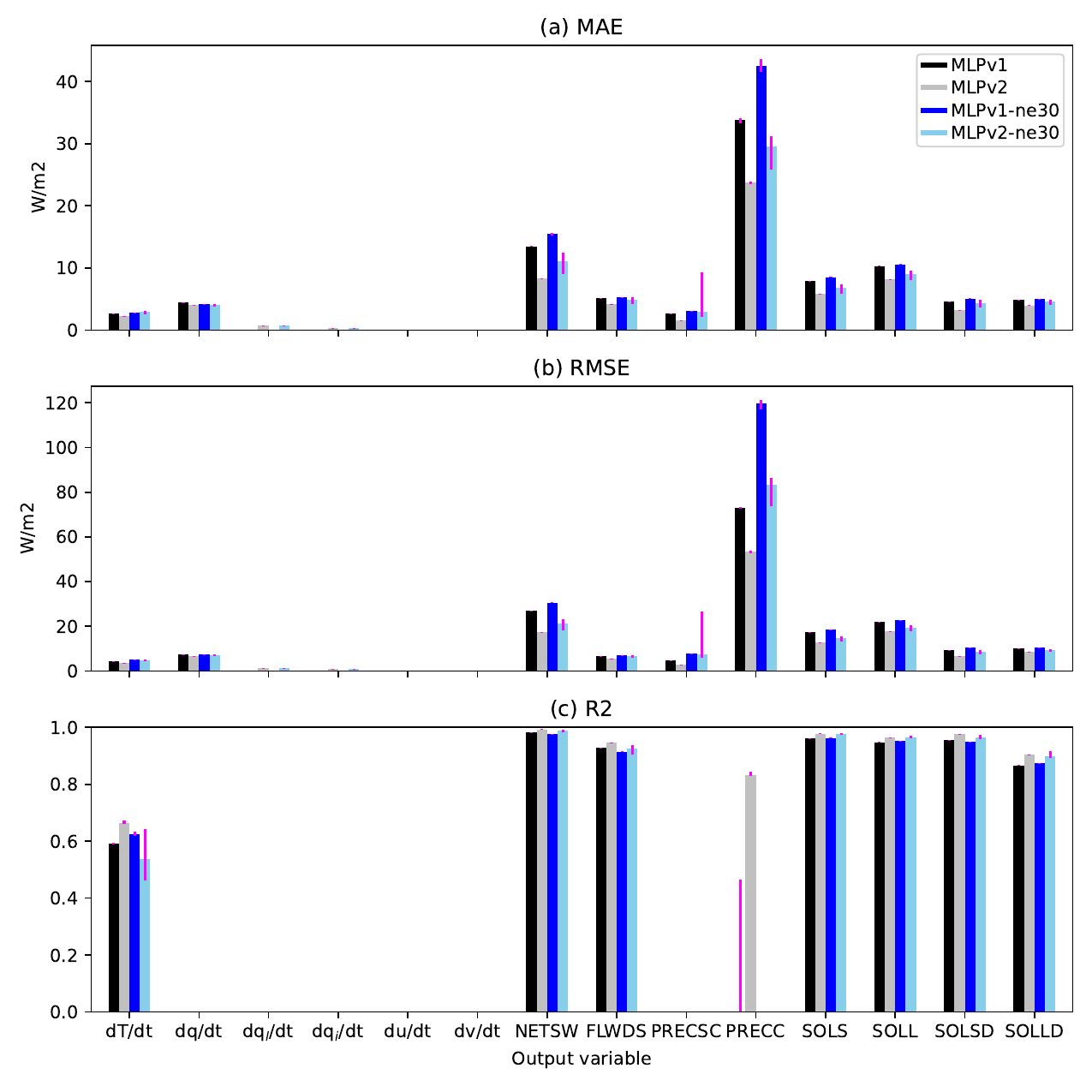}
    \caption{Equivalent to Figure S3, but for comparing the MLPv1 (subset emulation) and the MLPv2 (full vector emulation). In addition, MLP models trained with the high-resolution dataset (ne30) are shown here: MLPv1-ne30 and MLPv2-ne30. Bars show the median of the performance of top-20 models selected from the hyperparamter search (>8,000 trials), and magenta error bars show the range of the top-20 model performance.}
    \label{fig:SI_MLPv2_avg}
\end{figure}

\begin{figure}[H]
    \centering
    \includegraphics[width=.875\textwidth, trim={0.1cm 2.5cm 1cm 3.5cm},clip]{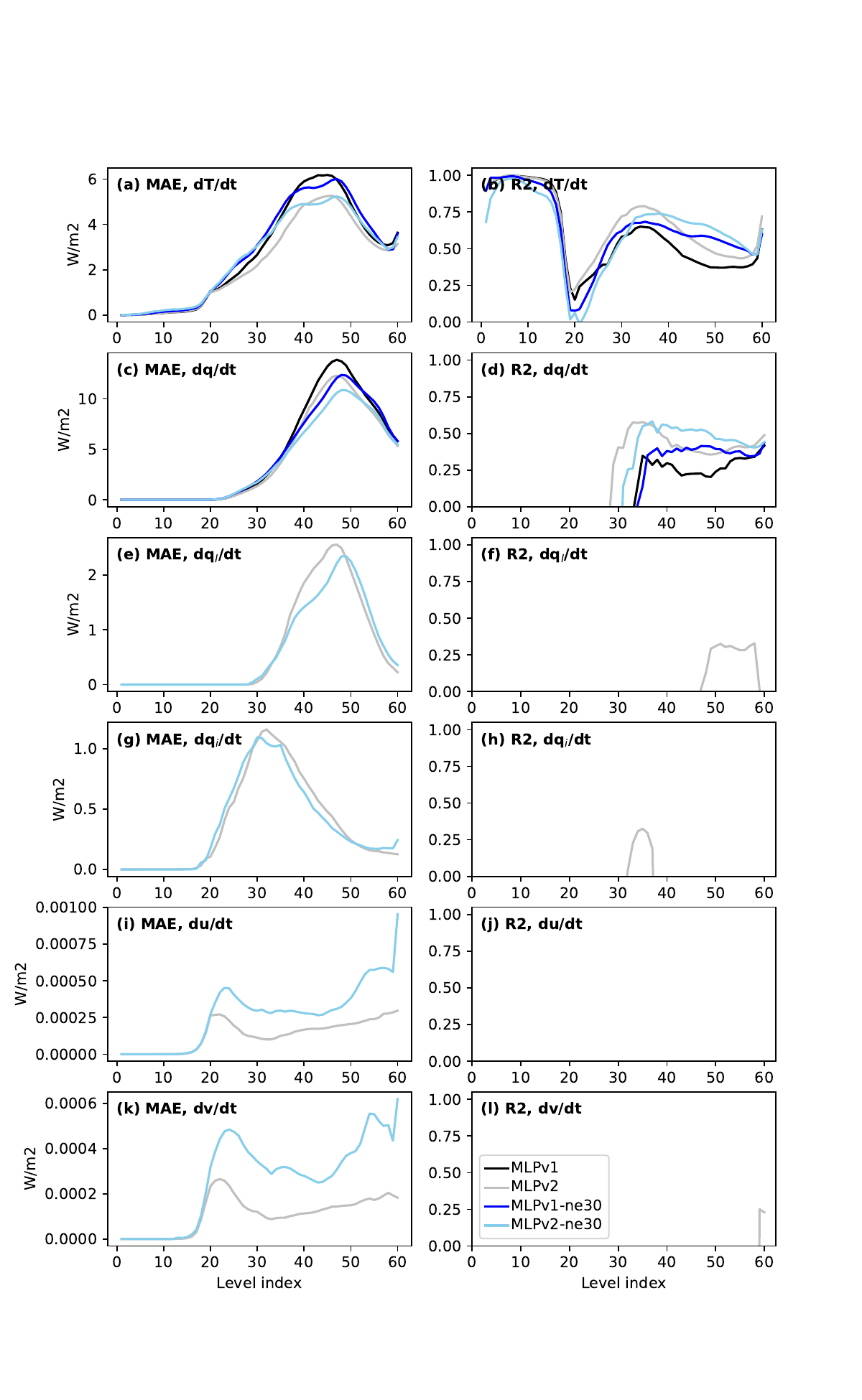}
    \caption{Equivalent to Figure 2, but for comparing the MLP v1 (subset emulation) and the MLP v2 (full vector emulation). In addition, MLP models trained with the high-resolution dataset (ne30) are shown here: MLPv1-ne30 and MLPv2-ne30. Out of the top model pools, MLP models shown in this figure are randomly chosen for visualizatoin.}
    \label{fig:SI_MLPv2_vert}
\end{figure}

\subsection{Scatter Plots}
\begin{figure}[H]
    \centering
    \includegraphics[width=.88\textwidth, trim={.7cm .5cm 0.cm .6cm},clip]{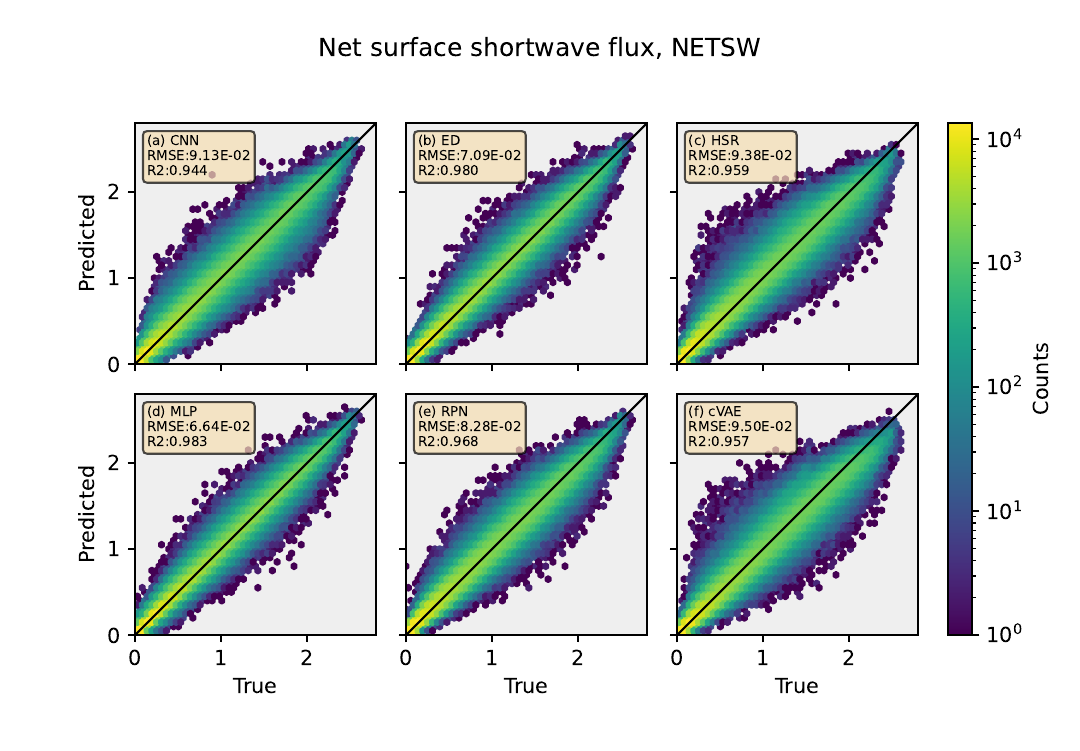}
    \includegraphics[width=.88\textwidth, trim={.7cm .5cm 0.cm .5cm},clip]{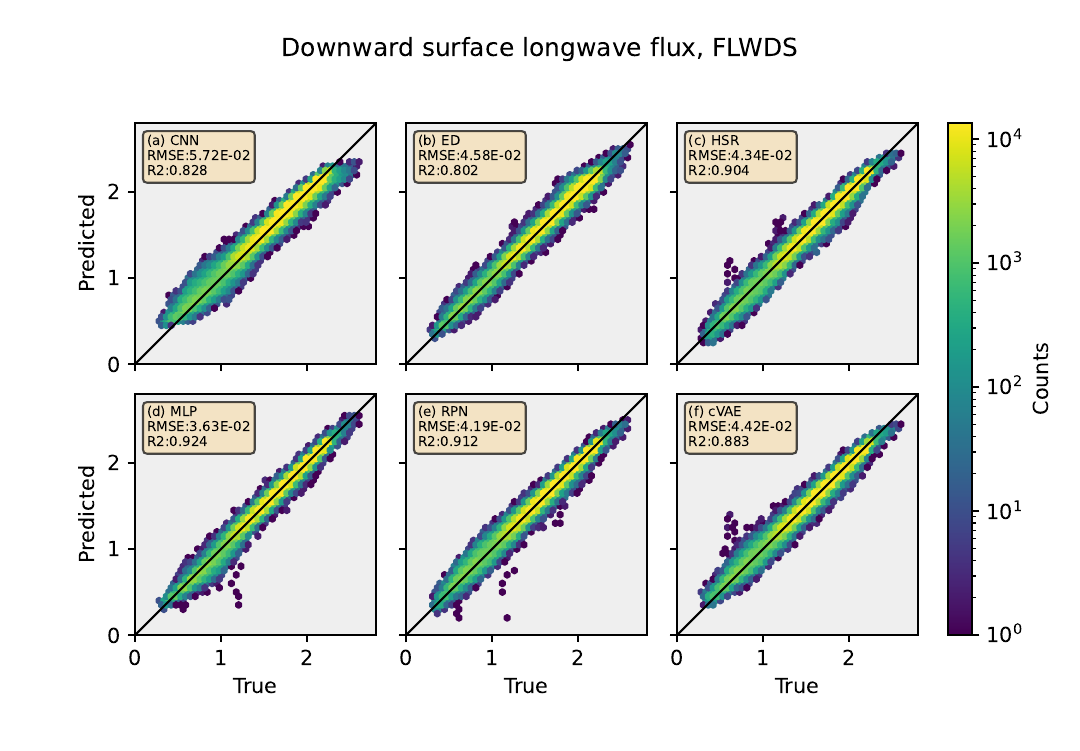}
     
    \caption{Hexagonally-binned representation of 2D target variables comparing the climate model simulation (``true"; x-axis) with the ML model prediction (``predicted"; y-axis). The color of each hexagonal bin corresponds to the number of data points enclosed.}
    \label{fig:SI_Scatter_2D_part1}
\end{figure}

\begin{figure}[H]
    \centering
    \includegraphics[width=.88\textwidth, trim={.7cm .1cm 0.cm 0},clip]{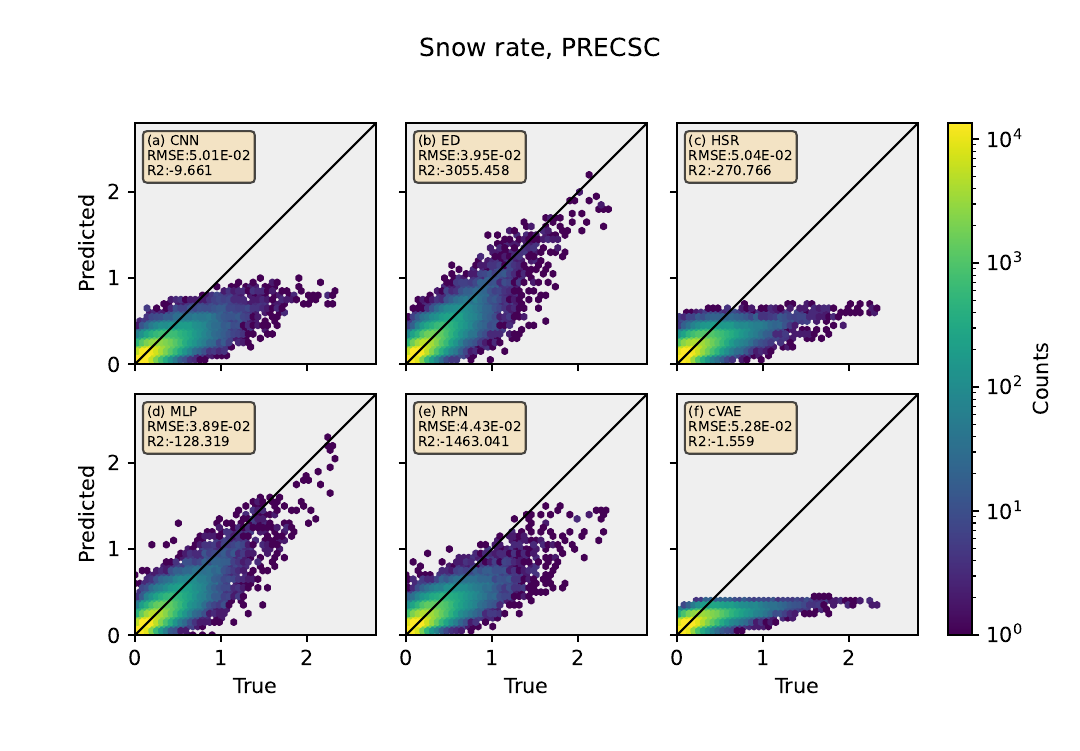}
    \includegraphics[width=.88\textwidth, trim={.7cm .1cm 0.cm 0},clip]{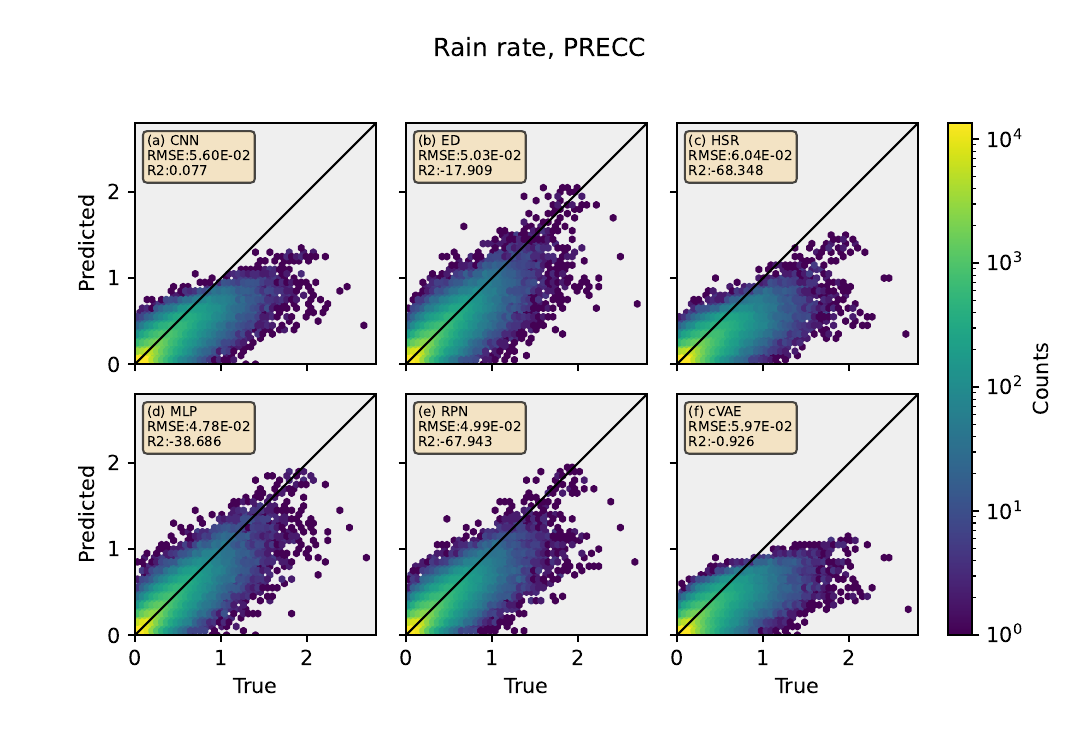}
     
    \caption{Hexagonally-binned representation of 2D target variables comparing the climate model simulation (``true"; x-axis) with the ML model prediction (``predicted"; y-axis). The color of each hexagonal bin corresponds to the number of data points enclosed.}
    \label{fig:SI_Scatter_2D_part2}
\end{figure}

\begin{figure}[H]
    \centering
    \includegraphics[width=.88\textwidth, trim={.7cm .1cm 0.cm 0},clip]{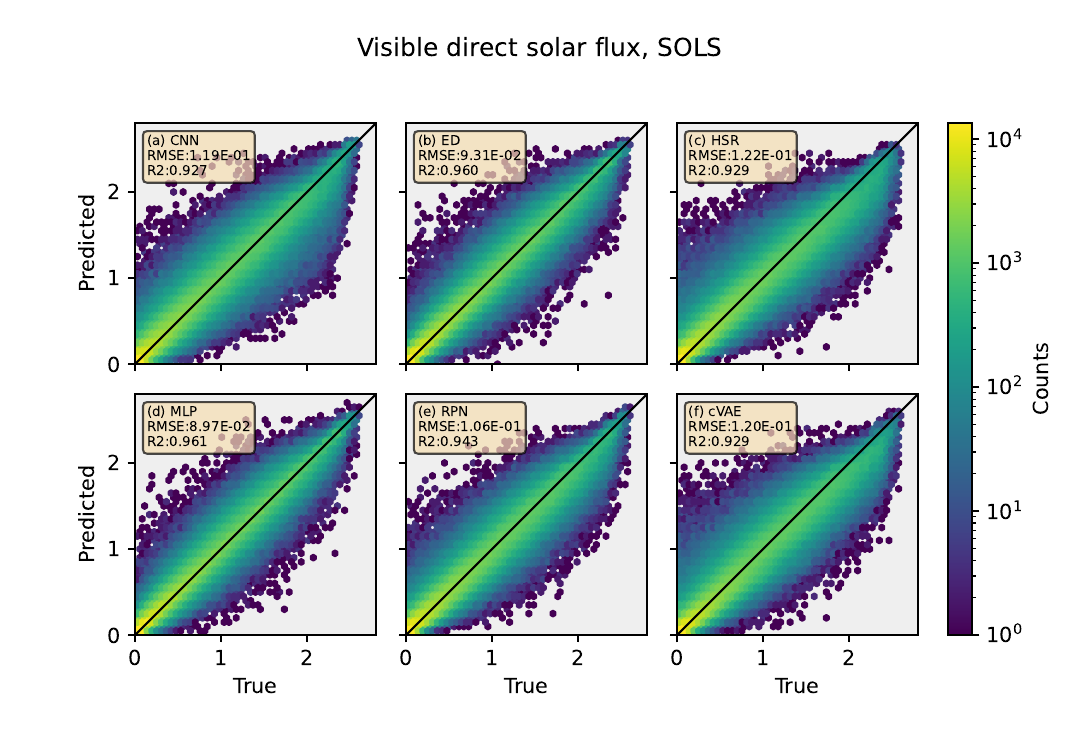}
    \includegraphics[width=.88\textwidth, trim={.7cm .1cm 0.cm 0},clip]{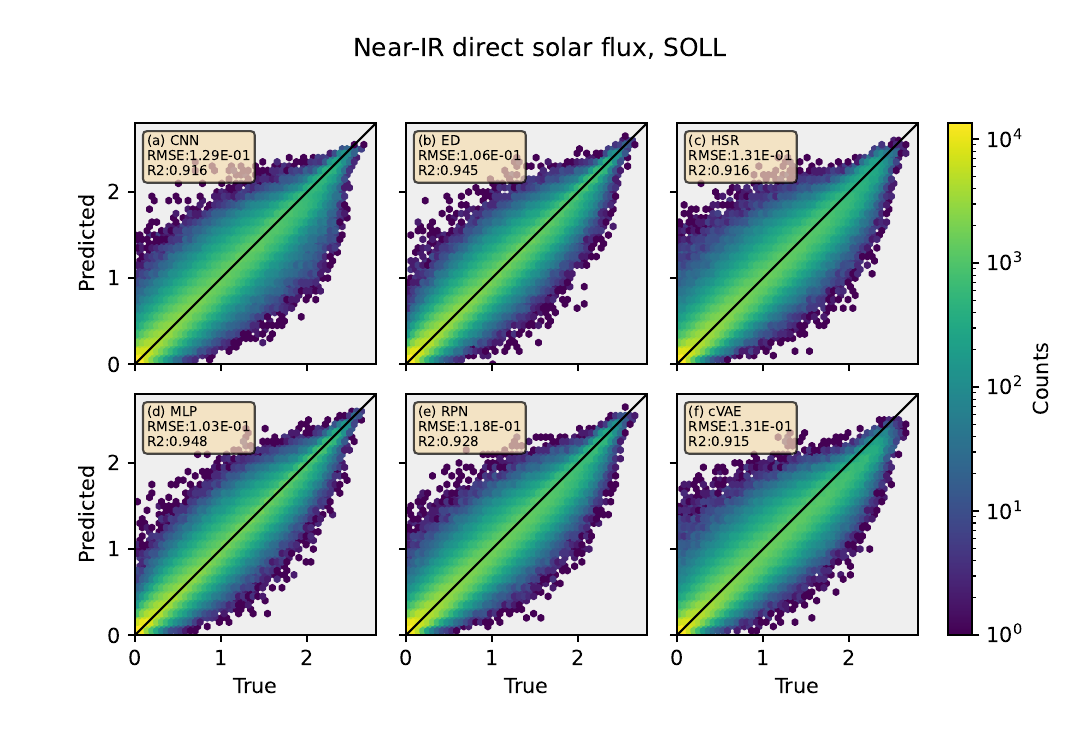}
     
    \caption{Hexagonally-binned representation of 2D target variables comparing the climate model simulation (``true"; x-axis) with the ML model prediction (``predicted"; y-axis). The color of each hexagonal bin corresponds to the number of data points enclosed.}
    \label{fig:SI_Scatter_2D_part3}
\end{figure}

\begin{figure}[H]
    \centering
    \includegraphics[width=.88\textwidth, trim={.7cm .1cm 0.cm 0},clip]{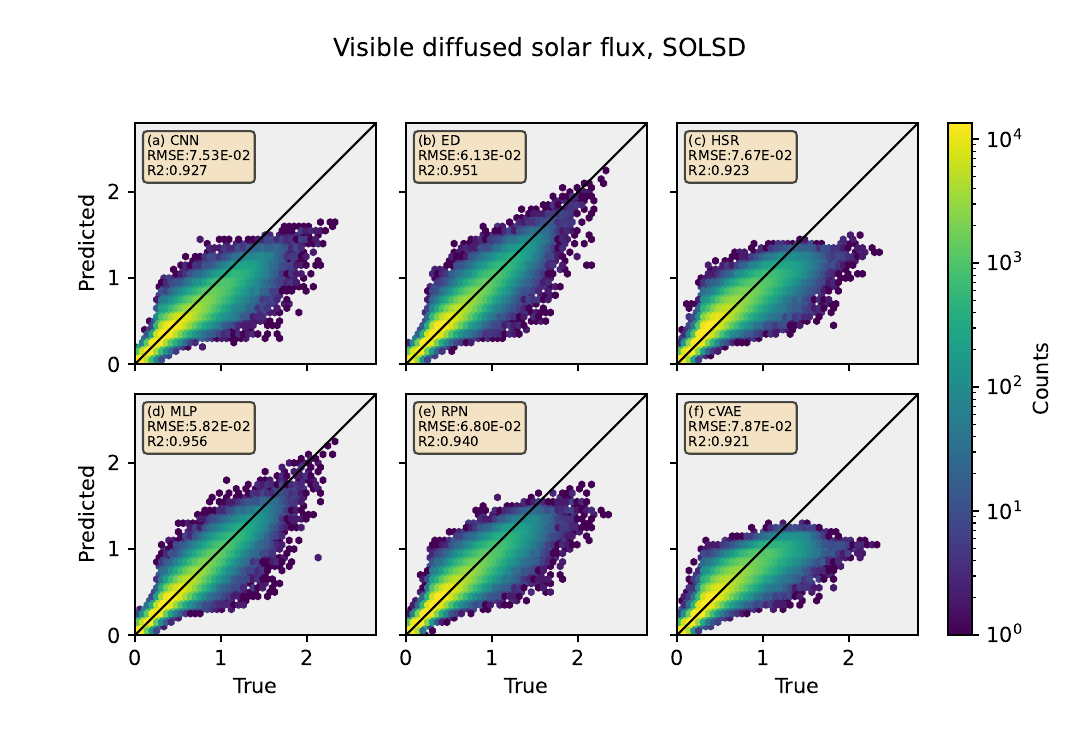}
    \includegraphics[width=.88\textwidth, trim={.7cm .1cm 0.cm 0},clip]{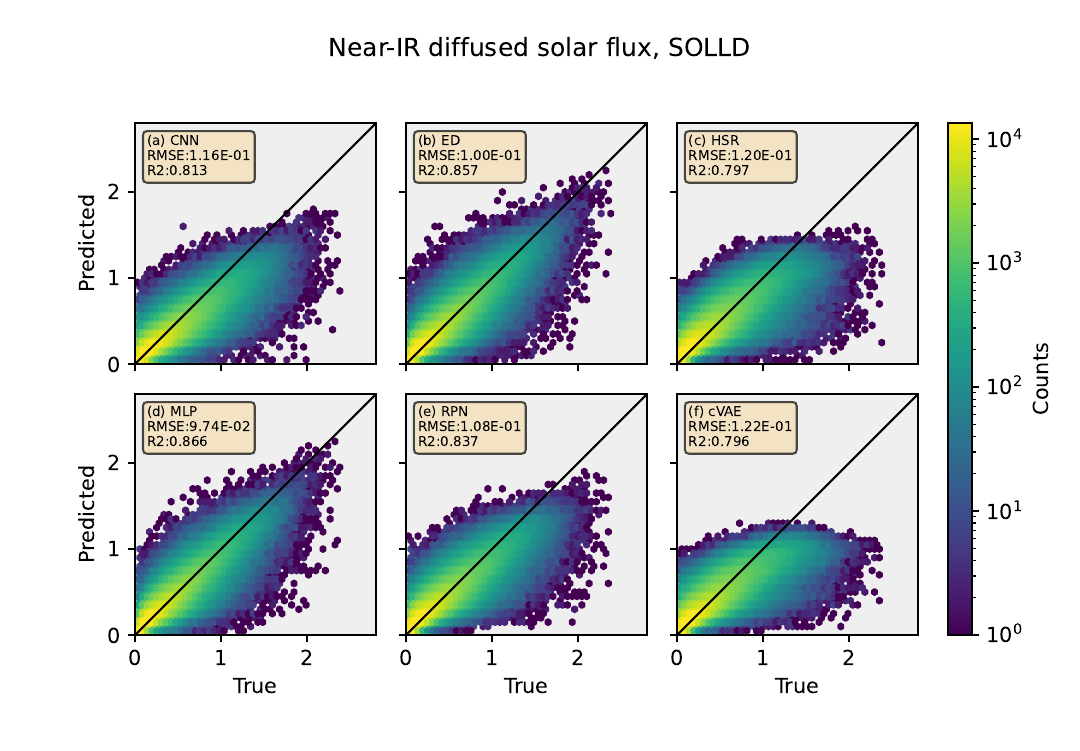}
     
    \caption{Hexagonally-binned representation of 2D target variables comparing the climate model simulation (``true"; x-axis) with the ML model prediction (``predicted"; y-axis). The color of each hexagonal bin corresponds to the number of data points enclosed.}
    \label{fig:SI_Scatter_2D_part4}
\end{figure}

\begin{figure}[H]
    \centering
    \includegraphics[width=.88\textwidth, trim={.7cm .1cm 0.cm 0},clip]{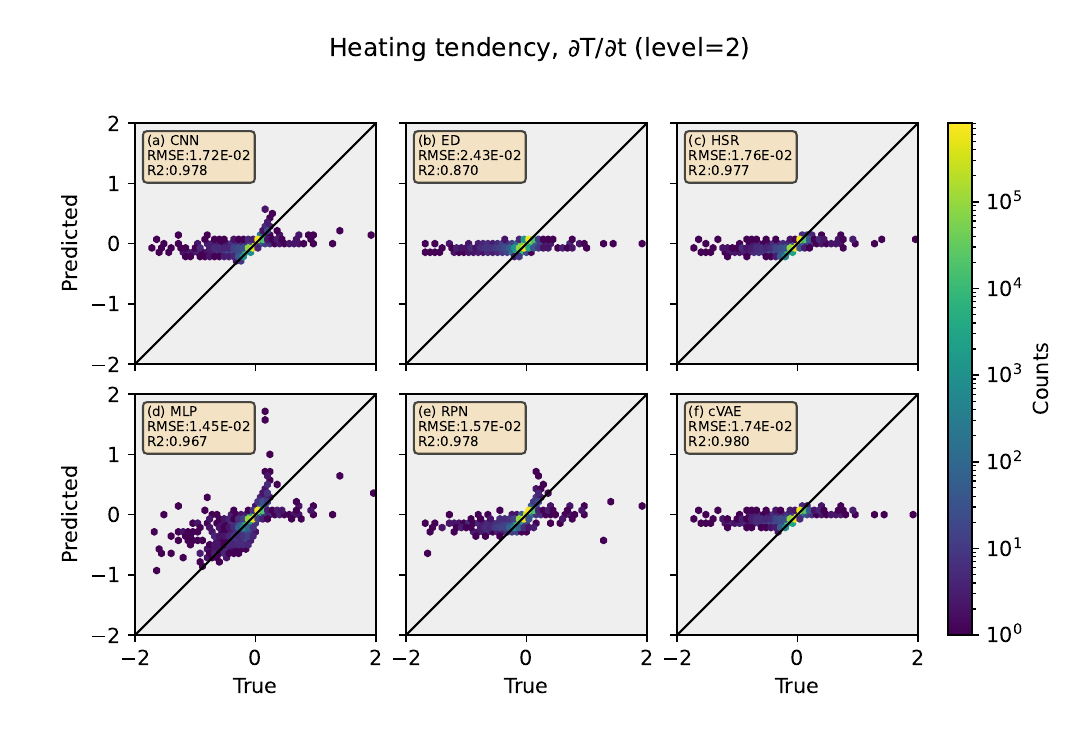}
    \includegraphics[width=.88\textwidth, trim={.7cm .1cm 0.cm 0},clip]{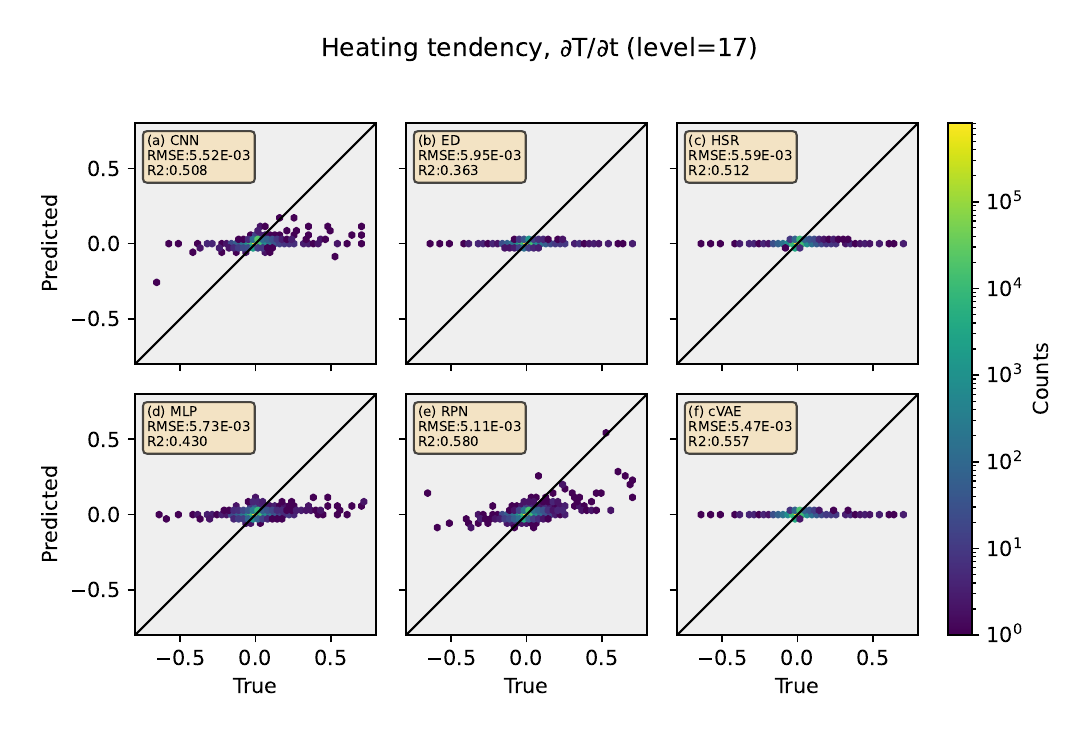}
     
    \caption{Hexagonally-binned representation of 3D (vertically-resolved) target variables comparing the climate model simulation (``true"; x-axis) with the ML model prediction (``predicted"; y-axis) at four different vertical levels. The color of each hexagonal bin corresponds to the number of data points enclosed.}
    \label{fig:SI_Scatter_3D_part1}
\end{figure}

\begin{figure}[H]
    \centering
    \includegraphics[width=.88\textwidth, trim={.7cm .1cm 0.cm 0},clip]{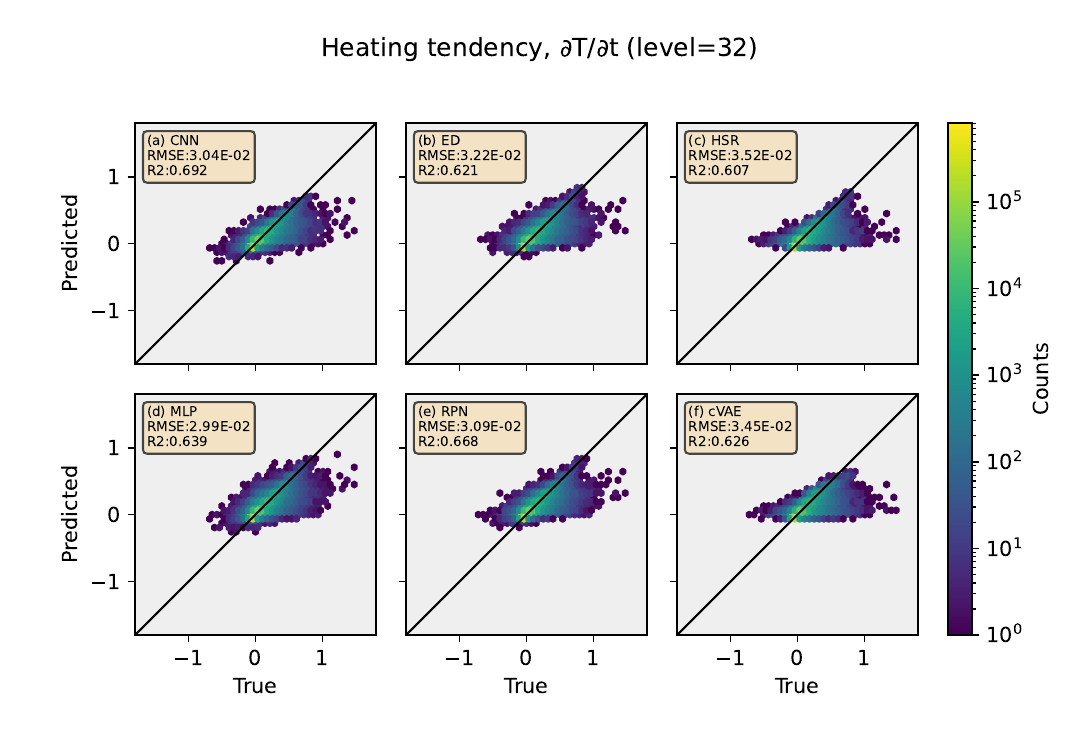}
    \includegraphics[width=.88\textwidth, trim={.7cm .1cm 0.cm 0},clip]{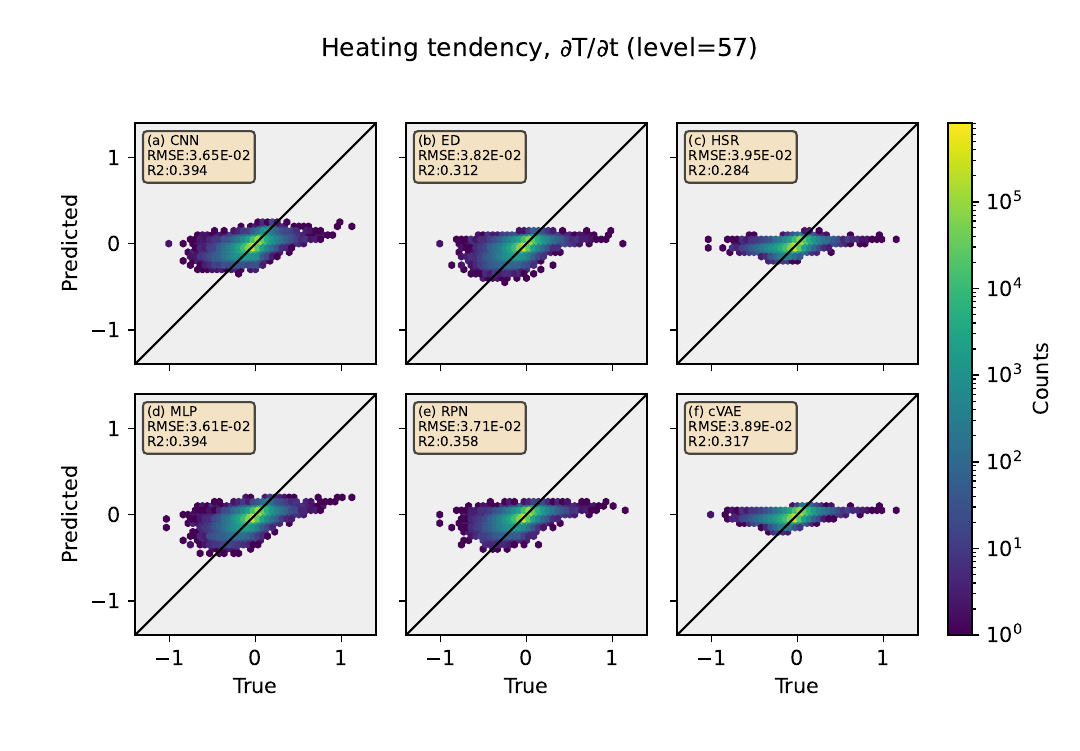}
     
    \caption{Hexagonally-binned representation of 3D (vertically-resolved) target variables comparing the climate model simulation (``true"; x-axis) with the ML model prediction (``predicted"; y-axis) at four different vertical levels. The color of each hexagonal bin corresponds to the number of data points enclosed.}
    \label{fig:SI_Scatter_3D_part2}
\end{figure}

\begin{figure}[H]
    \centering
    \includegraphics[width=.88\textwidth, trim={.7cm .1cm 0.cm 0},clip]{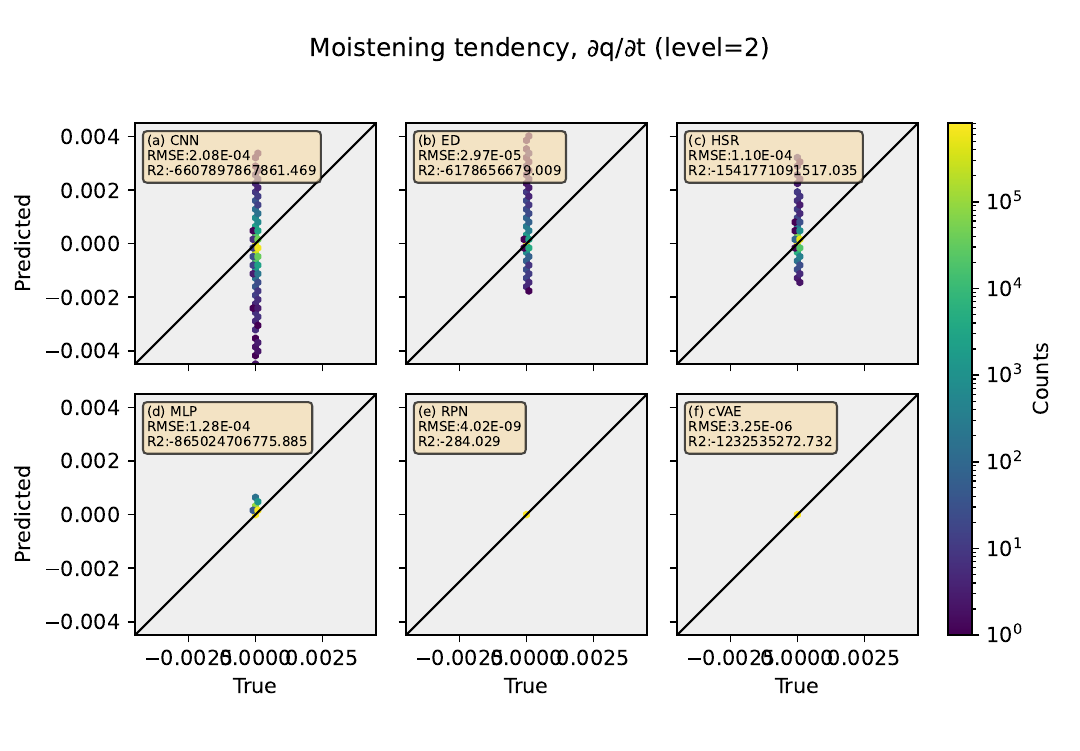}
    \includegraphics[width=.88\textwidth, trim={.7cm .1cm 0.cm 0},clip]{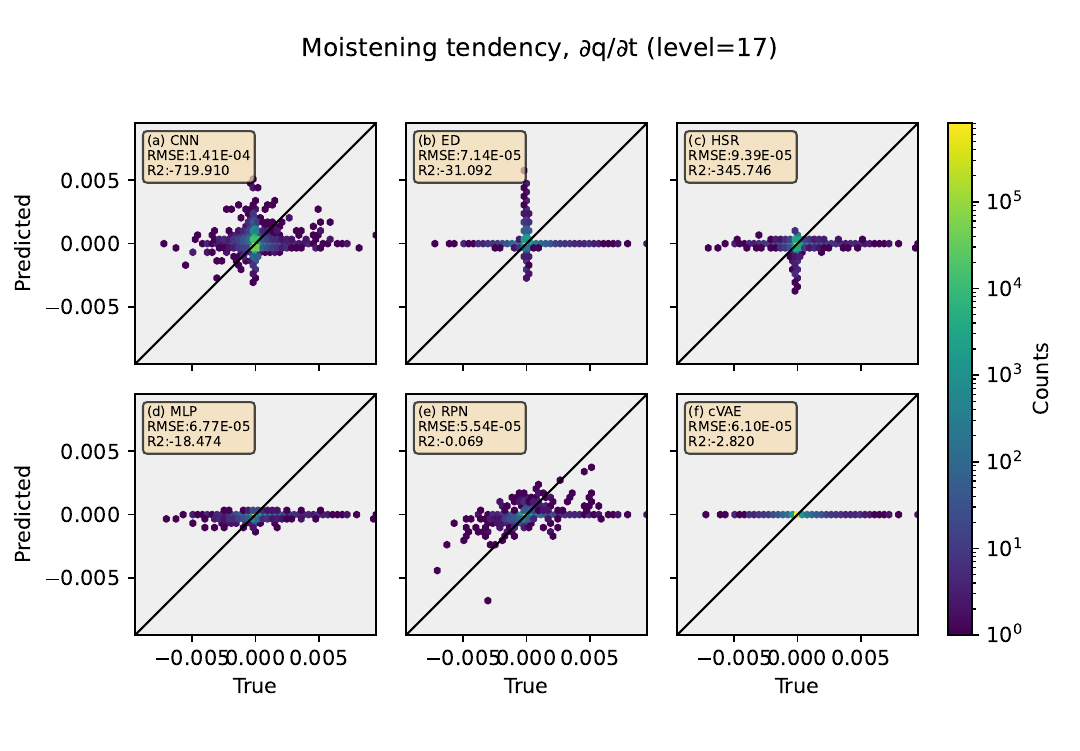}
     
    \caption{Hexagonally-binned representation of 3D (vertically-resolved) target variables comparing the climate model simulation (``true"; x-axis) with the ML model prediction (``predicted"; y-axis) at four different vertical levels. The color of each hexagonal bin corresponds to the number of data points enclosed.}
    \label{fig:SI_Scatter_3D_part3}
\end{figure}

\begin{figure}[H]
    \centering
    \includegraphics[width=.88\textwidth, trim={.7cm .1cm 0.cm 0},clip]{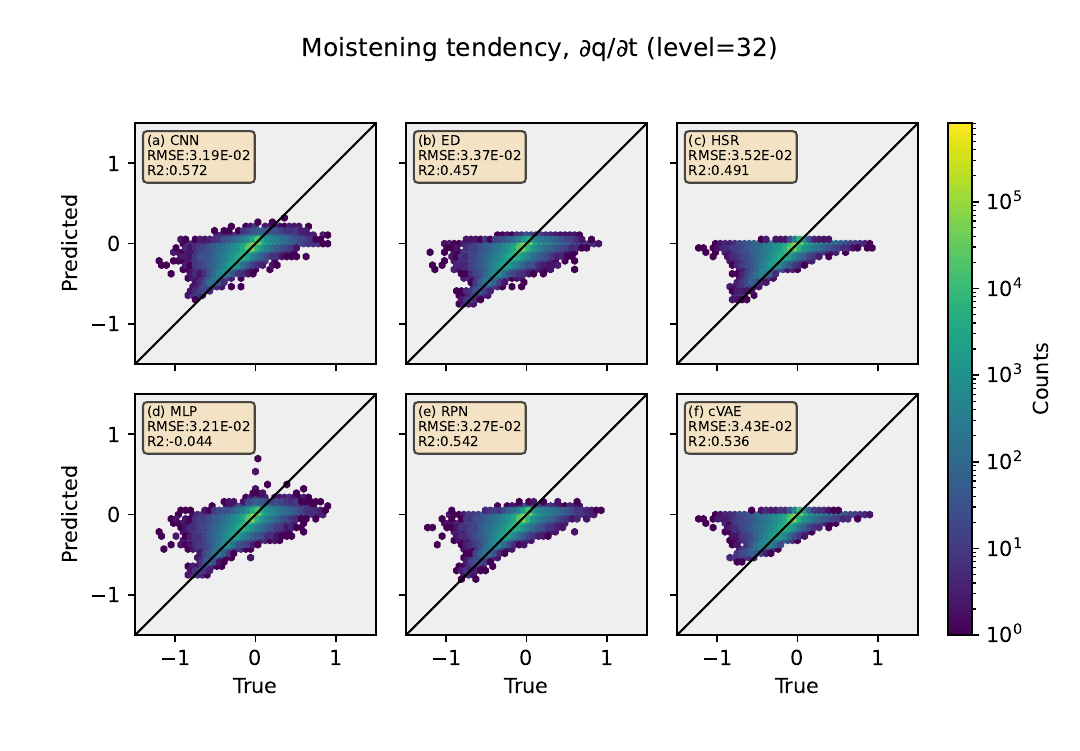}
    \includegraphics[width=.88\textwidth, trim={.7cm .1cm 0.cm 0},clip]{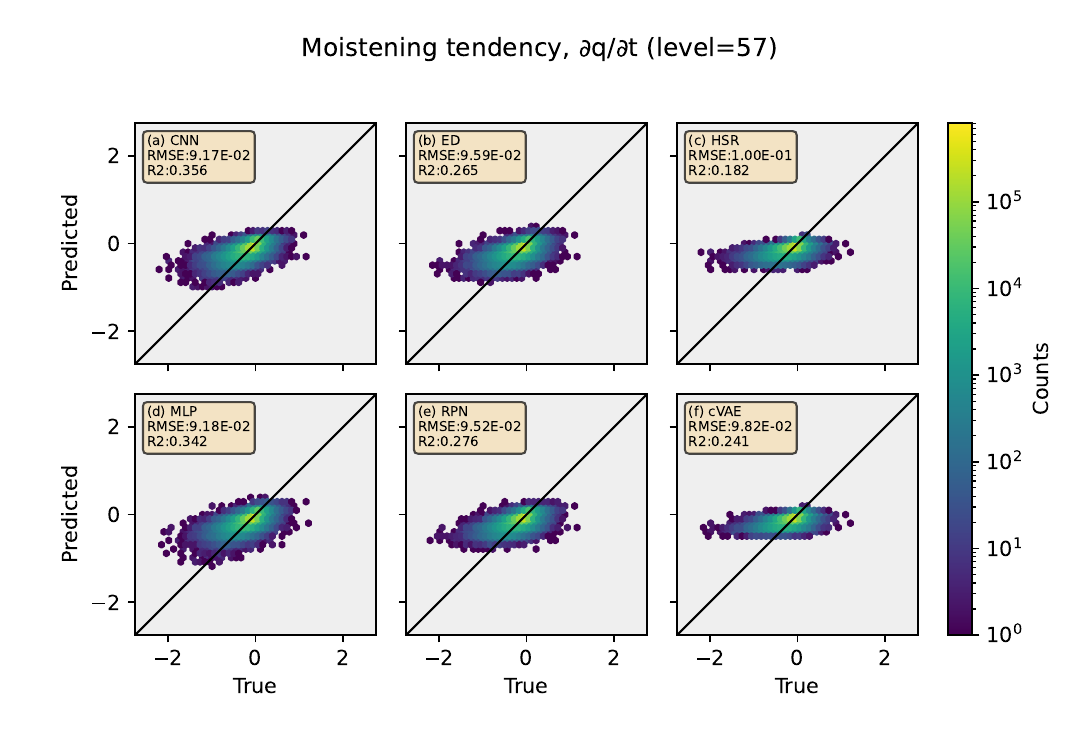}
     
    \caption{Hexagonally-binned representation of 3D (vertically-resolved) target variables comparing the climate model simulation (``true"; x-axis) with the ML model prediction (``predicted"; y-axis) at four different vertical levels. The color of each hexagonal bin corresponds to the number of data points enclosed.}
    \label{fig:SI_Scatter_3D_part4}
\end{figure}

\subsection{Global Maps of R$^2$}
\begin{figure}[H]
    \centering
    \includegraphics[width=.83\textwidth, trim={1.1cm .3cm 0 .3cm},clip]{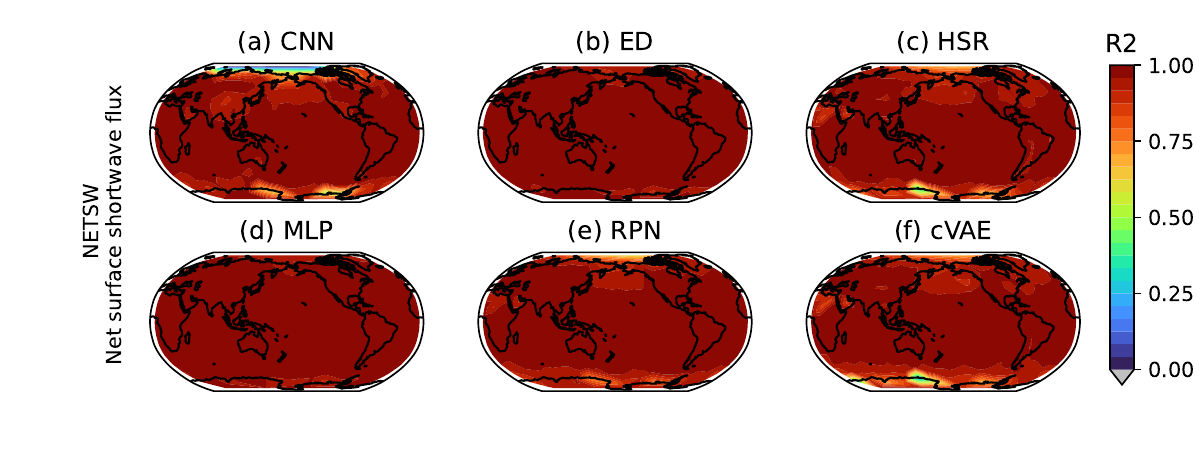}
    \includegraphics[width=.83\textwidth, trim={1.1cm .3cm 0 .3cm},clip]{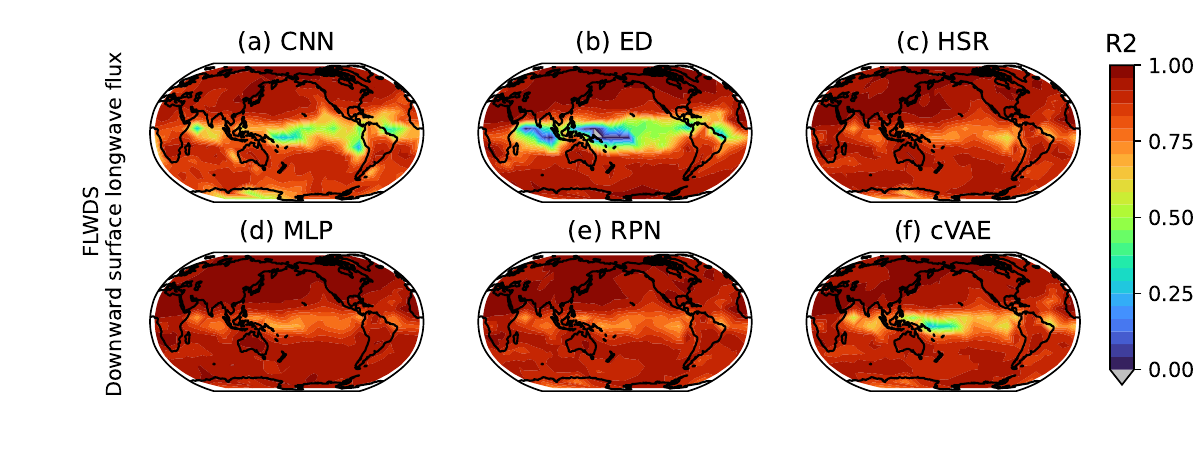}
    \includegraphics[width=.83\textwidth, trim={1.1cm .3cm 0 .3cm},clip]{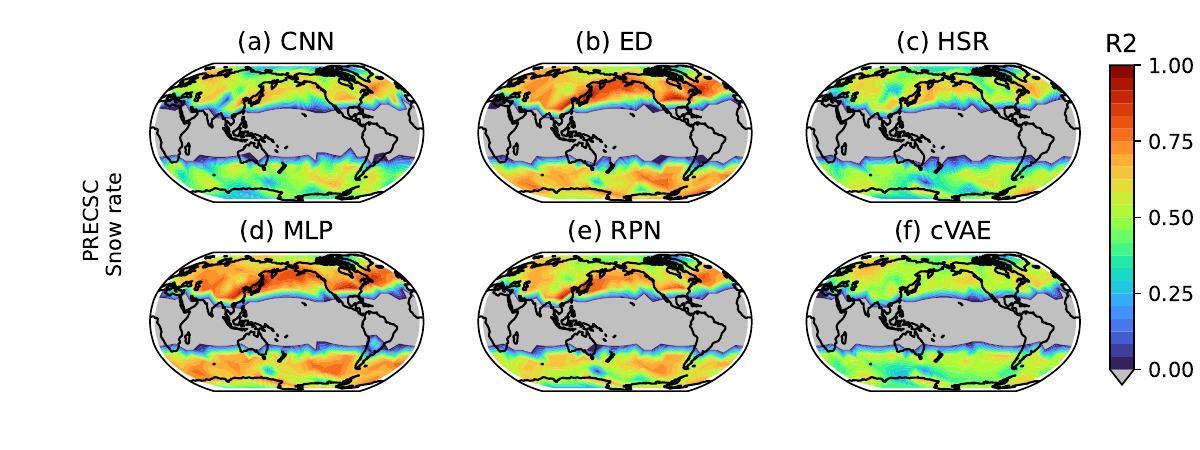}
    \includegraphics[width=.83\textwidth, trim={1.1cm .3cm 0 .3cm},clip]{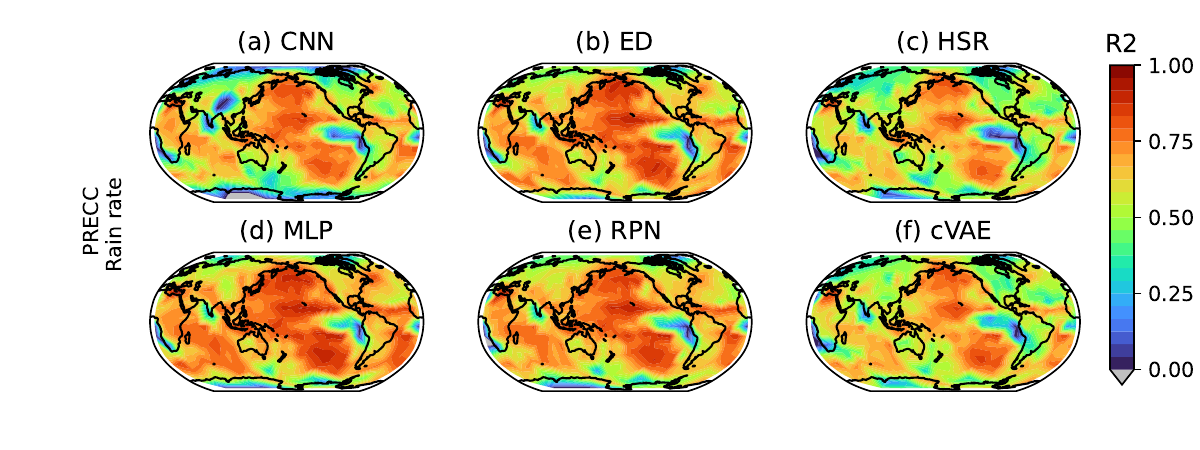}
     
    \caption{Global maps of R$^2$ of baseline models (built on the low-res, real-geography dataset). Grey shading shows locations with negative R$^2$ values.}
    \label{fig:SI_MAP_R2_2D_part1}
\end{figure}

\begin{figure}[H]
    \centering
    \includegraphics[width=.83\textwidth, trim={1.1cm 0.cm 0 0.cm},clip]{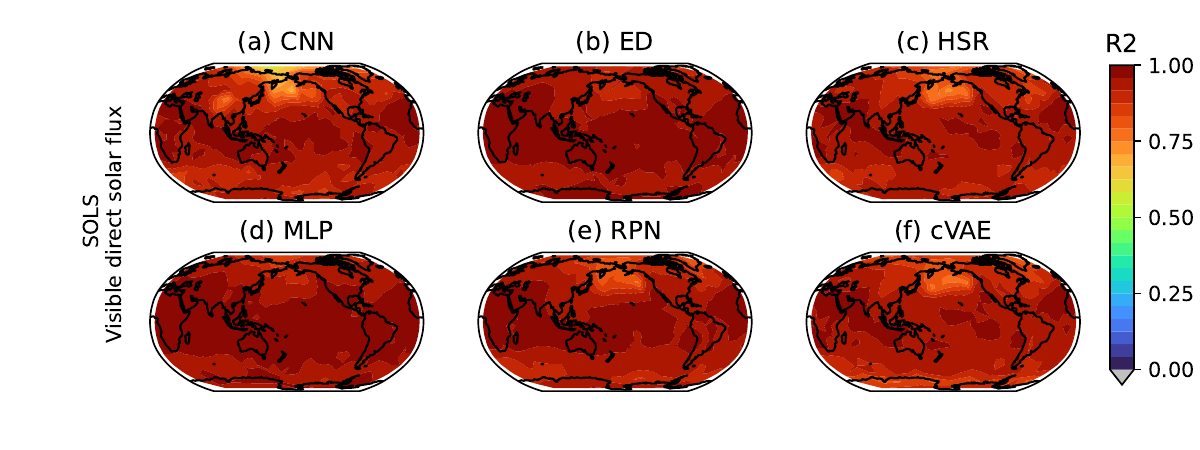}
    \includegraphics[width=.83\textwidth, trim={1.1cm 0.cm 0 0.cm},clip]{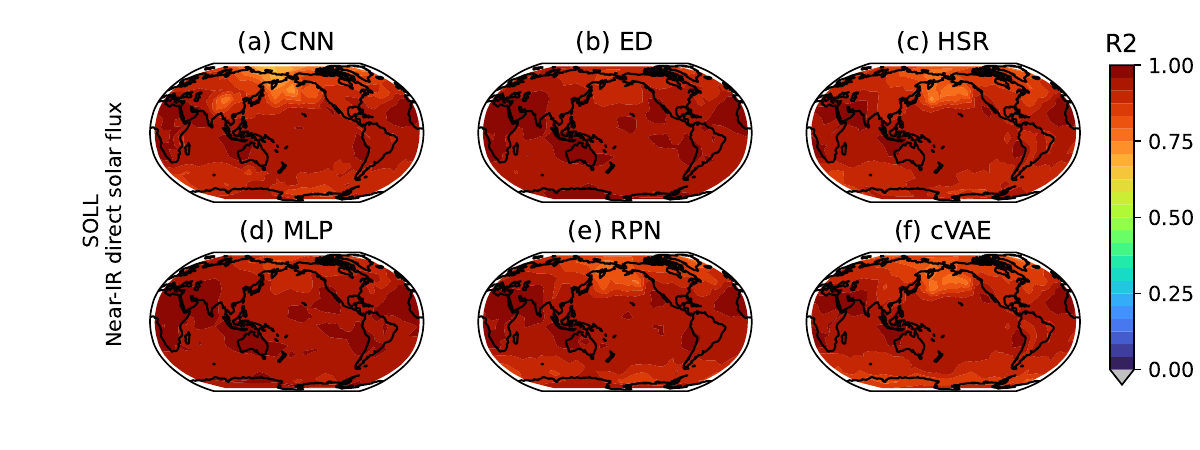}
    \includegraphics[width=.83\textwidth, trim={1.1cm 0.cm 0 0.cm},clip]{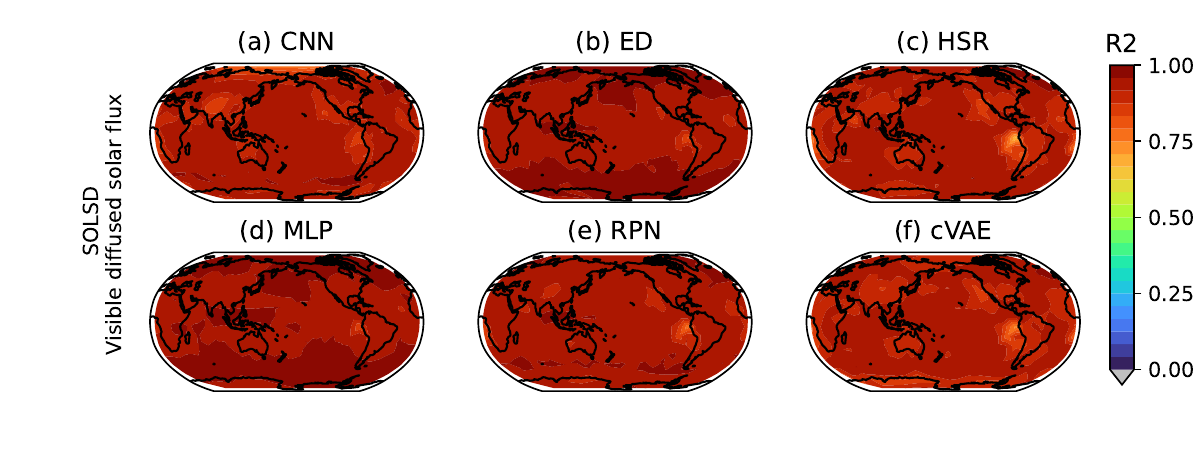}
    \includegraphics[width=.83\textwidth, trim={1.1cm 0.cm 0 0.cm},clip]{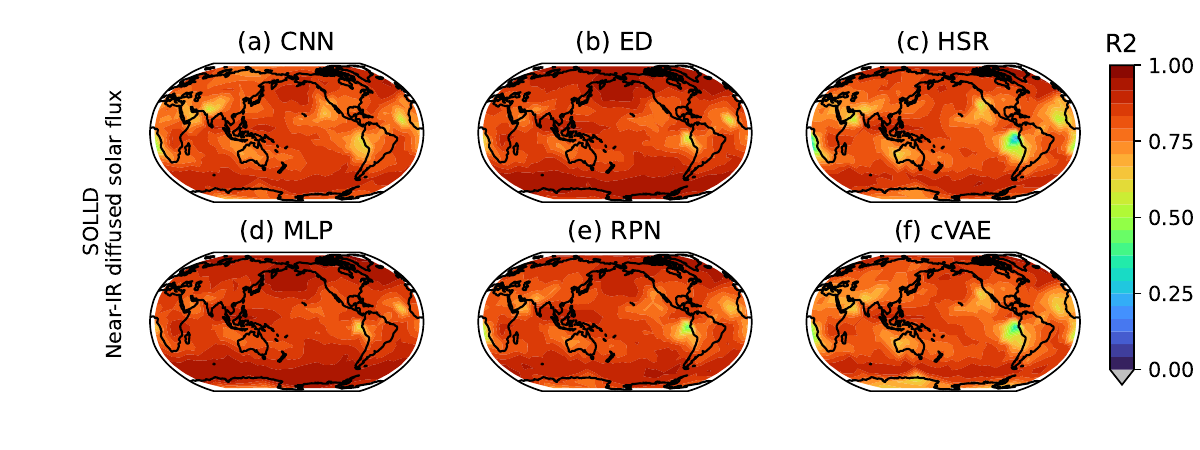}
     
    \caption{Global maps of R$^2$ of baseline models (built on the low-res, real-geography dataset). Grey shading shows locations with negative R$^2$ values.}
    \label{fig:SI_MAP_R2_2D_part2}
\end{figure}

\begin{figure}[H]
    \centering
    \includegraphics[width=.83\textwidth, trim={1.1cm 0.cm 0 0.cm},clip]{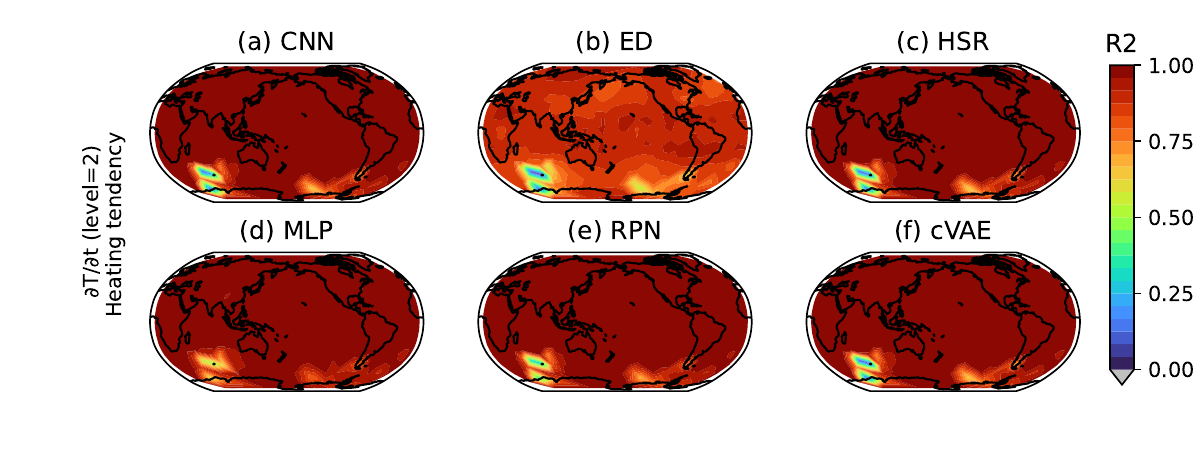}
    \includegraphics[width=.83\textwidth, trim={1.1cm 0.cm 0 0.cm},clip]{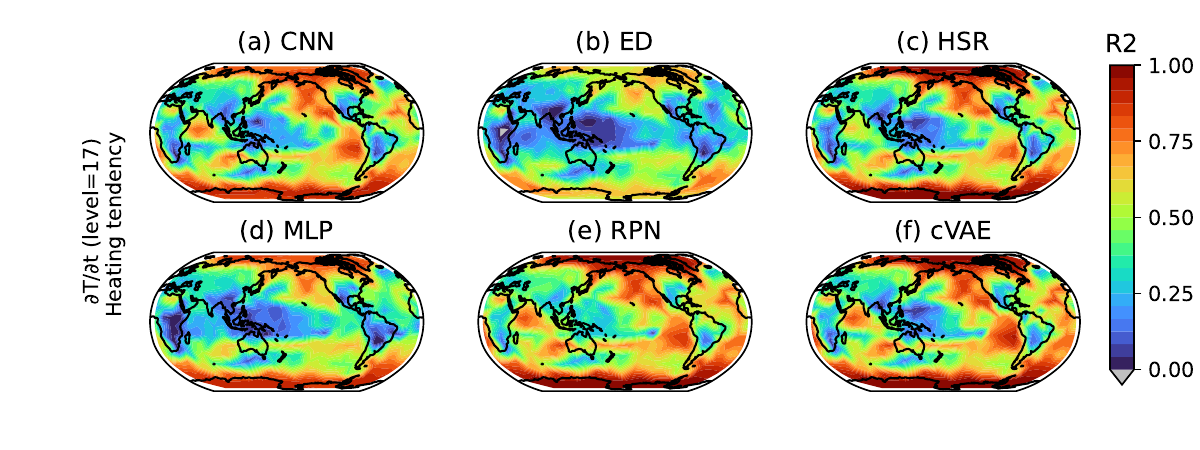}
    \includegraphics[width=.83\textwidth, trim={1.1cm 0.cm 0 0.cm},clip]{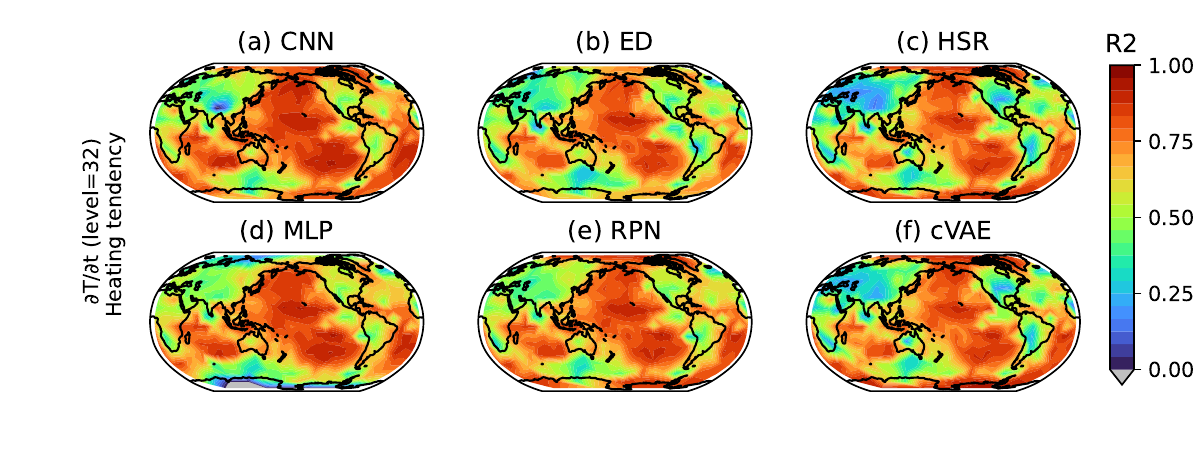}
    \includegraphics[width=.83\textwidth, trim={1.1cm 0.cm 0 0.cm},clip]{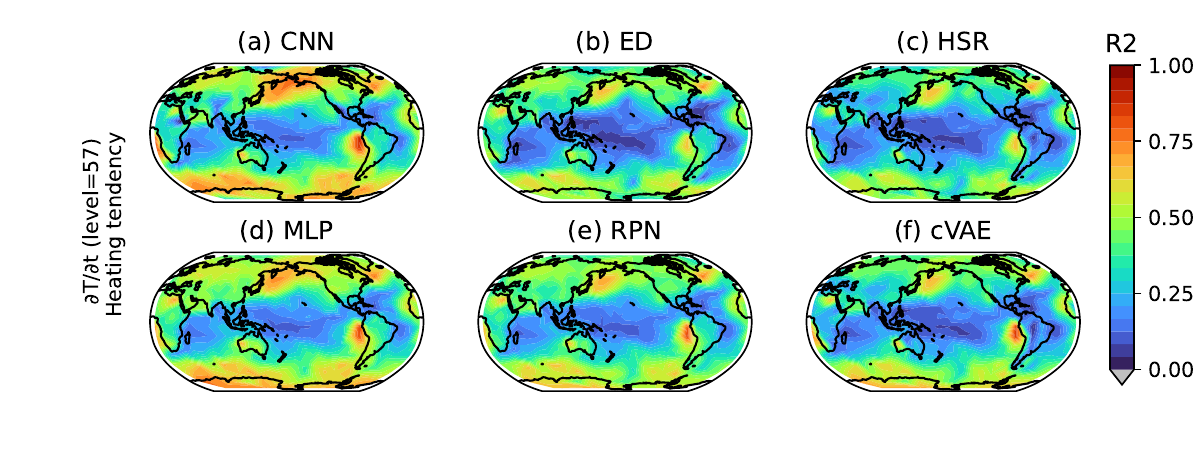}
     
    \caption{Global maps of R$^2$ of baseline models (built on the low-res, real-geography dataset). Grey shading shows locations with negative R$^2$ values.}
    \label{fig:SI_MAP_R2_3D_part1}
\end{figure}

\begin{figure}[H]
    \centering
    \includegraphics[width=.83\textwidth, trim={1.1cm 0.cm 0 0.cm},clip]{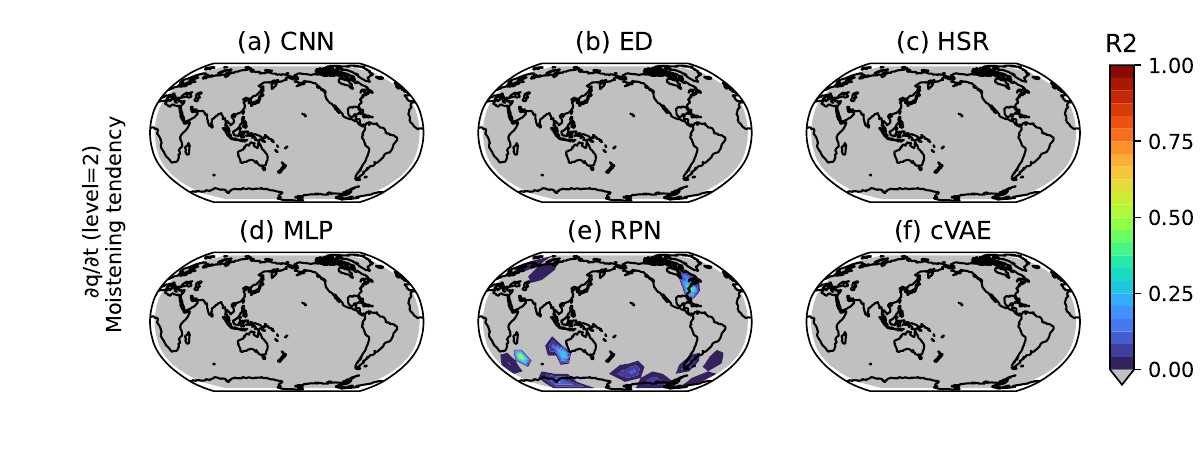}
    \includegraphics[width=.83\textwidth, trim={1.1cm 0.cm 0 0.cm},clip]{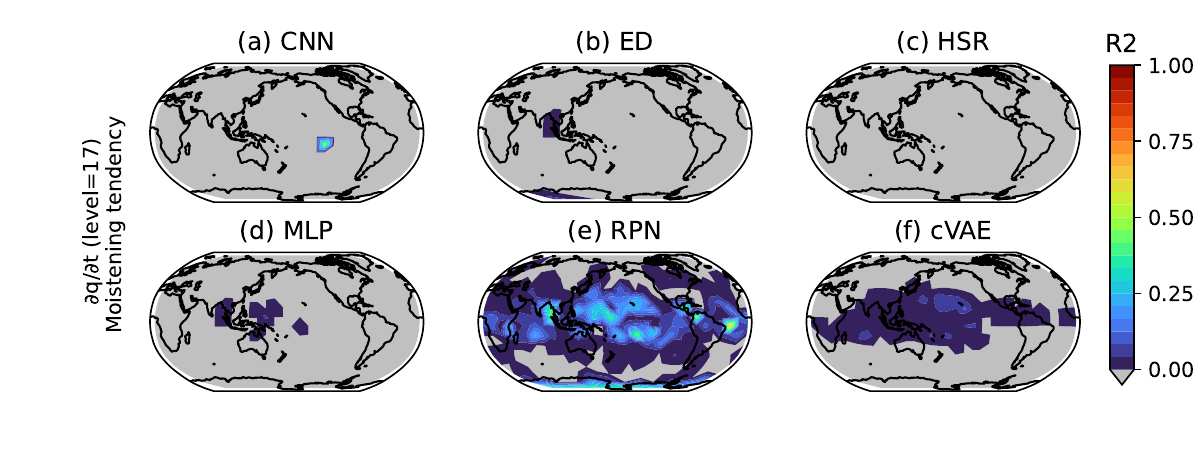}
    \includegraphics[width=.83\textwidth, trim={1.1cm 0.cm 0 0.cm},clip]{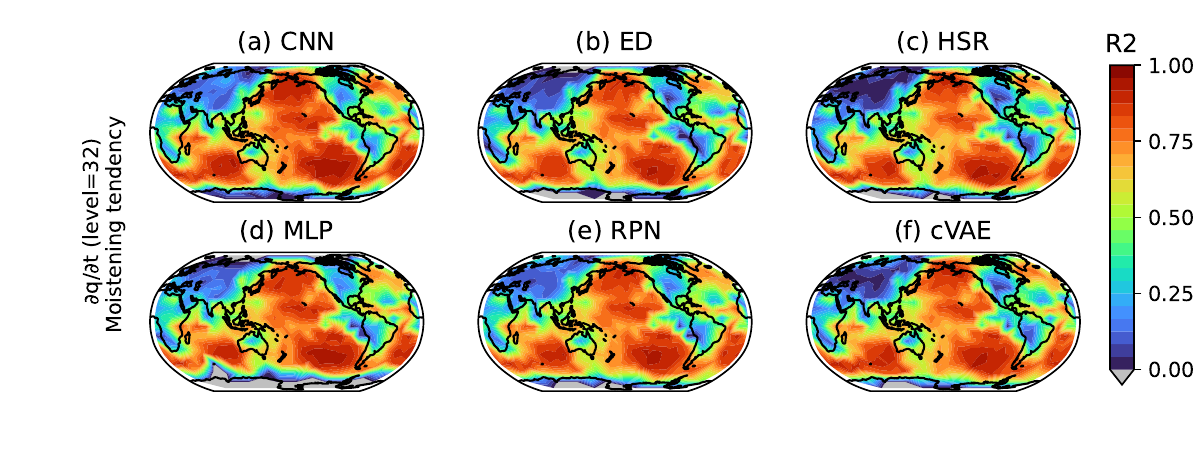}
    \includegraphics[width=.83\textwidth, trim={1.1cm 0.cm 0 0.cm},clip]{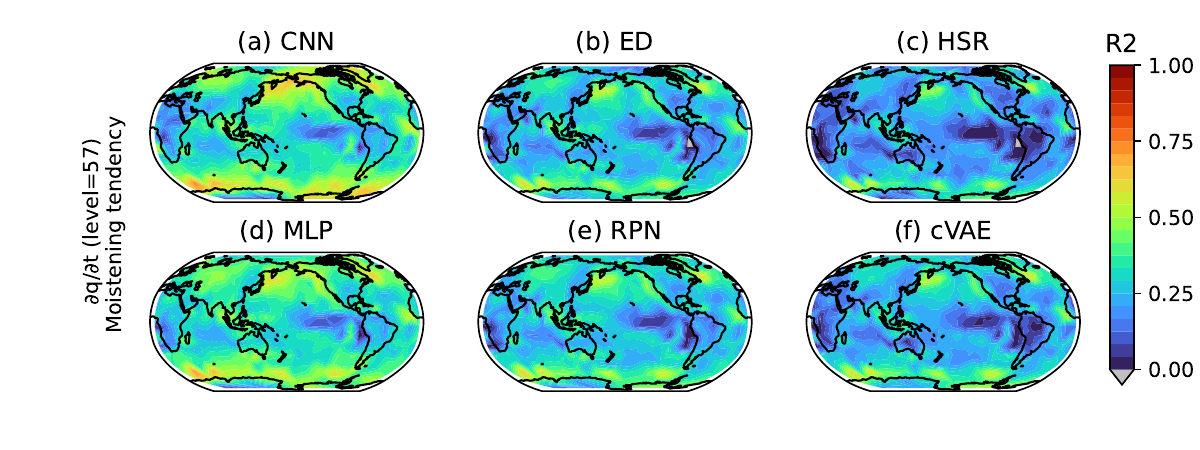}
     
    \caption{Global maps of R$^2$ of baseline models (built on the low-res, real-geography dataset). Grey shading shows locations with negative R$^2$ values.}
    \label{fig:SI_MAP_R2_3D_part1}
\end{figure}

\clearpage
% ======================================================
\section{Datasheet}
% ======================================================
% ===============================
\subsection*{Motivation}
% ===============================

\begin{enumerate}
    \item \textbf{For what purpose was the dataset created?}
        \textit{Our benchmark dataset was created to serve as a foundation for developing robust frameworks that emulate parameterizations for cloud and extreme rainfall physics and their interaction with other sub-resolution processes.}
    \item \textbf{Who created the dataset and on behalf of which entity?}
        \textit{The dataset was developed by a consortium of climate scientists and ML researchers listed in the author list.}
    \item \textbf{Who funded the creation of the dataset?}
        \textit{The main funding body is the National Science Foundation (NSF) Science and Technology Center (STC) Learning the Earth with Artificial Intelligence and Physics (LEAP). Other funding sources of individual authors are listed in the acknowledgment section of the main text.}
\end{enumerate}

% ===============================
\subsection*{Distribution}
% ===============================

\begin{enumerate}
    \item \textbf{Will the dataset be distributed to third parties outside of the entity (e.g., company, institution, organization) on behalf of which the dataset was created?}
        \textit{Yes, the dataset is open to the public.}
    \item\textbf{How will the dataset will be distributed (e.g., tarball on website, API, GitHub)?}
        \textit{The dataset will be distributed through Hugging Face and the code used for developing baseline models through GitHub.}
    \item \textbf{Have any third parties imposed IP-based or other restrictions on the data associated with the instances?}
        \textit{No.}
    \item \textbf{Do any export controls or other regulatory restrictions apply to the dataset or to individual instances?} 
        \textit{No.}
\end{enumerate}

% ===============================
\subsection*{Maintenance}
% ===============================

\begin{enumerate}
    \item \textbf{Who will be supporting/hosting/maintaining the dataset?}
        \textit{NSF-STC LEAP will support, host, and maintain the dataset.}
    \item \textbf{How can the owner/curator/manager of the dataset be contacted (e.g., email address)?}
        \textit{The owner/curator/manager(s) of the dataset can be contacted through following emails: Sungduk Yu (sungduk@uci.edu), Michael S. Pritchard (mspritch@uci.edu) and LEAP (leap@columbia.edu).}
    \item \textbf{Is there an erratum?}
        \textit{No. If errors are found in the future, we will release errata on the main web page for the dataset (https://leap-stc.github.io/ClimSim).}
    \item \textbf{Will the dataset be updated (e.g., to correct labeling errors, add new instances, delete instances)?}
        \textit{Yes, the datasets will be updated whenever necessary to ensure accuracy, and announcements will be made accordingly. These updates will be posted on the main web page for the dataset (https://leap-stc.github.io/ClimSim).}
    \item \textbf{If the dataset relates to people, are there applicable limits on the retention of the data associated with the instances (e.g., were the individuals in question told that their data would be retained for a fixed period of time and then deleted?)}
        \textit{N/A}
    \item \textbf{Will older version of the dataset continue to be supported/hosted/maintained?}
        \textit{Yes, older versions of the dataset will continue to be maintained and hosted.}
    \item \textbf{If others want to extend/augment/build on/contribute to the dataset, is there a mechanisms for them to do so?}
        \textit{No.}
\end{enumerate}

% ===============================
\subsection*{Composition}
% ===============================

\begin{enumerate}
    \item \textbf{What do the instance that comprise the dataset represent (e.g., documents, photos, people, countries?)}
        \textit{Each instance includes both input and output vector pairs. These inputs and outputs are instantaneous snapshots of atmospheric states surrounding detailed numerical calculations to be emulated.}
    \item \textbf{How many instances are there in total (of each type, if appropriate)?}
        \textit{The high-resolution dataset (ClimSim\_high-res) includes 5,676,480,000 instances, and each low-resolution dataset (ClimSim\_low-res and ClimSim\_low-res\_aqua-planet) includes 100,915,200 instances.}
    \item \textbf{Does the dataset contain all possible instances or is it a sample of instances from a larger set?}
        \textit{The datasets contain 80\% of all possible instances. The rest 20\% are reserved as the holdout test set, which will be released once enough models using \dataset{} are developed by independent groups.}
    \item \textbf{Is there a label or target associated with each instance?}
        \textit{Yes, each instance includes both input and target (prediction) variables.}
    \item \textbf{Is any information missing from individual instances?}
        \textit{No.}
    \item \textbf{Are there recommended data splits (e.g., training, development/validation, testing)?} 
        \textit{We have a hard split between the training/validation set and the test set. The first 8 simulation years-worth dataset is reserved for the training/validation set, and the last 2 simulation years-worth dataset is reserved for the test set. However, we do not have specific recommendations on the split within the training/validation set.}
    \item \textbf{Are there any errors, sources of noise, or redundancies in the dataset?}
        \textit{There is one redundancy. Input variable ``state\_pmid'' is redundant since it is a linear function of ``state\_ps''.}
    \item \textbf{Is the dataset self-contained, or does it link to or otherwise rely on external resources (e.g., websites, tweets, other datasets)?}
        \textit{The dataset is self-contained.}
    \item \textbf{Does the dataset contain data that might be considered confidential?}
        \textit{No.}
    \item \textbf{Does the dataset contain data that, if viewed directly, might be offensive, insulting, threatening, or might otherwise cause anxiety?}
        \textit{No.}
\end{enumerate}

% ===============================
\subsection*{Collection Process}
% ===============================

\begin{enumerate}
    \item \textbf{How was the data associated with each instance acquired?}
        \textit{The data associated with each instance is acquired from a series of simulations of a global climate model called E3SM-MMF. References for E3SM-MMF are provided in Section 3 of the main text.}
    \item \textbf{What mechanisms or procedures were used to collect the data (e.g., hardware apparatus or sensor, manual human curation, software program, software API)? }
        \textit{We used many NVIDIA A100 GPU nodes in a high-performance computing cluster called Perlmutter (operated by the U.S. Department of Energy) to run the E3SM-MMF simulations.}
    \item \textbf{Who was involved in the data collection process (e.g., students, crowdworkers, contractors) and how were they compensated (e.g., how much were crowdworkers paid)?}
        \textit{Regular employees (e.g., scientists and postdocs) at UC Irvine, LLNL, and SNL were involved in the data collection process. No crowdworkers were involved during the data collection process.}
    \item \textbf{Does the dataset relate to people?}
        \textit{No.}
    \item \textbf{Did you collect the data from the individuals in questions directly, or obtain it via third parties or other sources (e.g., websites)?}
        \textit{We obtained the dataset from computer simulations of Earth's climate.}
\end{enumerate}

% ===============================
\subsection*{Uses}
% ===============================

\begin{enumerate}
    \item \textbf{Has the dataset been used for any tasks already?}
        \textit{No, this dataset has not been used for any tasks yet.}
    \item \textbf{What (other) tasks could be the dataset be used for?}
        \textit{Please refer to Section 5 in the main manuscript for other applications.}
    \item \textbf{Is there anything about the composition of the dataset or the way it was collected and preprocessed/cleaned/labeled that might impact future uses?}
        \textit{The current composition of the datasets are self-sufficient to build a climate emulator. However, it misses some extra variables, which are not essential for such climate emulators but necessary to strictly enforce physical constraints (see Section 4.5 of the main text). We plan to include these extra variables in the next release. Any changes in the next release and update to user guidelines will be documented and shared through the dataset webpage (https://leap-stc.github.io/ClimSim).}
    \item \textbf{Are there tasks for which the dataset should not be used?}
        \textit{No.}
\end{enumerate}

\newpage
\bibliography{refs}